\documentclass{ametsocV6.1}

\usepackage{mathtools}
\usepackage{color}
\usepackage{multirow}
\usepackage{enumerate}
\usepackage{float}
\usepackage{subfig}
\usepackage{tabto}
\usepackage[ruled,vlined]{algorithm2e}
\renewcommand{\listalgorithmcfname}{Liste des algorithmes}
\SetKwInput{KwInit}{Init}
\SetKwInput{KwProcedures}{List of procedures}
\SetKwInput{KwPreproc}{Preprocessing}
\SetKwInput{KwPostproc}{Postprocessing}

\usepackage{tikz, tikz-3dplot}
\usepackage{pgfplots}
\tikzset{font=\normalsize}
\usetikzlibrary{decorations.text}
\usetikzlibrary{decorations.pathreplacing}
\usetikzlibrary{fadings}
\usetikzlibrary{positioning, fit, shapes.arrows, arrows.meta, shapes,calc,chains,scopes}
\usetikzlibrary{matrix,arrows,backgrounds}
\usetikzlibrary{tikzmark}
\usetikzlibrary{shapes.multipart}
\tikzset{elementwiseoperation/.style={circle, draw=green!55!blue, fill=green!55!blue, inner sep=0pt},
    elementwisefunction/.style={ellipse, draw=green!55!blue, fill=green!55!blue, inner sep=1pt},
    ct/.style={rectangle,, draw, minimum width=1cm, inner sep=1pt},
    sqt/.style={rectangle, draw=green!55!blue,fill=green!55!blue, minimum width=1cm, minimum height=1cm},
    sqt2/.style={circle, draw, fill=blue, minimum width=2cm, minimum height=2cm},
    gt/.style={rectangle, draw, minimum width=4mm, minimum height=3mm, inner sep=1pt},
    mylabel/.style={font=\scriptsize\sffamily},
    neuron/.style={ 
    circle,draw,thick, 
    inner sep=0pt, 
    minimum size=2.5em, 
    node distance=1ex and 2em, 
    fill=green!55!blue,
  },
  group/.style={ 
    rectangle,draw,thick, 
    inner sep=0pt, 
  },
  io/.style={ 
    neuron, 
    fill=gray!15, 
  },
  conn/.style={ 
    -{Straight Barb[angle=60:2pt 3]}, 
    thick, 
  },}%

\makeatletter
\renewcommand*\env@matrix[1][\arraystretch]{%
  \edef\arraystretch{#1}%
  \hskip -\arraycolsep
  \let\@ifnextchar\new@ifnextchar
  \array{*\c@MaxMatrixCols c}}
\makeatother

\newcommand{\bvec}[1]{{\mathbf{#1}}}
\newcommand{\x}[0]{\bvec{x}}
\newcommand{\s}[0]{\bvec{s}}
\newcommand{\y}[0]{\bvec{y}}

\title{Neural variational Data Assimilation with Uncertainty Quantification using SPDE priors}

\authors{Maxime Beauchamp,\aff{a,b} \correspondingauthor{Maxime Beauchamp, maxime.beauchamp@imt-atlantique.fr} 
Ronan Fablet,\aff{a} 
Simon Benaichouche\thanks{Simon Benaichouche's current affiliation: INRIA, Rennes, France},\aff{a} 
Pierre Tandeo,\aff{a} 
Nicolas Desassis,\aff{c} 
Bertrand Chapron,\aff{d} 
}

\affiliation{\aff{a}{IMT Atlantique, 655 Av. du Technopôle, 29280 Plouzané, France}\\
\aff{b}{Danish Meteorological Institute, Lyngbyvej 100, 2100 Copenhagen, Denmark}\\
\aff{c}{Mines ParisTech, Centre de Géosciences, 35 Rue Saint-Honoré, 77300 Fontainebleau, France}\\
\aff{d}{IFREMER, 1625 Rte de Sainte-Anne, 29280 Plouzané, France}
}

\abstract{The spatio-temporal interpolation of large geophysical datasets has historically been addressed by Optimal Interpolation (OI) and more sophisticated equation-based or data-driven Data Assimilation (DA) techniques. Recent advances in the deep learning community enables to address the interpolation problem through a neural architecture incorporating a variational data assimilation framework. The reconstruction task is seen as a joint learning problem of the prior involved in the variational inner cost, seen as a projection operator of the state, and the gradient-based minimization of the latter. Both prior models and solvers are stated as neural networks with automatic differentiation which can be trained by minimizing a loss function, typically the mean squared error between some ground truth and the reconstruction. Such a strategy turns out to be very efficient to improve the mean state estimation, but still needs complementary developments to quantify its related uncertainty. In this work, we use the theory of Stochastic Partial Differential Equations (SPDE) and Gaussian Processes (GP) to estimate both space-and time-varying covariance of the state. Our neural variational scheme is modified to embed an augmented state formulation with both state and SPDE parametrization to estimate. We demonstrate the potential of the proposed framework on a spatio-temporal GP driven by diffusion-based anisotropies and on realistic Sea Surface Height (SSH) datasets. We show how our solution reaches the OI baseline in the Gaussian case. For nonlinear dynamics, as almost always stated in DA, our solution outperforms OI, while allowing for fast and interpretable online parameter estimation.}

\begin{document}

\maketitle


\section{Introduction}

Over the last decade, the emergence of large spatio-temporal datasets both coming from remote sensing satellites and equation-based numerical simulations has been noticed in Geosciences. As a consequence,  the need for statistical methods able to handle both the size and the underlying physics of these data is growing. Data assimilation (DA) is the traditional framework used by geoscientists to merge these two types of information, data and model, by propagating information from observational data to areas with missing data. When no deterministic numerical outputs are available, the classic approach stems from the family of the so-called Optimal Interpolation (OI) techniques, also being at the core of the statistical DA methods \citep{asch_data_2016}. For the production of gridded geophysical maps, most of operational products either derived from OI and DA schemes allow us to estimate the mesoscale components of the targeted variables. In the specific case of Sea Surface Height (SSH) for instance, even if satellite altimetry provides capabilities to inform the mesoscale ocean geostrophic currents, the gridded product fails in merging observations and background model with consistent temporal and spatial resolutions able to retrieve fine mesoscale structures lower than 150-200 km at mid-latitudes, while they are key for general ocean circulation \citep{Su_2018}. Then, recent efforts of the DA community have been made to counteract this lack in the gridded SSH products, amongst them the dynamical optimal interpolation (DOI) \citep{Ubelmann_2015}, Multiscale Interpolation Ocean Science Topography (MIOST) \citep{Ubelmann_2021} or Back-and-Forth Nudging with Quasi-Geostrophic forward model approximation (BFN-QG) \citep{LeGuillou_2023}.\\

From another point of view, deep learning (DL)frameworks are currently knowing an intense period of scientific contributions to revisit statistical methods with neural network formulation. State-of-the-art methods leverage DL to better extract the information from the observations compared to the classical DA and OI approaches, in which fine-tuning of the DA scheme parametrizations (background and observation error covariances) demands itself some methodological and computational effort, see e.g. \citet{Tandeo_2020}. Together with the popular Gaussian simplifications involved in most of DA operational schemes \citep{asch_data_2016}, that may not be realistic, especially when leadtime increases in short-term forecasting, this makes DL appealing to get over simplified assumptions in the DA scheme, though some approaches already exist to alleviate the Gaussian assumptions in nonlinear geophysical problems, see e.g. \citet{Kurosawa_2023}. End-to-end learning architectures are also designed being backboned on DA schemes \citep{Boudier_2023,Rozet_2023, Fablet_2021}, so that they draw from the bayesian formalism to learn all the components of the DA procedure (prior model, observation operator, numerical solver, etc.) at once. Last, over the last few years, the successfull applications of generative models to geophysical datasets paves the way to combine DL together with Uncertainty Quantification (UQ) by efficient sampling strategies of the posterior distribution. Though, to our knowledge, no end-to-end-combination of DA, DL and UQ has yet been proposed. \\

That is why in this work, we draw from preliminary neural schemes inspired by variational data assimilation, see e.g. \citep{Fablet_2021}, to jointly learn the SPDE parametrization of a surrogate stochastic prior model of the evolution equation, together with the solver of the minimization problem. Because the parameters of the SPDE remain initially unknown, they are embedded in the optimization process using an augmented state formulation, commonly used in data assimilation, see e.g. \citep{Ruiz_2013}. The solver is still based on an iterative residual scheme \citep{Fablet_2021} to update the analysis state. Here, the analysis stands for the expectation of the state given the observations, together with the SPDE parametrization maximizing their likelihood given the true states used during the training. The SPDE equation can then be seen as a tangent linear model of the prior along the data assimilation window, from which we provide uncertainty quantification (UQ) throughout its precision matrix. Using such a stochastic prior entails the possibility of generating huge members in the posterior pdf, after conditioning of the prior samples by our neural variational scheme. Also, if the training dataset is large enough, the method provides an efficient way to estimate \textit{online} the SPDE parametrization for any new sequence of input observations, without any additional inference to make. In the end, the key contributions of this work are four-fold:
\begin{itemize} 
\item We develop the explicit representation of the considered SPDE prior. It relies on the analytical expression for the SPDE-based precision matrix of any state trajectory, based on a finite-difference discretization of the grid covered by the tensors involved in our neural scheme; 
\item We exploit this SPDE parametrization as surrogate prior model in the proposed variational formulation and leverage a trainable gradient-based  solver to address jointly the interpolation of the state trajectory and the estimation of SPDE parameters from irregularly-sampled observations. The end-to-end training of the solver targets both the expectation of the state given the observations, together with the SPDE parametrization maximizing its likelihood given the true states;
\item The SPDE prior paves the way to uncertainty quantification through the sampling of the prior pdf and the conditioning by the neural gradient-based solver;
\item Two applications of this framework are provided: first, a GP driven by a spatio-temporal SPDE with spatially varying diffusion parameters is compared to the optimal solution; then a Sea Surface Height realistic dataset is used to demonstrate how the proposed framework is also relevant for non-gaussian and non-linear dynamics.
\end{itemize} 

To present these contributions, the paper is structured as follows. In Section 2, we provide a preliminary background on SPDE-based Optimal Interpolation, Data assimilation and Deep Learning for DA, that we aims to bridge together in this work. We also remind how UQ is usually handled in these different fields. In Section 3, we present how we extend neural solvers to embed SPDE prior parametrizations for UQ. Finally, Section 4 provides  two applications of this framework in the Gaussian case and for realistic Sea Surface Height datasets. 

\section{Background: Data Assimilation, Uncertainty Quantification and Machine Learning}
\label{back}

In this work, we target the reconstruction of a probabilistic spatio-temporal state sequence $\mathbf{x}=\lbrace \mathbf{x}_k(\mathcal{D}) \rbrace, \x_k \in \mathbb{R}^m$ given the partial and potentially noisy observational dataset $\mathbf{y}(\Omega)=\lbrace \mathbf{y}_k(\Omega_k) \rbrace, \y_k \in \mathbb{R}^{p_k}$, where $\Omega=\lbrace \Omega_k \rbrace \subset \mathcal{D}$ denotes the subdomain with observations and index $k$ refers to time $t_k$. To do so, we aim at bridging Data Assimilation (DA), Uncertainty Quantification (UQ) and deep learning-based (DL) methods to propose SPDE-based extensions of the so-called 4DvarNet neural variational scheme \citep{Fablet_2021}, as a generic interpolation and short-term forecasting tools.  We provide here a brief presentation of these three literatures, focusing only on what is useful in our framework, with an additional description of how UQ is usually handled among these communities. For a more exhaustive review, \citet{Cheng_2023} provides a detailed presentation of machine learning techniques with data assimilation and uncertainty quantification for dynamical systems. 

\subsection{Optimal Interpolation (OI) and Data Assimilation (DA)} 

\paragraph{Classic formulations.} As very basic details to ease the link with the other components of this work, we remind that when no dynamical model is available as a prior information, the covariance-based Optimal Interpolation, see e.g. \citep{chiles_2012}, is given by:
\begin{align}
\label{covariance_OI}
\mathbf{x}^\star = \mathbf{P}_{\mathbf{x}\mathbf{y}}  \mathbf{P}_{\mathbf{y}\mathbf{y}}^{-1} \mathbf{y}
\end{align}
where $\mathbf{P}_{\mathbf{x}\mathbf{y}}$ and $\mathbf{P}_{\mathbf{y}\mathbf{y}}$ are covariance matrices coming from the covariance $\tilde{\mathbf{P}}$ of the observation and state vector $\begin{bmatrix} \mathbf{y} & \mathbf{x} \end{bmatrix}^\mathrm{T}$:
\begin{align}
\label{global_cov}
\tilde{\mathbf{P}} = \begin{bmatrix} \mathbf{P}_{\mathbf{x}\mathbf{x}} &  \mathbf{P}_{\mathbf{x}\mathbf{y}} \\ \mathbf{P}_{\mathbf{x}\mathbf{y}}^{\mathrm{T}}  & \mathbf{P}_{\mathbf{y}\mathbf{y}} \end{bmatrix}
\end{align}
Broadly speaking, when the prior information is available, typically as high dimensional numerical models in geosciences, see e.g. \citet{carassi_2018}, two main categories of DA \citep{evensen_data_2009, evensen_2022} exists: variational and sequential methods. They both aims at minimizing some energy or functional involving an equation-based dynamical prior and an observation term. Drawing from the link established with Gaussian Processes, we can also consider the case of noisy observations and ease the link with data assimilation formalism, see \citet{Sarkka_2012, Sarkka_2013, Grigorievskiy_2016}. The state space model corresponding to the GP regression problem writes:
\begin{align} 
\label{da_eqs}
 \begin{cases}
 \mathbf{x}_{t+dt} & = \mathbf{M}_{t+dt}\mathbf{x}_{t} + \boldsymbol{\eta}_t \\
  \y_t & = \mathbf{H}_t\x_t+\boldsymbol{\varepsilon}_t
 \end{cases}
\end{align}
where $\boldsymbol{\eta}_t$ is the m-dimensional noise process and the evolution equation is defined by the feedback linear operator matrix $\mathbf{M}_{t+dt}$. $\mathbf{H}_t$ is the observation operator at time $t$ mapping the state space to the observation space and $\boldsymbol{\varepsilon}_t$ the observational error with covariance matrix $\mathbf{R}_t$. Based on this time-dependent notations, we also consider global observation operator $\mathbf{H}$ with dimensions $(L \times p_k) \times (L \times m)$ and global observational error covariance matrix $\mathbf{R}$ with dimensions $(L \times p_k) \times (L \times p_k)$ as block diagonal matrices whose each block respectively contains the time-dependent observation operator and observational error covariance matrix $\mathbf{H}_t$ and $\mathbf{R}_t$.

\paragraph{The Stochastic Partial Differential Equation (SPDE) approach in DA.} The Optimal Interpolation implies to factorize dense covariance matrices which is an issue when the size of spatio-temporal datasets is large. Reduced rank approximations, see e.g. \citep{cressie_statistics_2015} have already been investigated to tackle this specific problem. More recently, the use of sparse precision matrices has also been proposed by using tapering strategies \citep{furrer_2006, bolin_2016} or by making use of the link seen by \citet{lindgren_2011} between Stochastic Partial Differential Equations (SPDE) and Gaussian Processes. For the latter, if the original link was made through the Poisson SPDE equation \citep{whittle_1953}:
\begin{align}
\label{isotropic_SPDE}
(\kappa^2 - \Delta)^{\alpha/2} X = \tau Z \ ;
\end{align}
where $\Delta = \sum_{i=1}^d \frac{\partial^2}{\partial s_i^2}$ denotes the Laplacian operator, $Z$ is a standard Gaussian white noise, $\kappa = 1/a$, $a$ denotes the range of the Gaussian Process $X$, $\alpha = \nu + d/2$ and $\tau$ relates to the marginal variance of $X$. It can be extended to more complex linear SPDE involving physical processes like advection or diffusion \citep{lindgren_2011,fuglstad_2015a, clarotto_2022}. We precise from here that in this study, we only use specific values of $\alpha$ that ensures $\alpha/2$ to be an integer, which considerably eases the discretization of the SPDE and  its integration in our work. Though, recent contributions has been done, see e.g. \citep{Bolin_2020}, to address the general case of the fractional operator that could be also considered here, with additional developments. The SPDE-based OI formulation uses precision matrix formalism, as the inverse of the covariance matrix $\mathbf{Q}=\tilde{\mathbf{P}}^{-1}=(dxdy)/\tau^2 \cdot \mathbf{B}^{\mathrm{T}}\mathbf{B}$, see Eq. (\ref{global_cov}) where $\mathbf{B}$ is the discretized version of the fractional differential operator $(\kappa^2-\Delta)^{\alpha/2}$, and $dx$, $dy$ are the spatial grid step sizes:
\begin{align}
\label{precision_OI}
  \mathbf{x}^\star=-\mathbf{Q}_{\mathbf{x}\mathbf{x}}^{-1}\mathbf{Q}_{\mathbf{x}\mathbf{y}}\mathbf{y}
\end{align}
By construction, $\mathbf{Q}$ is sparse which means that we solve a system with sparse Cholesky of complexity $\mathcal{O}(m^{3/2})$, while the general Cholesky algorithm is of complexity $\mathcal{O}(m^{3})$. Thus, it opens new avenue to cope with massive observational datasets in geosciences while making use of the underlying physics of such processes. Let note that the so-called SPDE-based approach can also be used as a general spatio-temporal model, even if it is not physically motivated, since it provides a flexible way to handle local anisotropies of a large set of geophysical processes. It has known numerous applications in the past few years, see e.g. \citet{Sigrist_2015, Fuglstad_2015}. Though, when considering a spatio-temporal advection-diffusion SPDE, the parameters generally vary continuously across space and/or time making their estimation an additional problem to the original interpolation task. This estimation generally relies on off-line strategies \citep{Fuglstad_2015} embedded in hierarchical models, which can be another computational issue while the set of parameters estimated does not automatically transfer to a new dataset. \\
Regarding the transfer of SPDE formulation in the DA formalism, rewriting the covariance-based Eq. (\ref{global_cov}) in terms of precision matrix leads to:
\begin{align}
\tilde{\mathbf{P}} &= 
\begin{bmatrix} \mathbf{P}_{\x\x}  & \mathbf{P}_{\x\x}  \mathbf{H}^{\mathrm{T}} \\
 \mathbf{H} \mathbf{P}_{\x\x}  & \mathbf{H} \mathbf{P}_{\x\x}  \mathbf{H}^{\mathrm{T}} + \mathbf{R}
\end{bmatrix}
, \ \ \ \tilde{\mathbf{Q}} = 
\begin{bmatrix} \mathbf{Q}_{\x\x} +  \mathbf{H}^{\mathrm{T}} \mathbf{R}^{-1} \mathbf{H} & -\mathbf{H}^{\mathrm{T}}\mathbf{R}^{-1}  \\  -\mathbf{R}^{-1}\mathbf{H}  & \mathbf{R}^{-1}
\end{bmatrix}
\end{align}
which gives an other version of Eq. (\ref{precision_OI}):
\begin{align}
\label{final_eq_oi}
\mathbf{x}^\star & = \left( \mathbf{Q}_{\x\x}  +  \mathbf{H}^{\mathrm{T}} \mathbf{R}^{-1} \mathbf{H} \right)^{-1}\mathbf{H}^{\mathrm{T}}\mathbf{R}^{-1} \y
\end{align} 
whose posterior precision matrix of state $\mathbf{x}^\star$ is $\mathbf{Q}(\x|\y)=\mathbf{Q}_{\x\x}  +  \mathbf{H}^{\mathrm{T}} \mathbf{R}^{-1} \mathbf{H}$. This type of formulation makes the link between DA framework and SPDE-based formalism.

\subsection{Machine and Deep Learning (ML/DL)}

From another point of view, deep learning frameworks are currently knowing an intense period of scientific contribution to revisit statistical methods with neural network formulation. The latter enables to use automatic differentiation embedded in the gradient-based optimization as a way of solving traditional inverse problems. Several approaches have been investigated, among others we can think of using DL as a substitute for one component of the DA procedure: surrogate dynamical models making extensive use of state-of-the-art neural networks such as LSTM \citep{Nakamura_2021}, UNet \citep{Doury_2023} or Transformers \citep{Tong_2022}, reduced order model to compute DA schemes in latent space or with super-resolution component \citep{brajard_2021}, modeling of model/observation error covariance \citep{Cheng_2022,Sacco_2022,Sacco_2023}. The other way rely on end-to-end learning of the entire DA system instead of using DL techniques to address one aspect of DA algorithm among its three main blocks: forward model and its error, observation operator and its error, and DA scheme (EnKF, 4DVar, etc.). Such approaches may rely on applying state-of-the-art neural architectures to map observation data to the hidden state sequence, see e.g. UNet \citep{Li_2022} ,Transformers \citep{Shi_2022} or LSTM architectures \citep{Martin_2023}. In particular, when the problem relates to space-time interpolation of partial and noisy observation of geophysical fields, DA-inspired neural schemes have been recently proposed, see e.g. \citet{Boudier_2023, Rozet_2023, Fablet_2021}. The latter specifically suggests a joint learning of prior models and solvers as an end-to-end data-driven variational data assimilation scheme. The so-called 4DVarNet neural scheme is introduced: it involves an implicit iterative gradient-based LSTM solver to minimize a variational cost, close to what is encountered in 4DVar data assimilation \citep{carassi_2018}. In this variational cost, the dynamical prior is no longer equation-based but is stated as a trainable neural network learnt during the training process. Then, automatic differentiation is used to compute the gradient of the variational cost during the gradient-based iterations, instead of requiring the computation of complex and costly adjoint models \citep{asch_data_2016}. Drawing from this framework, a neural optimal interpolation scheme has also been proposed \citep{beauchamp_2022b} to reach OI performance with a linear scaling of the solution on the number of space-time variables, leading to a significant speed up in the computation of the solution.\\ 

\subsection{Uncertainty Quantification} 

\paragraph{UQ in DA.} Based on its bayesian formalism, Optimal Interpolation provides uncertainty through a posterior covariance matrix $\mathbf{P}_{\x|\y}$. Because the background $\x^b$ is generally considered as stationary, this covariance matrix is mainly driven by the sampling of the observation, see e.g. \citep{Zhen_2020}, which may not be realistic for dynamical systems. Relying on a similar framework, DA schemes provide the deterministic part of the evolution equation in Eq. \ref{da_eqs} as numerical model outputs, and another probabilistic model has to be given for the distribution of the noise. In addition, under Gaussian assumption of the model likelihood, closed forms exist for the mean and standard deviation of the posterior prediction. Ensemble-based sequential schemes also provide a simple way to compute flow-dependant empirical posterior covariance \citep{Song_2013} that counteracts the sampling issue used in OI or in most of variational schemes.  Hybrid methods in DA \citep{asch_data_2016} aims at combining both ensemble-based methods together with variational schemes to benefit from the assets of each methods. 

\paragraph{UQ for ML approaches.} As stated above, DL for UQ generally relies on the estimation of the stochastic components of the DA scheme, namely the covariance model and/or observation errors. More recently, ML schemes aims at substituting the entire DA scheme, either with an explicit model of the posterior distribution, or by a sampling strategy of the latter. On the first approach, because the true posterior is both computationally and analytically intractable, a popular strategy is to estimate an approximate posterior distribution model with trainable parametrization that involves the minimization of the Kullback-Leibler Divergence between the two distributions, which is the similar that maximizing the Evidence Lower Bound (ELBO) \citep{Huang_2019} referred as variational inference \citep{Zhang_2021}. More recently, generative models are extensively used to draw samples in the prior distribution. Among them, GANs \citep{goodfellow_2014}, VAEs \citep{Kingma_2022}, normalizing flows \citep{Dinh_2017} and diffusion models \citep{Ho_2020} are the most popular. Once a way of sampling the prior is available, the posterior pdf can be obtained after conditioning through a classic DA scheme or method-related conditional generative models. This is the case in score-based data assimilation \citep{Rozet_2023} stated as diffusion models where Langevin iterative optimisation scheme can also embed the observation likelihood \citep{Ho_2020} to give a direct access to the posterior.  
 
\section{Neural variational schemes with SPDE priors}

\begin{figure}[H]
  \centering
  \includegraphics[width=12cm]{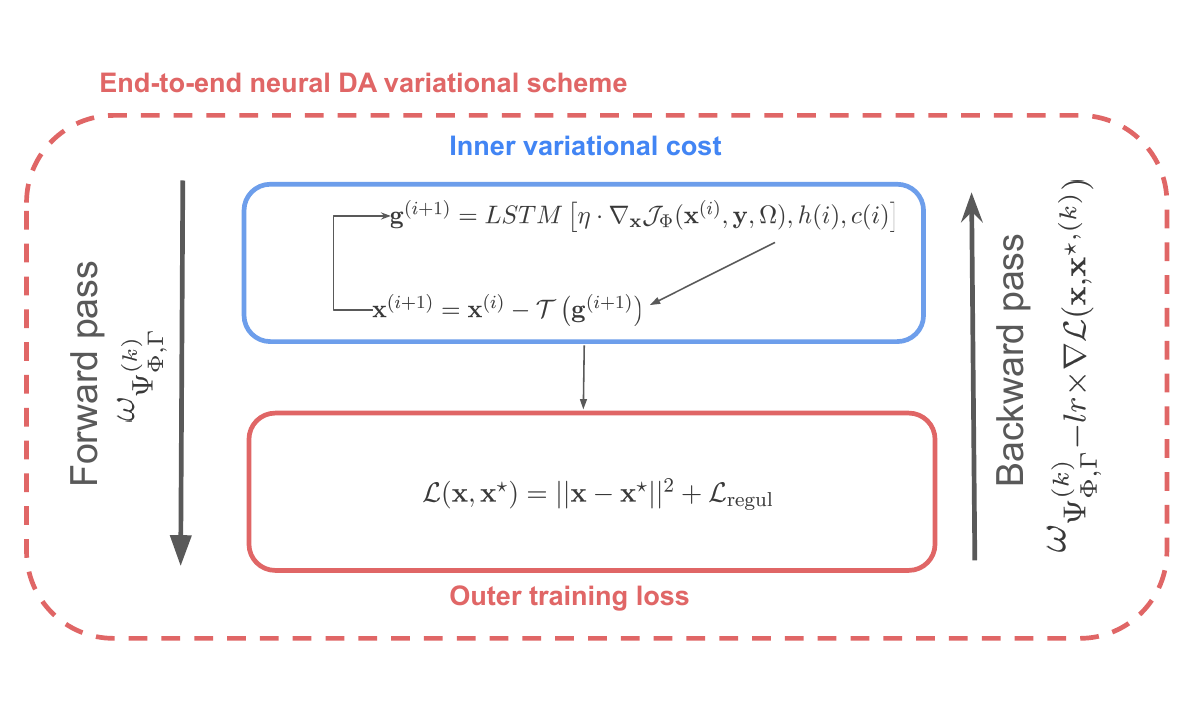}
   \caption{Schematic representation of the bi-level optimization scheme: the forward pass seeks to minimize the inner variational cost based on the current weights $\omega_{\Psi_{\Phi,\Gamma}}$ of both prior and solver, while the backward pass updates the previous weights based on the minimization of the outer training loss function}
   \label{4DVarNet_classic}
\end{figure}

In this section, we present how we draw from \citet{Fablet_2021} to embed an augmented state with SPDE prior parametrization in the neural variational scheme. The original strategy was based on the idea to replace every component of a classic data assimilation procedure, typically the forward deterministic model, the observation operator and the assimilation scheme by an end-to-end neural formulation whose every components are either predefined or trainable. Backboned on 4DVar formulation, the end-to-end neural scheme involves the optimization of the inner variational cost, see Fig. \ref{4DVarNet_classic}:
\begin{align}
  \label{var_cost}
  \mathcal{J}_{\Phi}(\mathbf{x},\mathbf{y},\Omega) & = || \mathcal{H}(\mathbf{x}) - \mathbf{y} ||^2_\Omega + \lambda || \mathbf{x} - \Phi(\mathbf{x}) ||^2
\end{align}
with a purely data-driven UNet-based projection operator $\Phi$ and a  trainable gradient-step descent:
\begin{equation}
\label{eq: lstm update}
\left \{\begin{array}{ccl}
     \mathbf{g}^{(i+1)}& = &  LSTM \left[ \eta \cdot \nabla_{\mathbf{x}}\mathcal{J}_{\Phi}(\mathbf{x}^{(i)} ,\mathbf{y},\Omega),  h(i) , c(i) \right ]  \\~\\
     \mathbf{x}^{(i+1)}& = & \mathbf{x}^{(i)} - {\mathcal{T}}  \left( \mathbf{g}^{(i+1)} \right )  \\
\end{array} \right.
\end{equation}
where $\mathbf{g}^{(i+1)}$ is the output of a two-dimensional convolutional Long-Short-Term-Memory (LSTM) \citep{arras_2019} using as input gradient $\nabla_{\mathbf{x}}\mathcal{J}_{\Phi}(\mathbf{x}^{(i)} ,\mathbf{y},\Omega)$, while $h(i)$ and $c(i)$ denotes the internal states of the LSTM. $\eta$ is a normalization scalar and ${\mathcal{T}}$ a linear or convolutional mapping. The overall 4DVarNet scheme, as a joint learning of both $\Phi$ and solver $\Gamma$ (set of $N$ predefined LSTM-based state updates) is denoted as $\Psi_{\Phi,\Gamma}$ with weights $\omega$ to learn. In \citep{beauchamp_2023a}, the latter are optimized based on a supervised outer loss function:   \\
\begin{equation}
\label{eq: loss 4dvar}
\mathcal{L}(\mathbf{x},\mathbf{x}^\star)= ||\mathbf{x}-\mathbf{x}^\star||^{\mathrm{2}} + \mathcal{L}_{\mathrm{regul}},
\end{equation}
typically the mean squared error between the reconstruction and the Ground Truth, with additional regularization terms, depending on both the specific application and the targeted ouputs of the end-to-end neural scheme. Because the end-to-end neural scheme involves both inner variational cost and outer training loss function, we sometimes use the term of bi-level optimization scheme to describe this class of methods.

\begin{figure}[H]
  \centering
  \includegraphics[width=12cm]{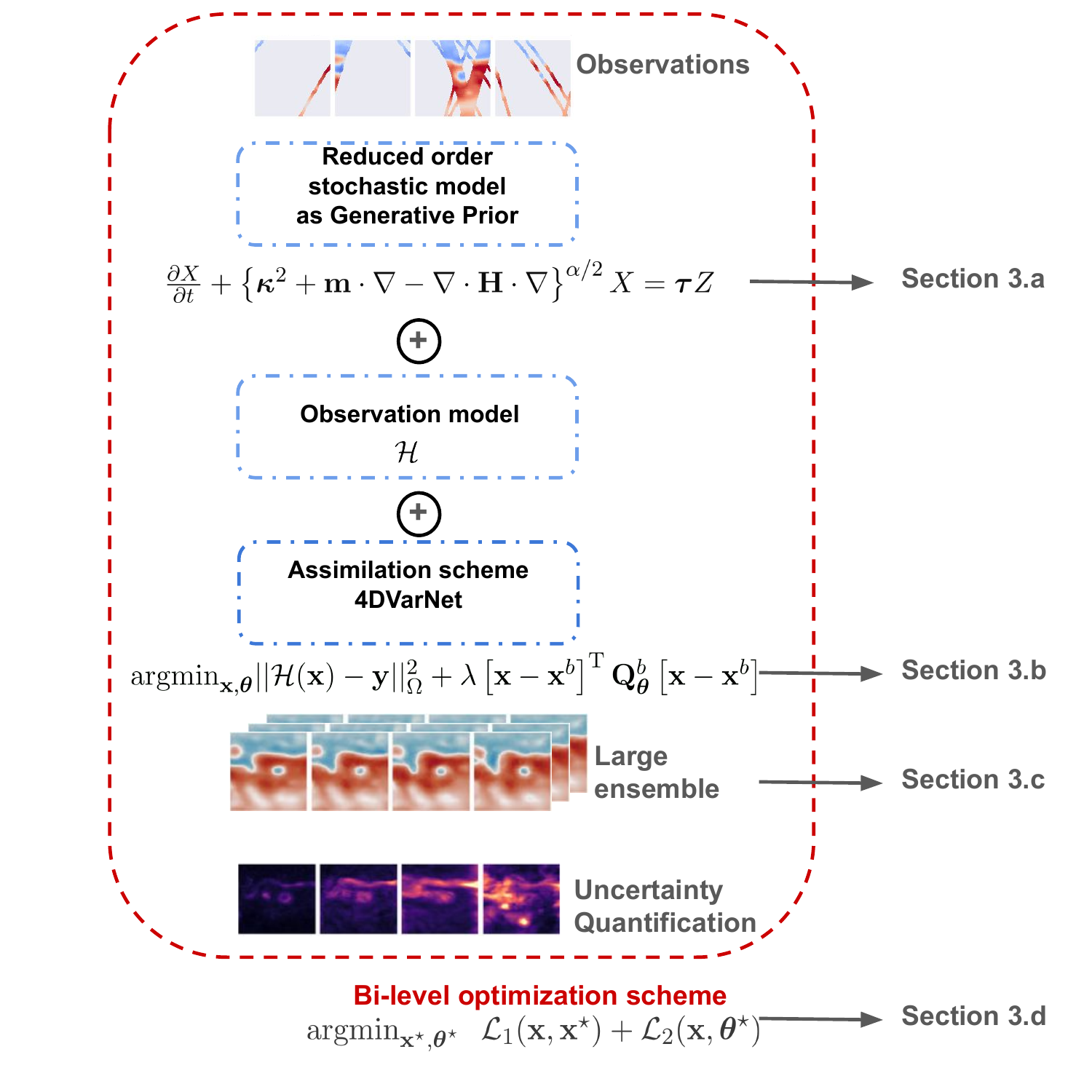}
   \caption{Adaptation of the original 4DVarNet scheme by replacing all the components of the data assimilation procedure by trainable components, namely a reduced order stochastic model used as generative prior, an observation model (predefined here as a masking operator but trainable if necessary), and an assimilation scheme with appropriate variational cost to optimize. All the components of the end-to-end neural architecture are optimized using an outer training loss function which is a combination of reconstruction loss and a likelihood regularization loss}
   \label{4DVarNet_SPDE}
\end{figure}

In this work, we propose to extend the previous scheme by demonstrating how we can draw from its generic formulation to derive a new bi-level optimization learning scheme (Section \ref{learning_scheme}), see also Fig. \ref{4DVarNet_SPDE}, in order to:
\begin{enumerate}
\item embed a trainable stochastic parametrization of the prior as physically-sounded reduced order models based on linear stochastic PDEs (Section \ref{spde_param});
\item reformulate the inner 4DVarNet variational cost in a similar way traditional DA  makes use of it, meaning that the regularization term involves the Mahanalobis distance between the state $\mathbf{x}$ and the SPDE-based prior distribution $\mathcal{N} \sim (\mathbf{x}^b,\mathbf{Q}^b_\theta)$ (Section \ref{neuralsolver});
\item produce fast and large set of realistic posterior ensembles leading to Uncertainty Quantification (Section \ref{uq_scheme}).
\end{enumerate}

\subsection{SPDE parametrization}
\label{spde_param}

In the original version of the neural scheme, the prior $\Phi$ is not easily interpretable: it acts as an encoding of the state $\mathbf{x}$ that helps in the gradient-based minimization process. In this work, we aim at bringing both explainability and stochasticity in the neural scheme by considering as a surrogate model for prior $\Phi$ an stochastic PDE (SPDE) instead of a given neural-based architecture. The continuous process associated with our state space describes the dynamical evolution of the spatio-temporal prior process $X_{\boldsymbol{\theta}}(\mathbf{s},t)$ as a spatio-temporal SPDE embedding the estimation of its parametrization $\boldsymbol{\theta}$, the latter controlling key physical behaviours such as local anisotropy, correlation range, and marginal variance: 
\begin{eqnarray}
\label{SPDE_st}
\mathcal{F}_{t,\boldsymbol{\theta}(t)} \lbrace{ X(\mathbf{s},t) \rbrace}=\tau(\mathbf{s},t) Z(\mathbf{s},t)
\end{eqnarray}
where operator $\mathcal{F}_{t,\boldsymbol{\theta}(t)}$ is here considered as linear, then the solution of Eq. (\ref{SPDE_st}) is a spatio-temporal Gaussian Process \citep{lindgren_2011}. Regarding the right-hand side noise of the SPDE, it is assumed separable, i.e. $Z(\mathbf{s},t) = Z_t(t) \otimes  Z_s(\mathbf{s})$ with $Z_t(t)$ a temporal white noise. $Z_s(\mathbf{s})$ can also be a spatial white noise or we may consider a colored noise to ensure more regularity on process $X$ accross time. \\

Following a state space formalism, the discretization of the stochastic process $X(\mathbf{s},t)$ is a multivariate gaussian vector $\x$ :
\begin{align}
\label{statespace_form}
 \x \sim \mathcal{N}\left(\x^b,\mathbf{Q}^b_{\boldsymbol{\theta}}\right)
\end{align}
where $\x^b$ is the deterministic mean of the state, typically \textit{the background}, a coarse approximation of the field in stationary formulations, or the \textit{forecast} in a dynamical data assimilation scheme. $\mathbf{Q}^b_{\boldsymbol{\theta}}$ is the precision matrix (inverse of the covariance matrix) of the state sequence  $\lbrace \x_0,\cdots,\x_{Ldt} \rbrace$. Because of the explicit link between linear stochastic PDEs and Gaussian Processes, we modify the variational formulation used in 4DVarNet schemes, see e.g. \citet{beauchamp_2023a}, by rewriting the matrix formulation of the regularization prior term of Eq. \ref{var_cost} as:
 \begin{align*}
    \mathbf{x}^\star &= \mathrm{argmin}_\mathbf{x} \mathcal{J}(\mathbf{x}, \mathbf{y}, \varOmega) = \mathrm{argmin}_\mathbf{x} ||\mathbf{H}\mathbf{x}-\mathbf{y}||_\varOmega^\mathrm{2}  + \lambda \left[ \mathbf{x}- \x^b \right]^\mathrm{T}\mathbf{Q}^b_{\boldsymbol{\theta}} \left[ \mathbf{x}- \x^b \right]\\
    \end{align*}
When $\x^b=\mathbf{0}$ (for simplification) and when denoting $\mathbf{L}$ the square root of the precision matrix $\mathbf{Q}^b_{\boldsymbol{\theta}}$, the identification with 4DVarNet schemes is direct for $\Phi=(1-\mathbf{L})$. Indeed, $|| \mathbf{x} ||^2_{\mathbf{Q}_{\boldsymbol{\theta}}} = \mathbf{x}^{\mathrm{T}}\mathbf{L}^{\mathrm{T}}\mathbf{L}\mathbf{x} = ||\mathbf{L}\mathbf{x}||^2 = ||\x -\Phi \cdot \x ||^2$. In this derived formulation, it is clear that if the SPDE is known, it can be embedded in the inner variational cost used by the LSTM iterative solver to optimize the outer training loss function. Drawing from the usual neural variational framework, the trainable prior is now SPDE-based, and the parameters $\boldsymbol{\theta}$ are embedded in the following augmented state formalism:
\begin{eqnarray}
 \tilde{\mathbf{x}} = \begin{bmatrix} \mathbf{x} & \boldsymbol{\theta} \end{bmatrix}^{\mathrm{T}}
\end{eqnarray}
The latent parameter $\boldsymbol{\theta}$ is potentially non stationary in both space and time and its size is directly related to the size of the data assimilation window $L$.

When dealing with geophysical fields, a generic class of non-stationary models generated by stochastic PDEs shall introduce some diffusion and/or advection terms. They are respectively obtained  by introducing a local advection operator $ \mathbf{m}(\mathbf{s},t) \cdot \nabla$ where $\mathbf{m}$ is a velocity field and a local diffusion operator $\nabla \cdot\mathbf{H}(\mathbf{s},t)\cdot\nabla$ where $\mathbf{H}$ acts as a two-dimensional diffusion tensor:
\begin{align} 
\label{spde_adv_diff}
\frac{\partial{X}}{\partial{t}}+\left\lbrace \boldsymbol{\kappa}^2(\mathbf{s},t) + \mathbf{m}(\mathbf{s},t) \cdot \nabla - \nabla \cdot\mathbf{H}(\mathbf{s},t)\cdot\nabla \right \rbrace^{\alpha/2} X(\mathbf{s},t)=\boldsymbol{\tau}(\mathbf{s},t) Z(\mathbf{s},t)
\end{align}

This way of handling spatio-temporal non-stationarities in SPDE models has been first mentioned in the original paper of \citet{lindgren_2011}, but because of the challenge of estimating the full set of space-time parameters, no works have been published to our knowledge pushing this framework into its completeness. Though, some parametrization of purely spatial diffusion process \citep{fuglstad_2015b} or stationary advection-dominated SPDE \citep{clarotto_2022} has been successfully applied.\\ 
Let stress that the advection-diffusion scheme is a good candidate for many geophysical datasets: this is the case in quasi-geostrophic approximation of Sea Surface Temperature (SST), see e.g. \citep{ubelmann_dynamic_2014}, but also in the dispersion of atmospheric pollutants \citep{chimere_2020} for instance. Then, this framework provides a generic and convenient way to bring more explainability in terms of space-time covariances of dynamical processes.\\

Such a model implies to estimate new parameters $\kappa(\mathbf{s},t)$, $\mathbf{H}_{2 \times 2}(\mathbf{s},t)$ and $\mathbf{m}_{2 \times 1}(\mathbf{s},t)=\begin{bmatrix} \mathbf{m}^1 & \mathbf{m}^2 \end{bmatrix}^{\mathrm{T}}$, all varying across space and time along the data assimilation window. In addition, $\kappa$ needs to be continuous while $\mathbf{m}$ and $\mathbf{H}$ additionnaly requires to be continuously differentiable. Regarding the diffusion tensor, we draw from the spatial statistics literature, see e.g. \citep{fuglstad_2015a}, to introduce the scalars $\gamma(\mathbf{s},t)$, $\beta(\mathbf{s},t)$, $\bf{v}_1(\bf{s},t)$ and $\bf{v}_2(\mathbf{s},t)$ as a generic decomposition of $\mathbf{H}(\mathbf{s},t)$ through the equation:
\begin{align*} 
\mathbf{H}(\mathbf{s},t) = \begin{bmatrix} \mathbf{H}^{1,1} & \mathbf{H}^{1,2} \\ \mathbf{H}^{1,2} & \mathbf{H}^{2,2} \end{bmatrix}(\mathbf{s},t)  = \gamma(\mathbf{s},t) \mathbf{I}_2 + \beta(\mathbf{s},t) \mathbf{v} \mathbf{v}^{\mathrm{T}}
\end{align*}
with $\mathbf{v}^{\mathrm{T}} = \begin{bmatrix} \bf{v}_1(\mathbf{s}) & \bf{v}_2(\mathbf{s}) \end{bmatrix}$, 
which models the diffusion tensor as the sum of an isotropic and anisotropic effects, the latter being described by its amplitude and magnitude. This is a valid decomposition for any symmetric positive-definite $2 \times 2$ matrix. This leads to the SPDE hyperparametrization $\boldsymbol{\theta}$ of size $m \times L \times 8$ parameters ($\mathbf{H}^{1,2}=\mathbf{H}^{2,1}$): it grows linearly with the potentially high dimensional state space. 
In the end, the SPDE hyperparametrization states as:
\begin{align*}
\boldsymbol{\theta} = \begin{bmatrix} \boldsymbol{\kappa} & \mathbf{m} & \mathbf{H} & \boldsymbol{\tau} \end{bmatrix}^{\mathrm{T}}
\end{align*}

\begin{figure}[H]
\centering
    \subfloat[Isotropic model: $\frac{\partial{\mathbf{x}}}{\partial{t}}+(\kappa^2(\mathbf{s},t)-\Delta)^{\alpha/2} \mathbf{x}(\mathbf{s},t)=\tau \mathbf{z}(\mathbf{s},t))$]{
    \includegraphics[width=2cm,trim=4.5cm 4.5cm 8cm 0cm,clip]{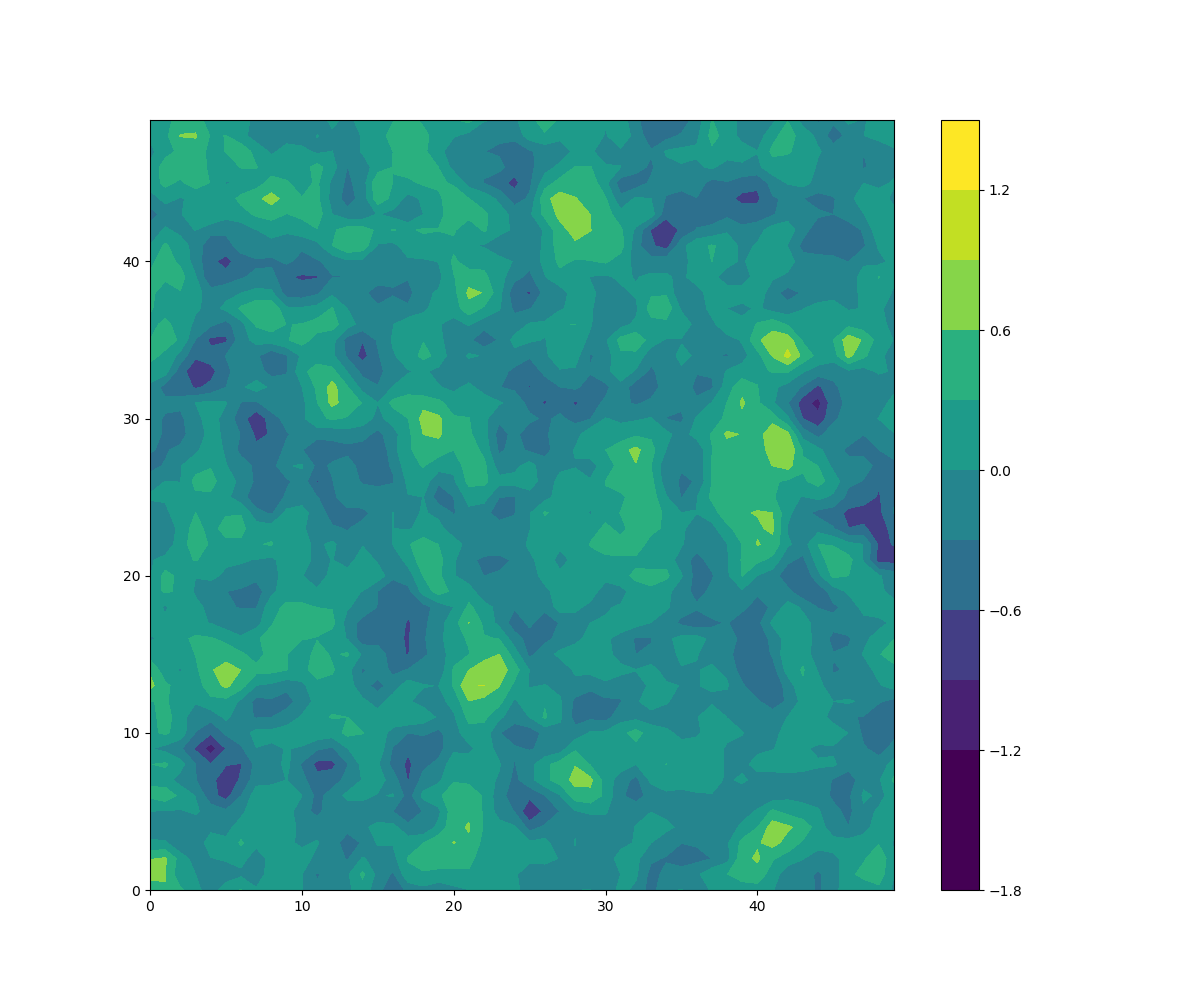}
    \includegraphics[width=2cm,trim=4.5cm 4.5cm 8cm 0cm,clip]{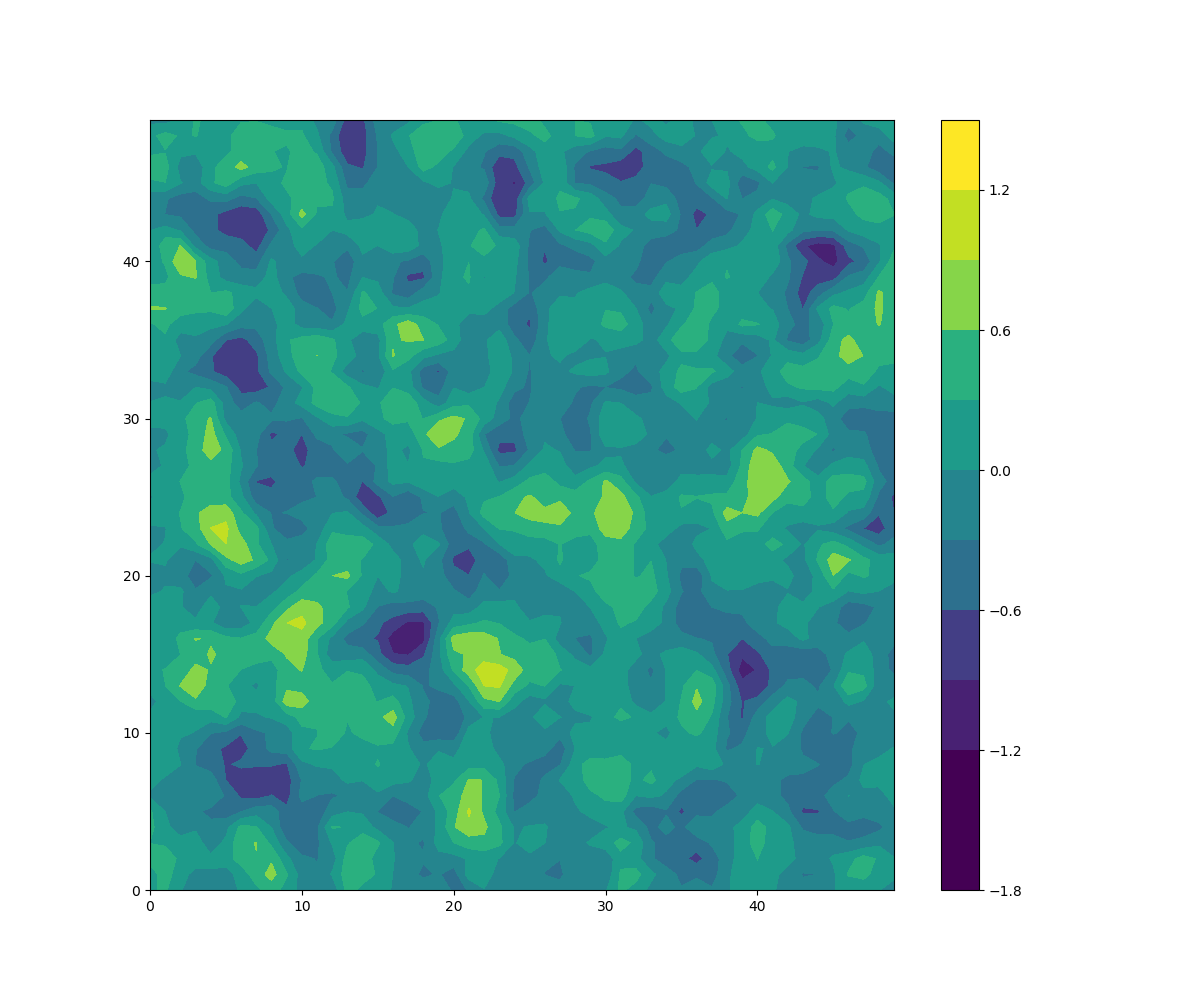}
    \includegraphics[width=1.9cm,trim=4.5cm 3.5cm 8cm 0cm,clip]{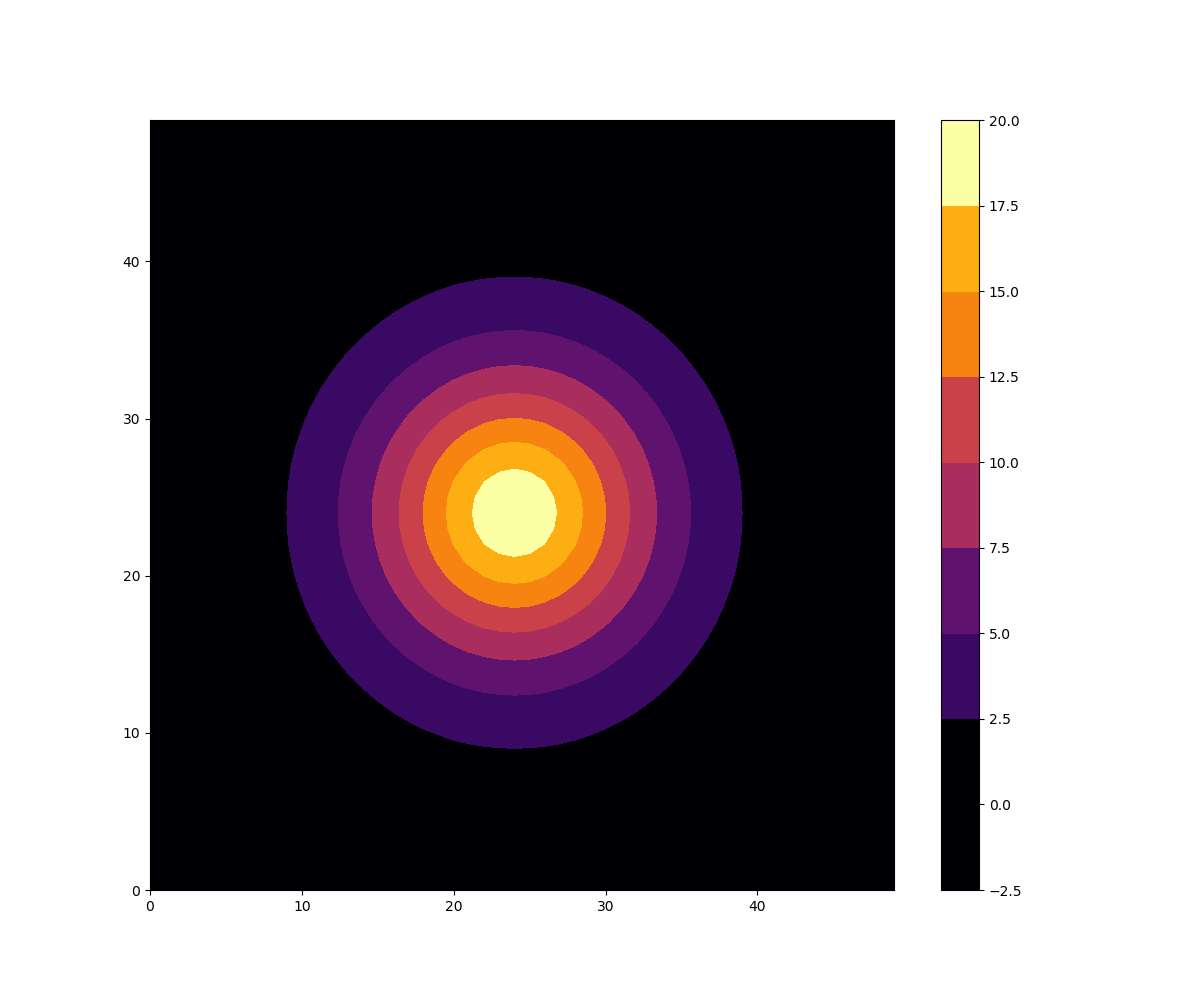}
    } \hfill
    \subfloat[Global anisotropy: $\frac{\partial{\mathbf{x}}}{\partial{t}}+\left\lbrace \kappa^2(\mathbf{s},t)- \nabla \cdot\mathbf{H}\nabla \right \rbrace^{\alpha/2} \mathbf{x}(\mathbf{s},t)=\tau \mathbf{z}(\mathbf{s},t)$]{
    \includegraphics[width=2cm,trim=4.5cm 4.5cm 8cm 0cm,clip]{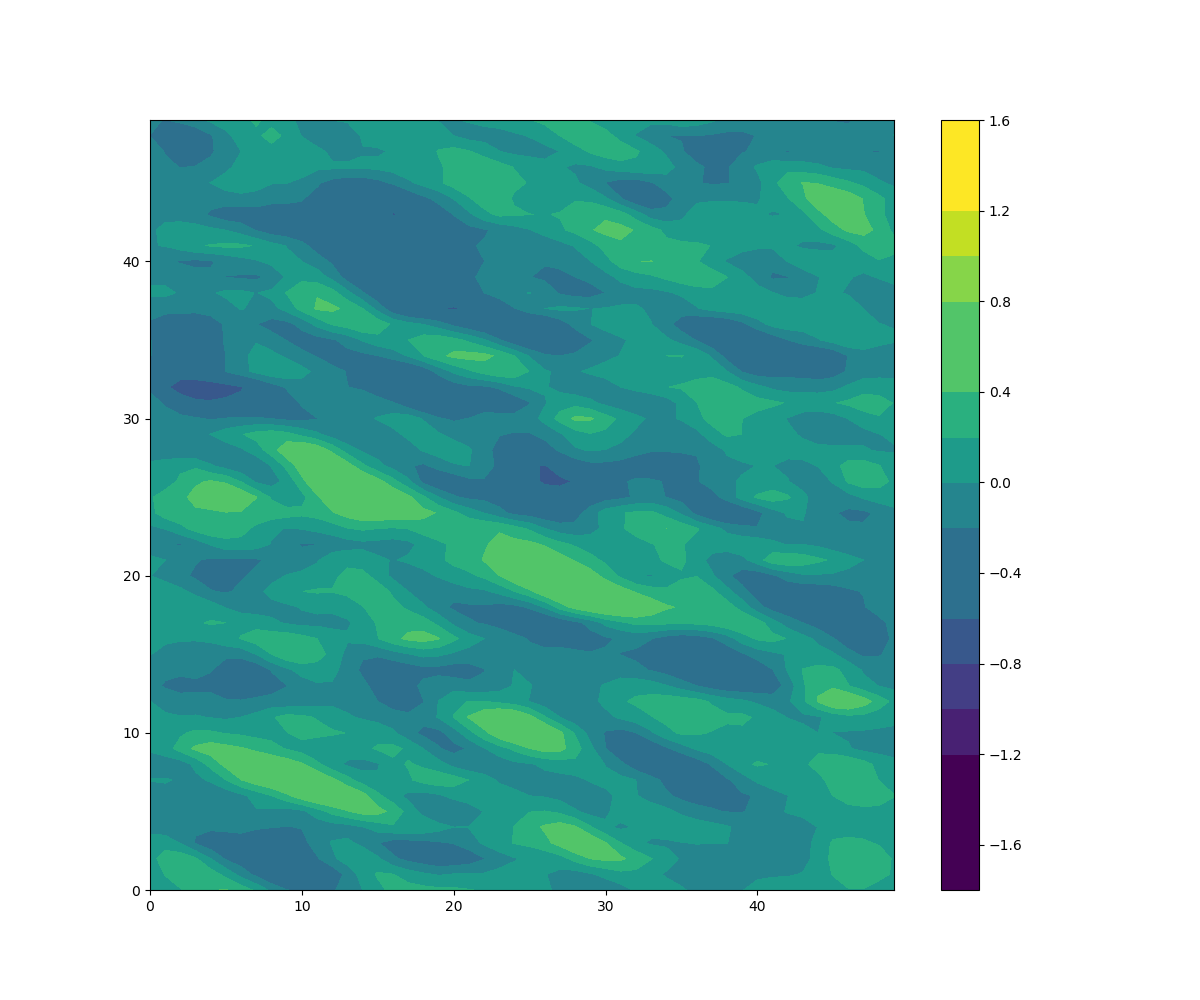}
    \includegraphics[width=2cm,trim=4.5cm 4.5cm 8cm 0cm,clip]{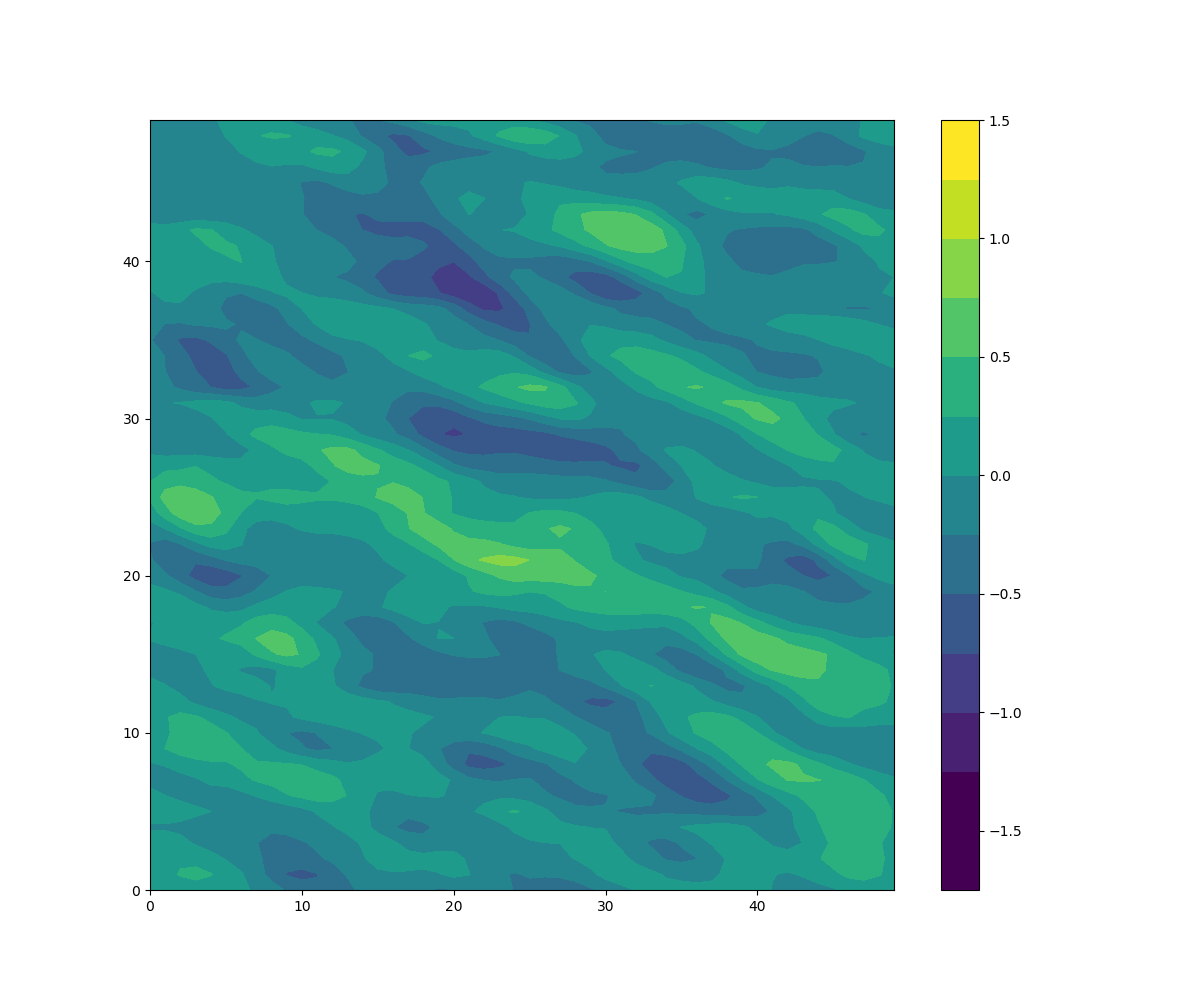}
    \includegraphics[width=1.9cm,trim=4.5cm 3.5cm 8cm 0cm,clip]{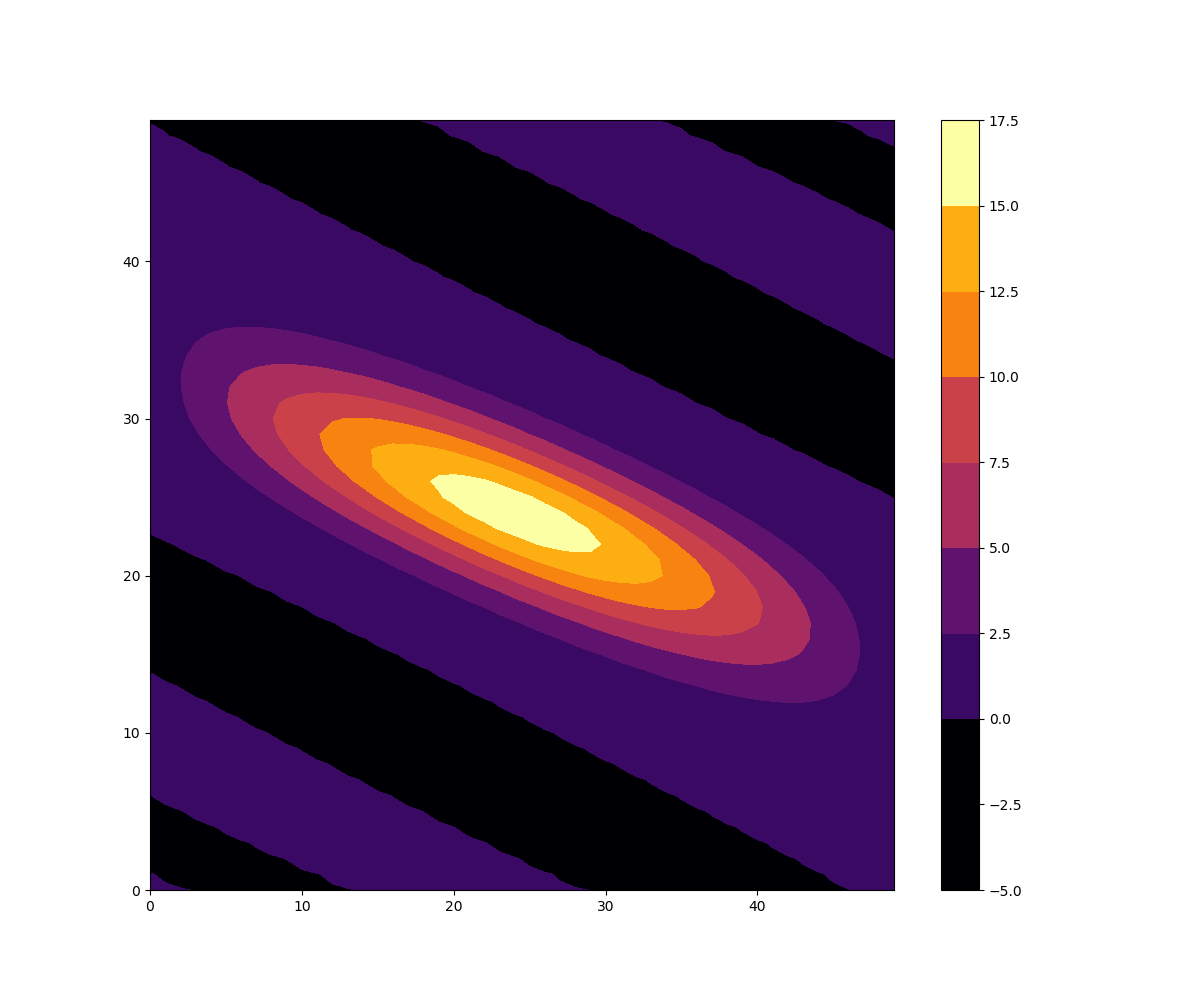}
    } \\
    \subfloat[Local anisotropy with diffusion: $\frac{\partial{\mathbf{x}}}{\partial{t}}+\left\lbrace \kappa^2(\mathbf{s},t) - \nabla \cdot\mathbf{H}(\mathbf{s})\nabla \right \rbrace^{\alpha/2} \mathbf{x}(\mathbf{s},t)=\tau \mathbf{z}(\mathbf{s},t)$]{
    \includegraphics[width=2cm,trim=4.5cm 4.5cm 8cm 0cm,clip]{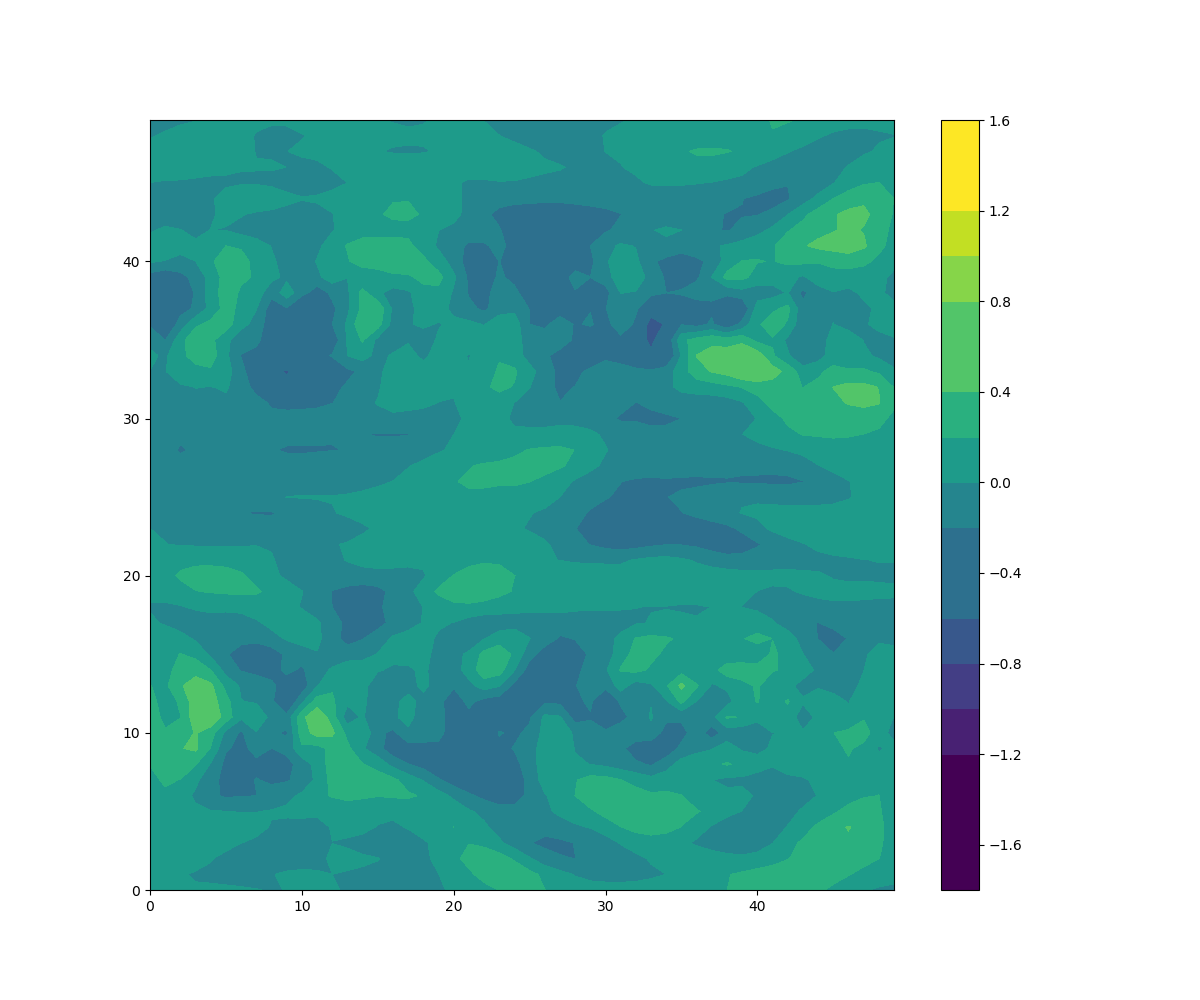}
    \includegraphics[width=2cm,trim=4.5cm 4.5cm 8cm 0cm,clip]{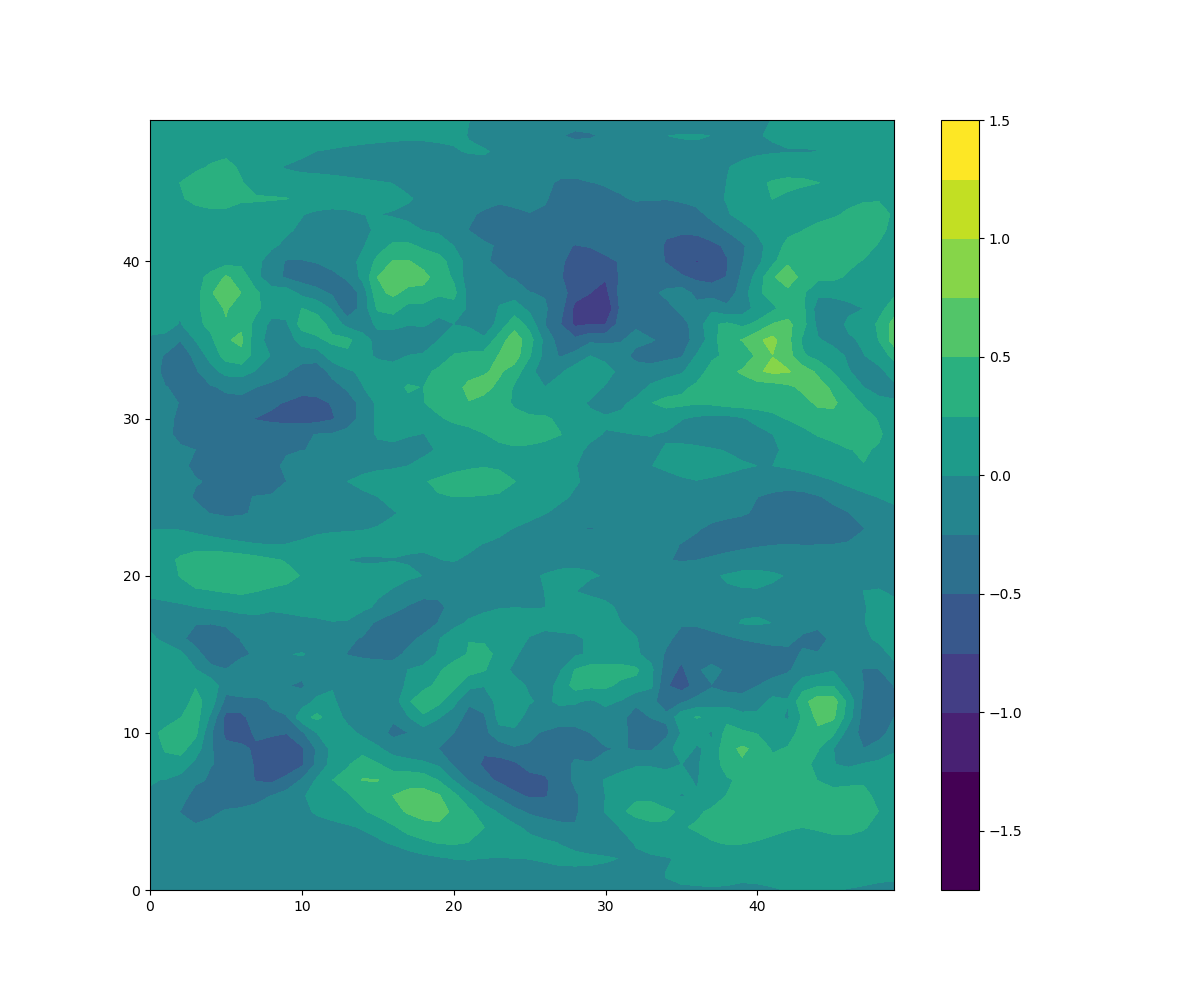}
    \includegraphics[width=1.9cm,trim=4.5cm 3.5cm 8cm 0cm,clip]{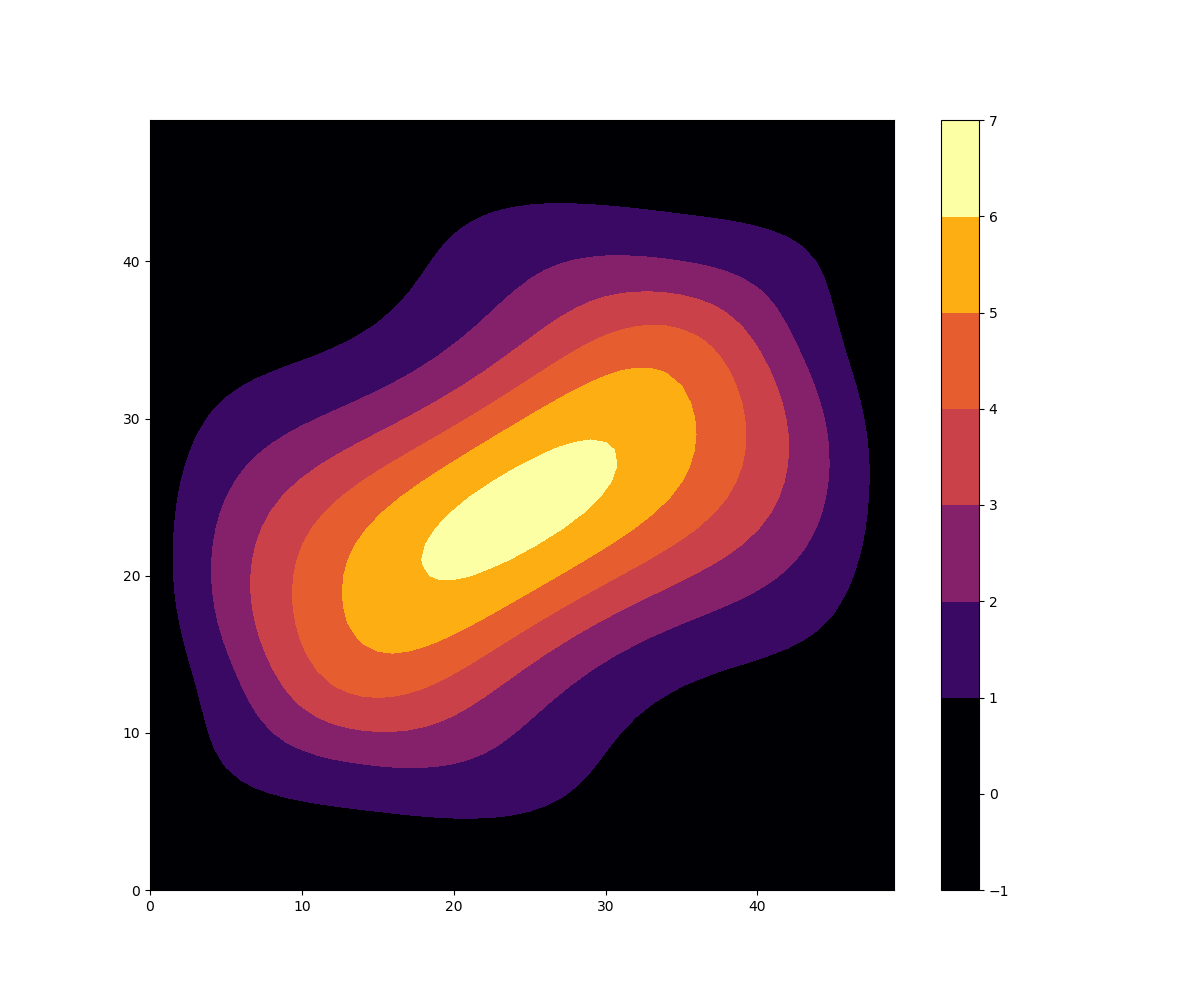}
    } \hfill
    \subfloat[Local anisotropy + Global advection: $\frac{\partial{\mathbf{x}}}{\partial{t}}+\left\lbrace \kappa^2(\mathbf{s},t) + \mathbf{m} \cdot \nabla - \nabla \cdot\mathbf{H}(\mathbf{s})\nabla \right \rbrace^{\alpha/2} \mathbf{x}(\mathbf{s},t)=\tau \mathbf{z}(\mathbf{s},t)$]{
    \includegraphics[width=2cm,trim=4.5cm 4.5cm 8cm 0cm,clip]{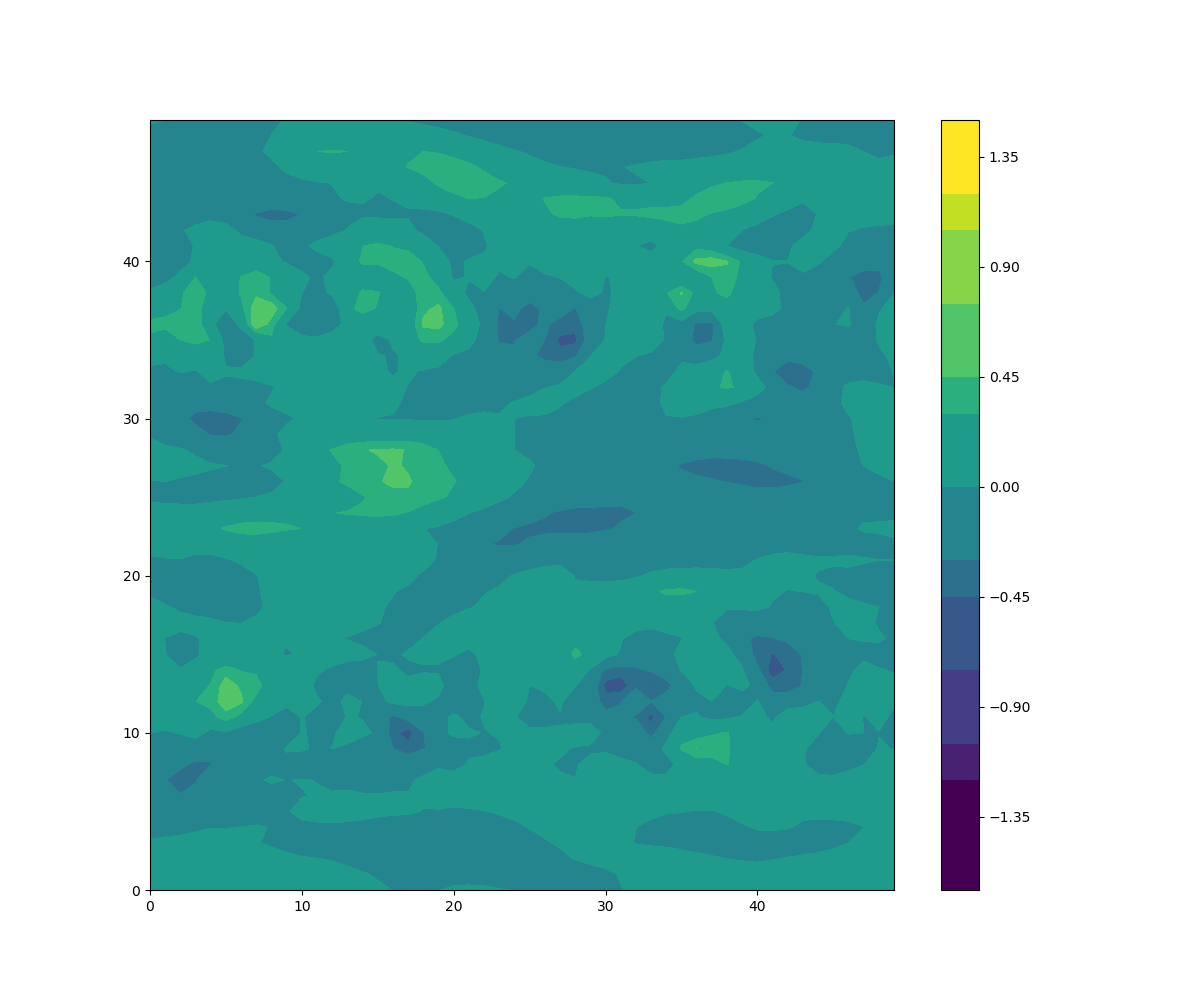}
    \includegraphics[width=2cm,trim=4.5cm 4.5cm 8cm 0cm,clip]{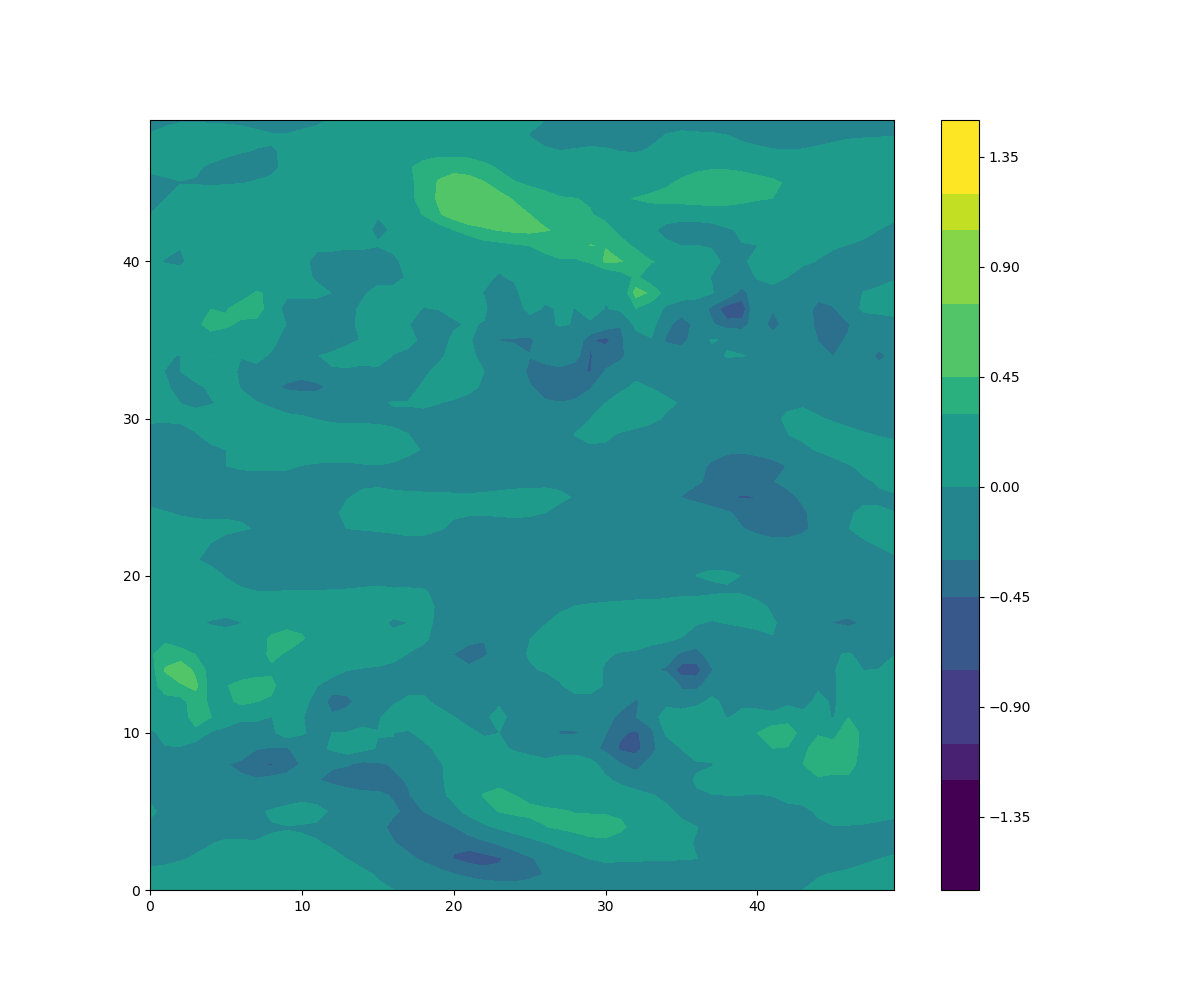}
    \includegraphics[width=1.9cm,trim=4.5cm 3.5cm 8cm 0cm,clip]{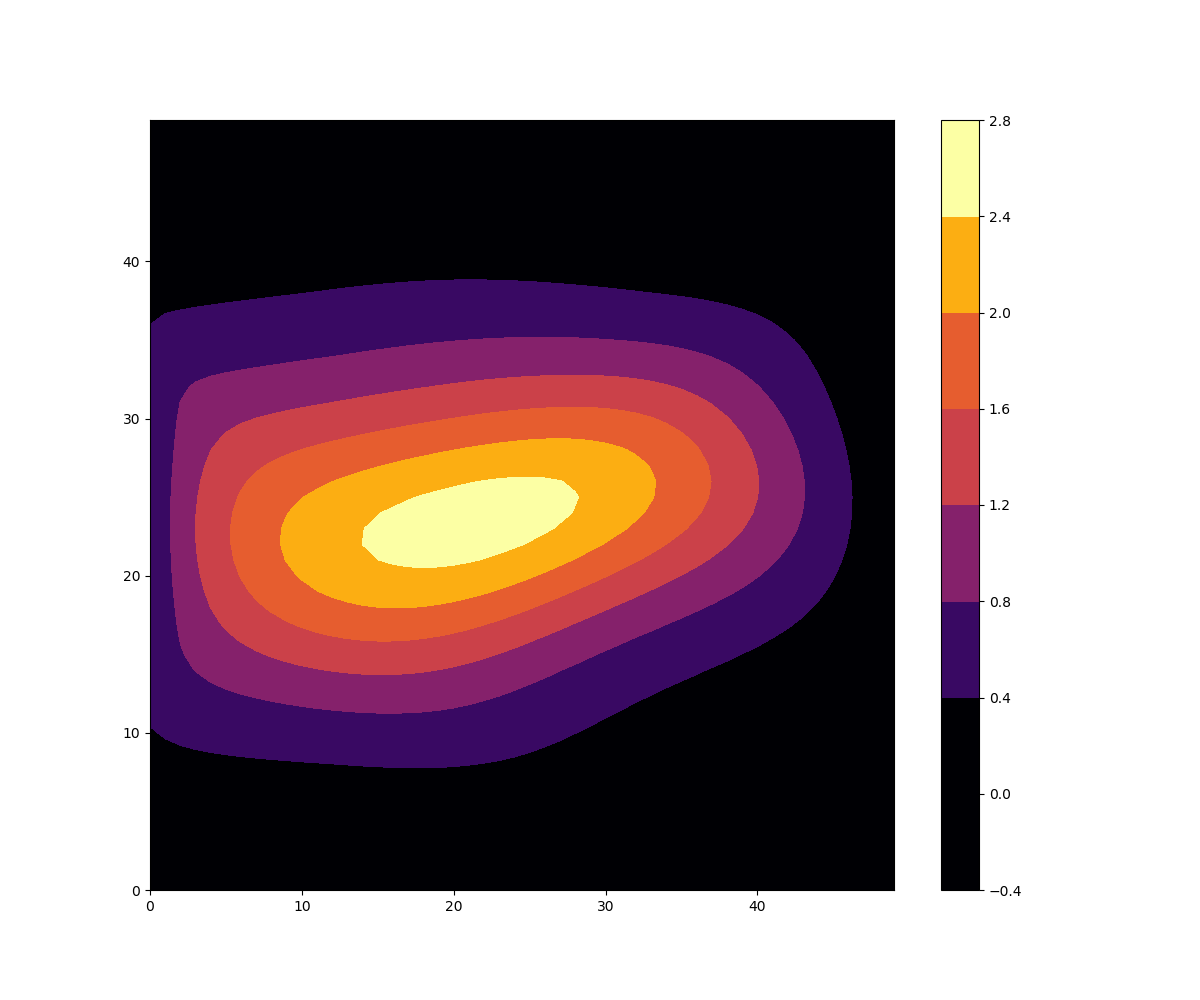}
    }
    \caption{For (a), (b), (c) and (d), one realization of the corresponding SPDE-driven GP at time $t=10$ (left panel), $t=20$ (middle panel) and covariance (right panel) with central point of domain $\mathcal{D}=[0,1]\times[0,1]$}
\label{ex_spde}
\end{figure}

Fig.\ref{ex_spde} provide some specific examples to show how the modifications in the fractional differential operator leads to more complex spatio-temporal anisotropies. In this four SPDE parametrizations, $\kappa=0.33$, $\tau=1$ and $\alpha=4$.

The space-time SPDE is discretized based on a numerical implicit Euler scheme:
\begin{eqnarray*}
\frac{\mathbf{x}_{t+dt}-\mathbf{x}_{t}}{dt}+\mathbf{B}_{t+dt}\mathbf{x}_{t+dt}=\frac{\tau}{\sqrt{dt}} \mathbf{z}_{t+dt}
\end{eqnarray*} 
where $\mathbf{x}_{t}$ and $\mathbf{B}_t$ respectively denote the state space and the finite difference discretization of the fractional differential operator $\frac{\partial{}}{\partial{t}}+\left\lbrace \boldsymbol{\kappa}^2(\mathbf{s},t) + \mathbf{m}(\mathbf{s},t) \cdot \nabla - \nabla \cdot\mathbf{H}(\mathbf{s},t)\nabla \right \rbrace^{\alpha/2}$  at time $t=0,\cdots,L$. The noise $\mathbf{z}_{t+dt}$ is white in space and $\mathbf{M}_{t+dt}=(\mathbf{I}+dt\mathbf{B}_{t+dt})^{-1}$ denotes the matrix operator that emulates the dynamical evolution of state $\x$ from time $t$ to $t+dt$. In this case, $\mathbf{T}_{t+dt}=\tau\sqrt{dt}\mathbf{M}_{t+dt}$ corresponds to the dynamical linear model regularized by the product of the noise variance with the square root of the SPDE time step. In case of a more general right-hand term with non-uniform regularization variance $\lbrace \boldsymbol{\tau}_t$, $t>0 \rbrace$, and $\lbrace \mathbf{z}_t$, $t>0 \rbrace$ are independent realizations of a colored noised driven by a spatial isotropic SPDE:
\begin{equation}
 \label{isotropic_spatial_SPDE}
 (\kappa_s^2 - \Delta)^{\alpha_s/2} Z_s(\mathbf{s}) = W(\mathbf{s})
\end{equation}
with $W(\mathbf{s})$ a white noise with unit variance, the spatial FDM together with time Euler discretization leads to:
\begin{equation}
\label{implicit_scheme2}
\mathbf{x}_{t+dt}  = \mathbf{M}_{t+dt}\mathbf{x}_{t} + \widetilde{\mathbf{T}}_{t+dt}\mathbf{z}_{t+dt} 
\end{equation}
where $\widetilde{\mathbf{T}}_{t+dt}=\sqrt{dt}\mathbf{M}_{t+dt} \boldsymbol{\tau}_t \mathbf{L}_{s}$  and $\mathbf{L}_{s}$ stands for the Cholesky decomposition of the discretized spatial precision matrix $\mathbf{Q}_s$ introduced of the stochastic process $Z_s$ introduced by Eq. (\ref{isotropic_spatial_SPDE}). 
In the case of advection-dominated SPDEs, we involve state-of-the-art upwind schemes (UFDM) for stabilization of the numerical system, by letting the advective transport term, which is the dominating term, collect its information in the flow direction, i.e., upstream or upwind of the point in question. All the calculation details are given in Appendix \ref{App:AppendixA}. \\

In a compact formulation, using centered finite differences on the diffusion term, and by denoting $\mathbf{a}^{1,t,+}_{i,j} = \max(\mathbf{m}^{1,t}_{i,j},0)$, $\mathbf{a}^{1,t,-}_{i,j} = \min(\mathbf{m}^{1,t}_{i,j},0)$, $\mathbf{a}^{2,t,+}_{i,j} = \max(\mathbf{m}^{2,t}_{i,j},0)$, $\mathbf{a}^{2,t,-}_{i,j} = \min(\mathbf{m}^{2,t}_{i,j},0)$, the resulting UFDM scheme is:
\begin{align*}
\mathbf{x}^{t+1}_{i,j} = \mathbf{x}^{t}_{i,j} + dt &\Big[\kappa^t_{i,j}\mathbf{x}^t_{i,j} + 
\left( \mathbf{a}^{1,t,+}_{i,j}\mathbf{m}^{1,t,-}_{i,j} + \mathbf{a}^{1,t,-}_{i,j}\mathbf{m}^{1,t,+}_{i,j}\right) + 
\left( \mathbf{a}^{2,t,+}_{i,j}\mathbf{m}^{2,t,-}_{i,j} + \mathbf{a}^{2,t,-}_{i,j}\mathbf{m}^{2,t,+}_{i,j}\right) \\
&+ \mathbf{H}^{1,1,t}_{i,j} \frac{\mathbf{x}^{t}_{i+1,j} -2\mathbf{x}^{t}_{i,j} + \mathbf{x}^{t}_{i-1,j}}{dx^2} + \mathbf{H}^{2,2,t}_{i,j} \frac{\mathbf{x}^{t}_{i,j+1} -2\mathbf{x}^{t}_{i,j} + \mathbf{x}^{t}_{i,j-1}}{dy^2} \\
& + \mathbf{H}^{1,2,t}_{i,j} \frac{\mathbf{x}^{t}_{i+1,j+1} -\mathbf{x}^{t}_{i+1,j-1} - \mathbf{x}^{t}_{i-1,j+1} + \mathbf{x}^{t}_{i-1,j-1}}{2dxdy} + \tau^t_{i,j} \mathbf{z}^{t+1}_{i,j} \Big]
\end{align*}

Starting from this numerical scheme, the modified 4DVarNet scheme requires the precision matrix $\mathbf{Q}^b_{\boldsymbol{\theta}}$ of the state sequence $\lbrace \x_0,\cdots,\x_{Ldt} \rbrace$. Here, $\x_0 \sim \mathcal{N}(\mathbf{0},\mathbf{P_0})$ denotes the initial state and $\mathbf{Q}_0=\mathbf{P_0}^{-1}$ is always taken as the precision matrix obtained after a given stabilization run, i.e. the evolution of the dynamical system over $N$ timesteps using as stationary parameters the initial parametrization $\boldsymbol{\theta}_0$ of the SPDE at time $t=0$, then we can rewrite :
\[ \lbrace \x_0,\cdots,\x_{Ldt} \rbrace = \mathbf{M}_G \left[\begin{array}{c}\mathbf{x}_0\\ \mathbf{z}\end{array}\right]\]
with $\mathbf{z}=[\mathbf{z}_1,\dots,\mathbf{z}_t]^{\mathrm{T}}$ and 
\[ \mathbf{M}_G = \left[\begin{array}{ccccccc} 
\mathbf{I} & 0  & 0 & 0 & 0 &\dots& 0 \\
\mathbf{M}_1 & \mathbf{T}_1 & 0 &0  & 0 & \dots &0 \\
\mathbf{M}_2\mathbf{M}_1 & \mathbf{M}_2\mathbf{T}_1 & \mathbf{T}_2 & 0& 0 &\dots& 0\\
\mathbf{M}_3\mathbf{M}_2\mathbf{M}_1 & \mathbf{M}_3\mathbf{M}_2\mathbf{T}_1 & \mathbf{M}_3\mathbf{T}_2 & \mathbf{T}_3 & 0 &\dots & 0\\
\vdots &\ddots &\ddots &\ddots & \ddots & \ddots & 0 \\
\vdots &\ddots &\ddots &\ddots & \ddots & \ddots & 0 \\
\vdots &\ddots &\ddots &\ddots & \ddots & \ddots & \mathbf{T}_L \\
\end{array}\right]\]
With the additional notation $\mathbf{S}_k=\mathbf{T}_k\mathbf{T}_k^{\mathrm{T}}$, see Eq. (\ref{implicit_scheme2}), the precision matrix $\mathbf{Q}^b$ writes, see Appendix \ref{App:AppendixB} for all the details:
\begin{align}
\hspace{-1.5cm}
\label{global_prec_mat_2}
\mathbf{Q}^b_{\boldsymbol{\theta}}  &= \frac{1}{dt} 
\begin{bsmallmatrix} 
\mathbf{P}_0^{-1}+\widetilde{\mathbf{Q}}_{s,1}  & -\widetilde{\mathbf{Q}}_{s,1}\mathbf{M}_1^{-1} & 0 & 0 & 0 &\dots& 0 \\
-\left(\mathbf{M}_1^{\mathrm{T}}\right)^{-1}\widetilde{\mathbf{Q}}_{s,1} & \mathbf{M}^{\mathrm{T}}_1\widetilde{\mathbf{Q}}_{s,1}\mathbf{M}_1 +\widetilde{\mathbf{Q}}_{s,2} & -\widetilde{\mathbf{Q}}_{s,2}\mathbf{M}_2^{-1} &0  & 0 & \dots &0 \\
0 & -\left(\mathbf{M}_2^{\mathrm{T}}\right)^{-1}\widetilde{\mathbf{Q}}_{s,2} & \mathbf{M}^{\mathrm{T}}_2\widetilde{\mathbf{Q}}_{s,2}\mathbf{M}_2 + \widetilde{\mathbf{Q}}_{s,3} & -\widetilde{\mathbf{Q}}_{s,3}\mathbf{M}_3^{-1} & 0 & \dots & 0\\
0 &\ddots &\ddots &\ddots & \ddots & \ddots & 0 \\
0 &\ddots &\ddots &\ddots & \ddots & \ddots & 0 \\
\vdots &\ddots &\ddots &\ddots & -\left(\mathbf{M}_{L-1}^{\mathrm{T}}\right)^{-1}\widetilde{\mathbf{Q}}_{s,L-1} & \mathbf{M}^{\mathrm{T}}_{L}\widetilde{\mathbf{Q}}_{s,L-1} \mathbf{M}_{L-1} + \widetilde{\mathbf{Q}}_{s,L} & -\widetilde{\mathbf{Q}}_{s,L}\mathbf{M}_{L}^{-1} \\
0 &\ddots &\ddots &\ddots & 0 & -\left(\mathbf{M}_{L}^{\mathrm{T}}\right)^{-1}\widetilde{\mathbf{Q}}_{s,L} & \mathbf{M}^{\mathrm{T}}_{L}\widetilde{\mathbf{Q}}_{s,L}\mathbf{M}_{L} \\
\end{bsmallmatrix}
\end{align}
where $\widetilde{\mathbf{Q}}_{s,t}$ is the precision matrix of the colored noise weighted by the non-uniform regularization variance $\boldsymbol{\tau}_t$. As clearly visible, the sparsity of $\mathbf{Q}^b_{\boldsymbol{\theta}}$ is high, which is key in traditional SPDE-based GP inference, but also in our approach, see Section \ref{neuralsolver}.

\subsection{Neural solver with augmented state}
\label{neuralsolver}

\begin{figure}[H]
  \centering
  \includegraphics[width=16cm]{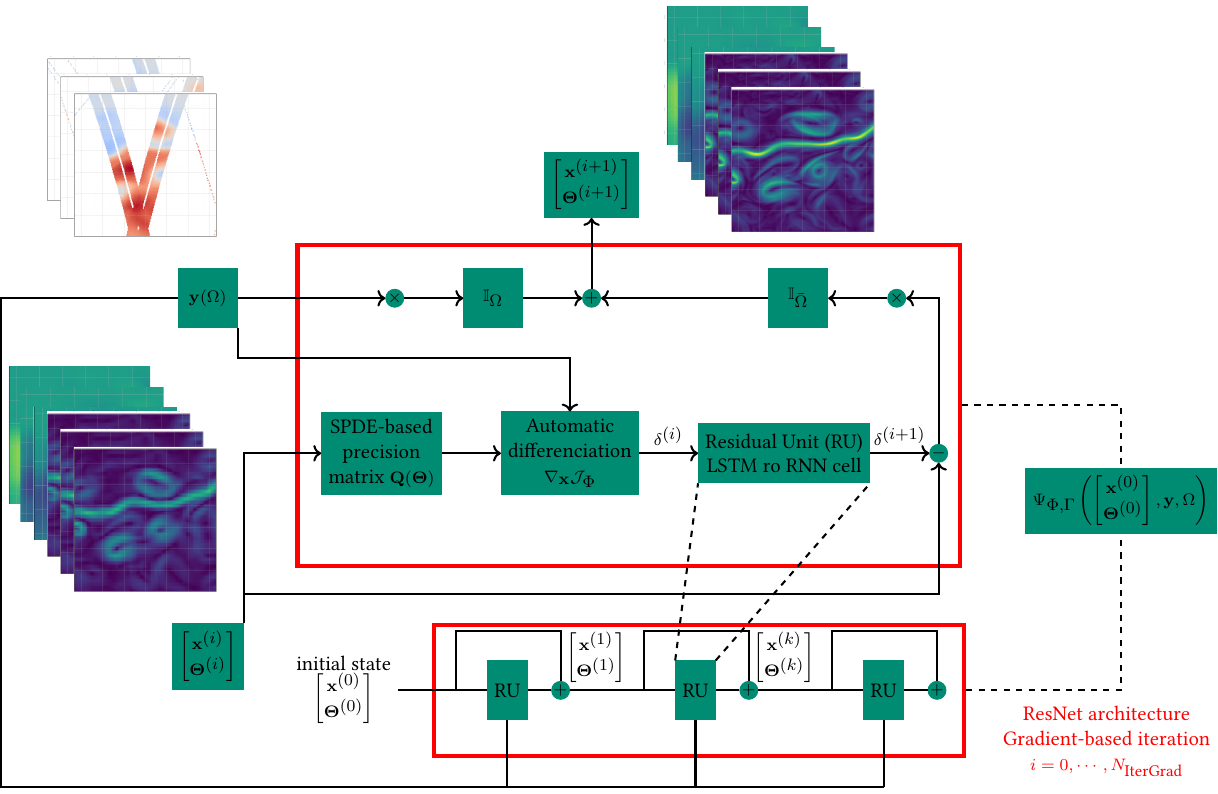}
   \caption{Sketch of the gradient-based neural emulation at the core of our neural architecture. Based on the single observations and the initialization of the augmented state, a set of predefined LSTM residual units are involved to reach convergence of the optimal state. $\mathbf{I}_\Omega$ acts as a masking operator for any spatio-temporal location not in $\Omega$.}
   \label{sketch_GB}
\end{figure}

Overall, let denote by $\Psi_{\boldsymbol{\theta},\Gamma}(\tilde{\mathbf{x}}^{(0)},\mathbf{y},\Omega )$ the output of the end-to-end learning scheme given the SPDE-based dynamical model with parameters $\boldsymbol{\theta}$ and the neural residual architecture for the solver $\Gamma$, see Fig. \ref{sketch_GB} and Algorithm \ref{algoSPDE_4DVarNet}, the initialization $\tilde{\mathbf{x}}^{(0)}$ of augmented state $\tilde{\mathbf{x}}$ and the observations $\mathbf{y}$ on domain $\Omega$.
\vspace{-.5cm}

\begin{center}
\centering
\begin{algorithm}[H]
 \label{algoSPDE_4DVarNet}
 \SetAlgoLined
 \KwData{\\
 \tabto{.3cm} $\x \in \mathbb{R}^{T \times m}=\lbrace{ \x_k \rbrace},\ k=1,\cdots,T$ \\
 \tabto{.3cm} $\y_{\Omega}= \lbrace{ \y_{k,\Omega_k} \rbrace},\ k=1,\cdots,T$: observations on domains $\Omega_k \subset \mathcal{D}$ \\
 \tabto{.3cm} $N_I$: number of iterations \\
 \tabto{.3cm} $\eta$: gradient step \\
 }
 \KwInit{\\
   \tabto{.3cm} $\tilde{\mathbf{x}}^{\star,(0)}$
 }
 \KwProcedures{\\
  \tabto{.3cm} Train\_$\Psi_{\boldsymbol{\theta},\Gamma}$: end-to-end learning procedure with: \\
  \tabto{.6cm} $\boldsymbol{\theta}$: parameter of the SPDE-based prior operator; \\
  \tabto{.6cm} $GradLSTM$: residual NN-based representation of $\nabla_\x \mathcal{J}(\x)$ \\
  \tabto{.6cm} $\Gamma$: iterative gradient-based update operator:\\
  \tabto{.8cm}  $i=0$\\
  \tabto{.8cm}  \While{$i<N_I$}{    
     $\mathbf{Q}^b_{\boldsymbol{\theta^{\star,(i)}}} = \mathbf{P}_{G}^{-1}(\boldsymbol{\theta^{\star,(i)}}) = {\mathbf{M}_G^{-1}}(\boldsymbol{\theta^{\star,(i)}})      ^{\mathrm{T}}\left[\begin{array}{llll}
\mathbf{P}_0^{-1} & 0 & \dots & 0\\
0  & \mathbf{I} &\dots & 0\\
\vdots & \ddots & \ddots & \vdots\\
0 & 0 & \dots & \mathbf{I}
\end{array}\right] \mathbf{M}_G^{-1}(\boldsymbol{\theta^{\star,(i)}})$\\
     $\tilde{\mathbf{x}}^{(i+1)} \leftarrow \tilde{\mathbf{x}}^{(i)}-\eta \times GradLSTM(\tilde{\mathbf{x}}^{(i)})$ \\
     $N_I \nearrow$\ ;\ $\eta \searrow$ ; \ $i \leftarrow i+1$ \\
   }
 }
 \For{$i \in 0,\cdots,n_{epochs}$}{
   $\omega_{\Psi_{\boldsymbol{\theta},\Gamma}}^{(i+1)} \leftarrow \omega_{\Psi_{\boldsymbol{\theta},\Gamma}}^{(i)}-lr \times \nabla \mathcal{L}(\mathbf{x},\mathbf{x}^{\star,(i)},\boldsymbol{\theta^{(i)}})$\\
 }
 \KwResult{$\tilde{\mathbf{x}}^\star \leftarrow \Psi_{\boldsymbol{\theta},\Gamma}(\tilde{\mathbf{x}}^{(0)},\mathbf{y},\Omega )$ } 
 \caption{Variational scheme with SPDE-based GP prior and implicit neural solver}
\end{algorithm}

\end{center}

Then, the joint learning for the weights $\omega_{\boldsymbol{\theta},\Gamma}$ of the neural scheme given the SPDE formulation that is chosen (isotropic or not, non-stationary or not, etc.) and the architecture of operator $\Gamma$ is stated as the minimization of the mixed loss function $\mathcal{L}(\mathbf{x},\mathbf{x}^{\star},\boldsymbol{\theta}^{\star})$, fully explained in Section \ref{learning_scheme}:
\begin{equation}
\label{eq: E2E loss}
   \omega^\star_{\Psi_{\boldsymbol{\theta},\Gamma}} = \arg \min_{\boldsymbol{\theta},\Gamma} \Big[ \mathcal{L}(\mathbf{x},\mathbf{x}^{\star},\boldsymbol{\theta}^{\star}) \Big] \mbox{  s.t.  } 
   \tilde{\mathbf{x}}^\star = \Psi_{\boldsymbol{\theta},\Gamma}  (\tilde{\mathbf{x}}^{(0)},\mathbf{y},\Omega)
\end{equation} 
It is important to note that as most of traditional convolutional neural networks, the layers acts as stationary filters so that all the weights are shared across all locations. What changes across the locations is the SPDE parameters $\theta$ estimated as latent variables in the augmented state formalism, which makes the SPDE, hence the prior, not stationary. 

\paragraph{Initialization of the augmented state.} As initial condition $\x^{(0)}$, the non-observed parts of the domain are filled with the global mean of the training dataset (0 when normalized). It could be something else, like pure noise as in diffusion models \citep{Ho_2020}, but we found that using noise as initial condition requires more iterations to reach the scheme convergence.. Because there is obviously no observation of the SPDE parameters, we might guide the training process at the first iteration of the solver. This proved to be particularly helpful when dealing with realistic geophysical datasets for which the advection diffusion scheme is meaningful. In that case, see for instance Application 2 on real SSH datasets in Section \ref{results}, we used first and second-order derivatives of the initial state $\x^{(0)}$ as initial parametrizations for the advection and diffusion process:
\begin{align*}
\mathbf{m}^{(0)} = \begin{pmatrix} \frac{\partial{\x^{(0)}}}{\partial{x}} \\ \frac{\partial{\x^{(0)}}}{\partial{y}}   \end{pmatrix}, \ \mathrm{and} \  \mathbf{v}^{(0)} = \begin{pmatrix} \frac{\partial^2{\x^{(0)}}}{\partial{x^2}} \\ \frac{\partial^2{\x^{(0)}}}{\partial{y^2}}   \end{pmatrix} \\
\end{align*}
$\boldsymbol{\kappa}^{(0)}$ and $\boldsymbol{\tau}^{(0)}$ are both using absolute values of the normalized gradient norms, while $\beta^{(0)}$ and $\gamma^{(0)}$ are resp. set to 1 and 0. Because $\kappa$, $\tau$ and $\gamma$ are both strictly positive, see Section \ref{spde_param}, we use ReLu activation function on these three parameters to ensure their consistency.

\paragraph{Computational aspects.} The sparse formulation of the precision matrix $\mathbf{Q}^b_{\boldsymbol{\theta}}$ is key in the memory-saving component of the algorithm because the latter relies on a set of $N_{I}$ gradient-based iterations, meaning that for a single interpolation task, the precision matrix $\mathbf{Q}^b_{\boldsymbol{\theta}}$ is stored  $N_{I}$ times along the computational graph with updated values of the SPDE parameters $\boldsymbol{\theta}^{(i)}$.

\subsection{UQ scheme}
\label{uq_scheme}
\begin{figure}[H]
\centering
\includegraphics[width=10cm]{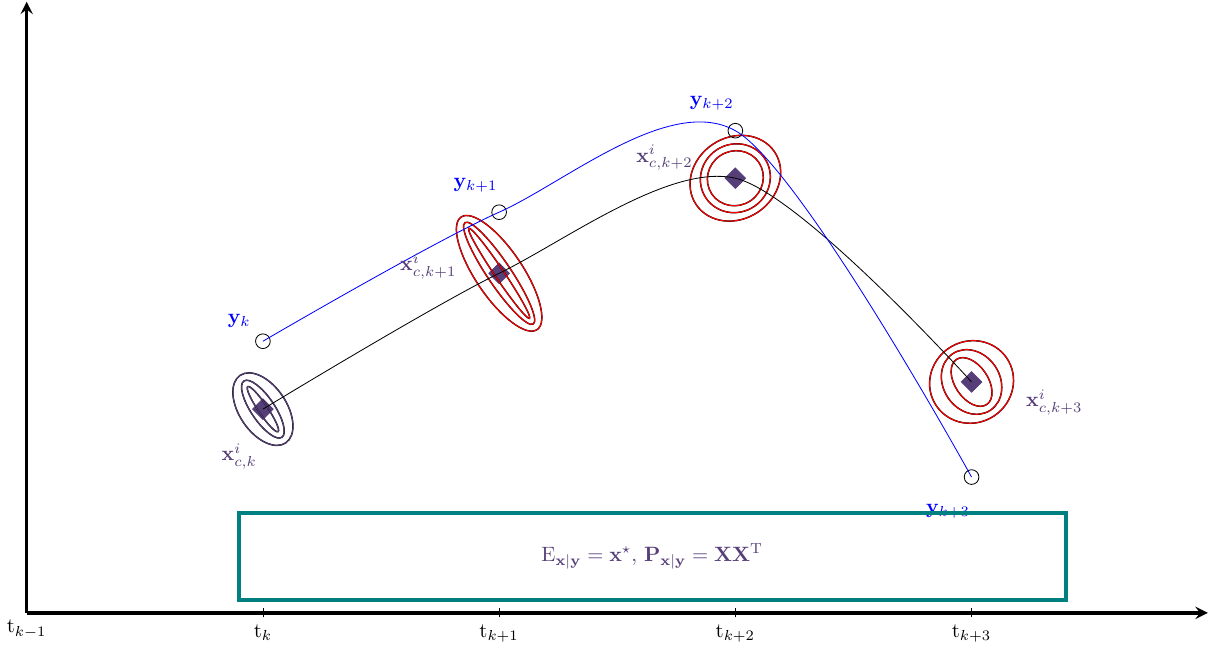}
\caption{Ensemble-based neural variational scheme with SPDE-based GP prior: the assimilation is not sequential. The inversion scheme embeds the global precision matrix of the state sequence $\lbrace \x_0,\cdots,\x_{Ldt} \rbrace$ in the inner variational cost to minimize. Ensemble members are generated from the Gaussian prior surrogate SPDE model, then conditioned by the neural solver so that the posterior pdf is no longer Gaussian. The posterior pdf is empirically ensemble-based computed: $\mathbf{P}_{\mathbf{x}|\mathbf{y}} = \mathbf{X}\mathbf{X}^{\mathrm{T}}$ with $\mathbf{X}=(1/\sqrt{N-1})\begin{bmatrix} \mathbf{x}_i^{\star} - \mathbf{x}^\star \end{bmatrix}$}
\label{EnOI_spde} 
\end{figure}

Because the PDE is stochastic, it provides an easy way to generate a set of $N$ gaussian prior simulations $\mathbf{x}_i$, $i=1,\cdots,N$:
\begin{align*}
\mathbf{x}_i = \x^b + \mathbf{L}^b_{\boldsymbol{\theta}}\mathbf{z}_i
\end{align*}
where $\mathbf{L}^b_{\boldsymbol{\theta}}$ stands for the Cholesky decomposition of $\mathbf{Q}^b_{\boldsymbol{\theta}}$ and $\mathbf{z}_i$ is a white noise. From a geostatistical point of view, this can be seen as SPDE-based spatio-temporal non conditional simulations of state $\x$, meaning that we produce surrogate simulations sharing the same physical properties than the true states. These simulations are then conditioned by the neural solver given the observations available $\mathbf{y}$, see Fig. \ref{EnOI_spde}. To do so, we draw from traditional geostatistics to realize SPDE-based spatio-temporal non conditional simulations with a kriging-based conditioning \citep{wackernagel_2003}. Except that we replace the kriging algorithm by our neural approach, which has to be seen as a generic interpolation tool here:
\begin{align}
  \label{simu_cond}
 \mathbf{x}_i^{\star} (\mathbf{s},t) = \mathbf{x}^\star(\mathbf{s},t) + (\mathbf{x}_i (\mathbf{s},t) - \widehat{\mathbf{x}}_i(\mathbf{s},t))
\end{align}
where $\mathbf{x}^\star$ denotes the neural-based interpolation, $\mathbf{x}_i$ is one SPDE  non-conditional simulation of the process $\mathbf{x}$ based on the parameters $\boldsymbol{\theta}^\star$ and $\widehat{\mathbf{x}}_i$ is the neural reconstruction of this non-conditional simulation, using as pseudo-observations a subsampling of $\mathbf{x}_i$ based on the actual data locations. Because $\mathrm{E}[\mathbf{x}_{i} - \widehat{\mathbf{x}}_i]=0$, the resulting simulation is well conditioned by the observations at data locations. \\

Running an ensemble of $N$ conditional simulations gives an approximation of the probability distribution function $p_{\mathbf{x}|\mathbf{y}}$ of state $\mathbf{x}^\star=\mathbf{x}|\mathbf{y}=\lbrace \mathbf{x}_0|\mathbf{y}, \cdots, \mathbf{x}_L|\mathbf{y}\rbrace$. The ensemble mean $\overline{\mathbf{x}^{\star}_i}$ will be $\mathbf{x}^\star$ in the limits of $N \rightarrow +\infty$:
\begin{align*}
\frac{1}{N} \sum_i \mathbf{x}_i^{\star} (\mathbf{s},t) \xrightarrow[N \to +\infty]{} \mathbf{x}^\star(\mathbf{s},t)
\end{align*}
Such an approach has already been successfully tested in \citet{beauchamp_2023c} when using analog operator strategy \citep{tandeo_combining_2015} to draw non-conditional simulation in the prior distribution. 

Let note that in this formulation, the idea is to run SPDE-based conditional simulation of the prior Mat\'{e}rn field $X(\mathbf{s},t)$. One ensemble member is obtained by running one space-time simulation, and two space-time neural reconstructions along the data assimilation window, then combined through Eq. (\ref{simu_cond}). As a consequence, there is no sequential assimilation. Though, in the idea, such a conditioning is exactly similar to the one produced by EnKF simulations: in geostatistical terms we can interpret the forecast step of the EnKF as being unconditional simulations at time $t-1$ generating $N$ realizations of a non-stationary random function (SPDE-based here) for time $t$. Both prior mean and covariance matrix are then computed directly on this set of unconditional realizations, before the analysis step, i.e. their conditiong with the observations. The key advantage of the EnKF is its flow-dependency that propagates the uncertainties at each time step with the evolution model, while the classic EnOI method generally used by geoscientists, see e.g. \citet{asch_data_2016,counillon_2008} replaces the flow-dependent EnKF error covariance matrix by a stationary matrix calculated from an historical ensemble. This is an important difference with our approach: while the neural architecture can be seen as a way to learn neural ensemble-based optimal interpolation models and solvers, our GP prior encoded by its SPDE precision matrix, built according to the FDM scheme with varying parameters over space and time, still allows for flow dependency, based on the learning of the SPDE parametrization given the input observations.\\ 

\begin{figure}[H]
\centering
\tikz[remember picture]{\node(1BL){\includegraphics[width=10cm]{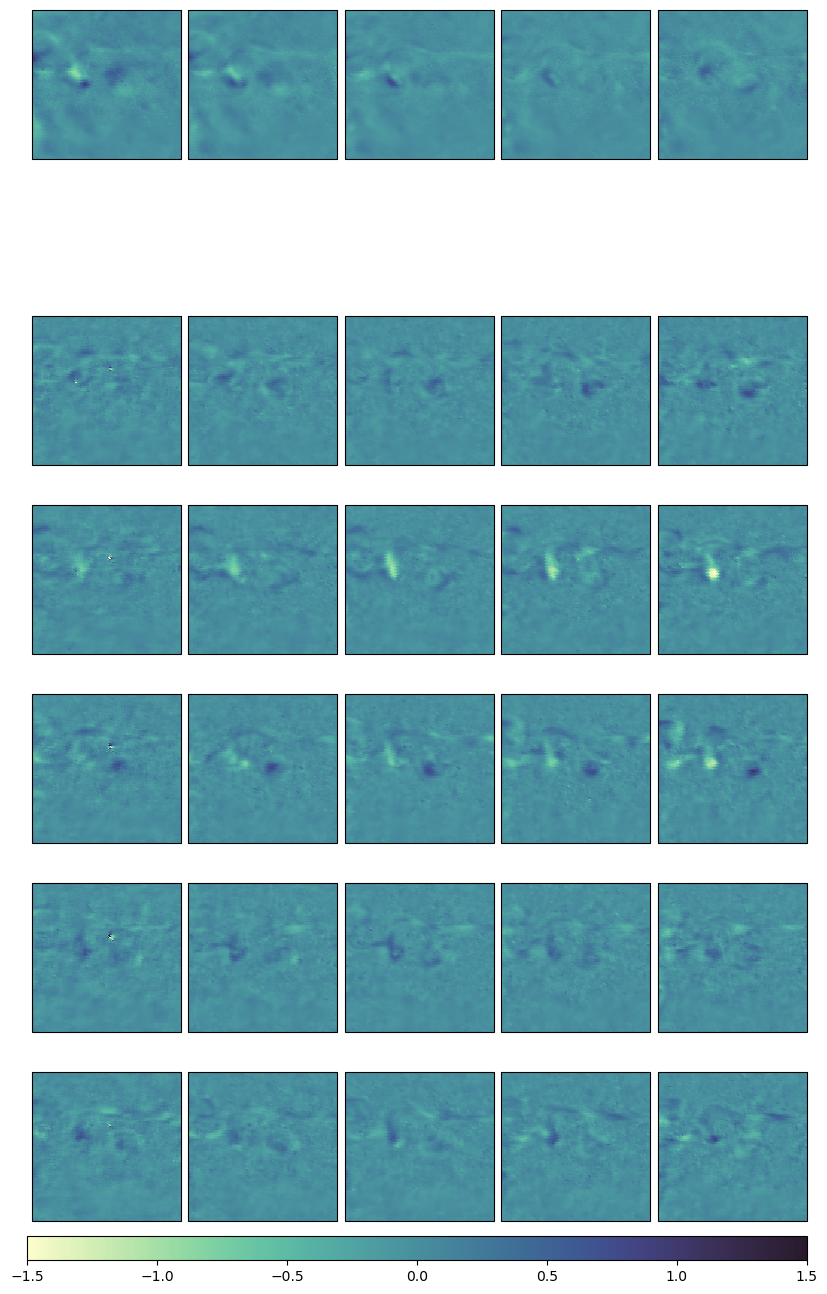}};}%
\tikz[overlay,remember picture]{
    \node[inner sep=0pt] at (1BL.center) {%
        \includegraphics[width=10cm]{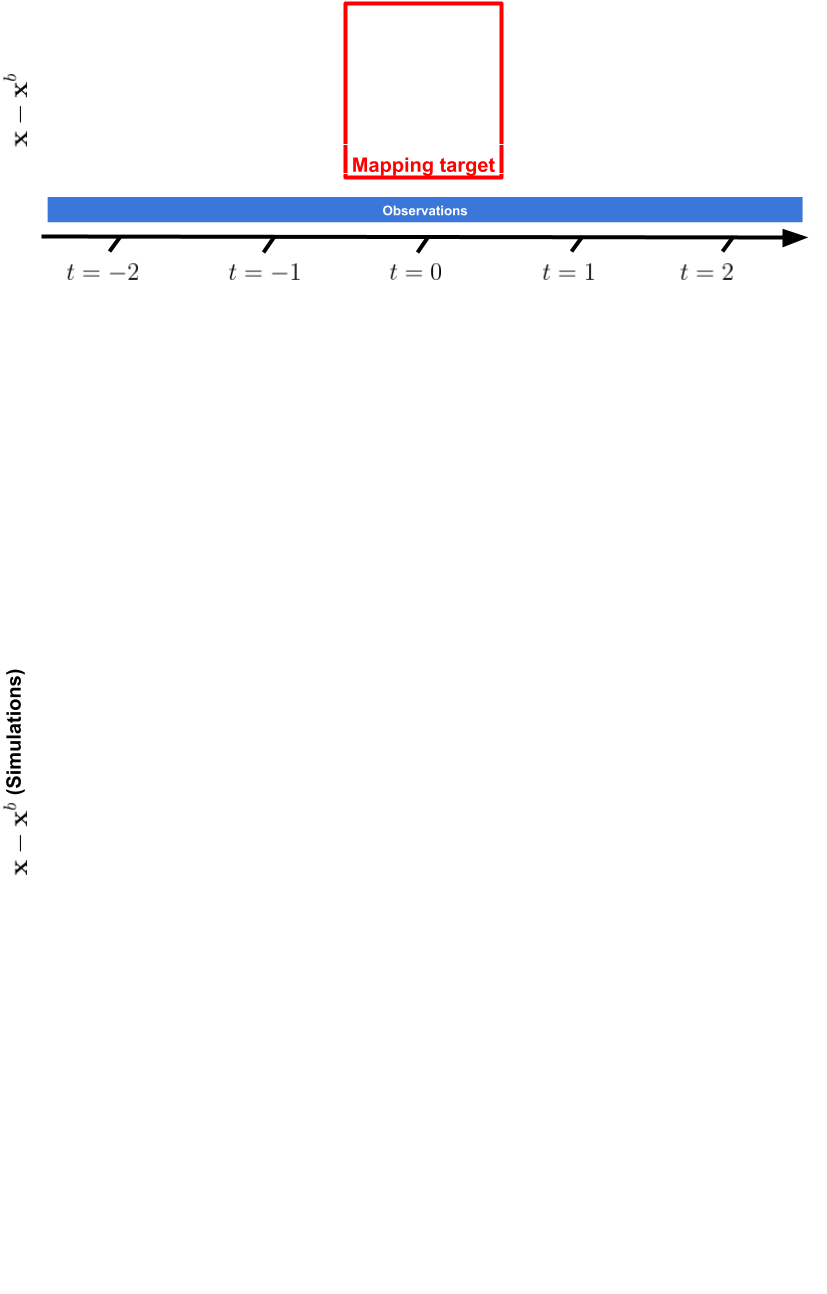}%
    };
}
\caption{Top panel: an example of the Ground Truth Sea Surface Height anomaly between $\x$ and $\x^b$ on the Gulf Stream domain used in our second realistic experiment (see Section 4.\ref{realistic_SSH}); Bottom panel: five simulations of the same anomaly based on the SPDE parametrization $\boldsymbol{\theta}$ learned after training. Because we are looking at anomaly SSH fields normalized for training consideration here, the most energetic part of the Gulf Stream domain relates to pale yellow/dark blue areas in the maps, that are consistent between what we see in the sample Ground Truth displayed in the top panel.}
\label{ex_simu}
\end{figure}

Such a strategy also shares many similarities with the so-called conditional generative models in deep learning, see \citep{goodfellow_2014, Kingma_2022, Dinh_2017, Ho_2020}. Instead of using neural networks to learn how to simulate in the prior distribution, we use here a generic class of advection-diffusion SPDE which we see as a first step towards physically-sounded generatives models in learning-based methods. Fig. \ref{ex_simu} displays an example of such non-conditional simulations for the term $\x-\x^b$, again for Application 2 on realistic SSH datasets over some energetic area along the Gulf Stream, see Section \ref{results}, i.e. the anomaly between the true state sequence and the deterministic mean $\x^b$, retrieved from a preliminary 4DVarNet coarse resolution scheme. We can appreciate how both the generic class of SPDE selected here, together with the training of its parameter, leads to realistic anomalies with a clear increase of the variance along the main meander of the Gulf Stream, due to a correct distribution of SPDE parameter $\tau^\star(\s,t)$. For a detailed analysis of the parameter estimation, please report to Section \ref{results}.

\subsection{Learning scheme}
\label{learning_scheme}

\paragraph{Training loss.} The joint problem of estimating the best reconstruction and infering realistic SPDE parametrizations is difficult because according to the size and nature of the dataset, the spatio-temporal interpolation may not always benefit from knowing the exact set of true SPDE parameters. Indeed, if the degree of sparsity of the observation dataset is low, the reconstruction may be good despite a poor estimation of the covariance matrix. In the other way, if the degree of sparsity is high, much more difficult will be the estimation of the SPDE underlying parameters. In this section, we benefit from the supervised configuration of neural variational scheme to train this joint problem. In this Section, we show how to embed the global precision matrix defined by Eq. \ref{global_prec_mat_2} in our neural scheme and define which training loss is the more appropriate to handle the bi-level optimization scheme (of both the inner variational cost and the training loss). 

For the training process, we have to consider two different loss functions:
\begin{itemize} 
\item $\mathcal{L}_1(\mathbf{x},\mathbf{x}^\star)=||\mathbf{x}-\mathbf{x}^\star||^2$ is the L2-norm of the difference between state $\mathbf{x}$ and reconstruction $\mathbf{x}^\star$
\item $\mathcal{L}_2(\mathbf{x},\boldsymbol{\theta}^\star) =  -|\mathbf{Q}^b_{\boldsymbol{\theta}^\star}| + \mathbf{x}^{\mathrm{T}} \mathbf{Q}^b_{\boldsymbol{\theta}^\star}\mathbf{x} $ is the negative log-likelihood of the true states given the estimated precision matrix $\mathbf{Q}^b_{\boldsymbol{\theta}^\star}$, thus ensuring consistency between the actual ground truth and the SPDE parameters. 
\end{itemize}  

Using $\mathcal{L}_1$ will lead to satisfactory reconstructions without any constraints on the SPDE parameters. The single use of $\mathcal{L}_2$ should lead to satisfactory results if the analytical solver, i.e. the inversion of the linear system or the gradient-based minimization of the variational cost, were used. But because the solver is trained, it also needs to be constrained by an appropriate loss function for the reconstruction, meaning $\mathcal{L}_1$. The best solution is then to create a mixed loss function, combination of $\mathcal{L}_2$ to estimate at best the SPDE parameters and optimize the prior model, and $\mathcal{L}_1$ to satisfy the reconstruction criteria and optimize the solver. \\
The log-determinant of the precision matrix $\log|\mathbf{Q}^b_{\boldsymbol{\theta}^\star}|$ is usually difficult to handle when computing $\mathcal{L}_2$. Hopefully, based on the particular structure and the notations already introduced for the spatio-temporal precision matrix $\mathbf{Q}^b_{\boldsymbol{\theta}}$ in Eq.(\ref{global_prec_mat}), it writes, see e.g. \citet{clarotto_2022}: 
\begin{align}
\label{log_det_Q}
\log|\mathbf{Q}^b_{\boldsymbol{\theta}^\star}| & = \log|\mathbf{P}_0^{-1}| + \log|\mathbf{S}_1^{-1}| + \cdots + \log|\mathbf{S}_L^{-1}| \nonumber \\
               & = \log|\mathbf{P}_0^{-1}| + \sum_{i=1}^L \log \left(|\mathbf{L}_i\mathbf{L}_i^{\mathrm{T}}|\right) \nonumber \\
               & = \log|\mathbf{P}_0^{-1}| + 2 \sum_{i=1}^L \sum_{j=1}^m \log \mathbf{L}_i (j,j)
\end{align}       
where $\mathbf{L}_i$ denotes here the Cholesky decomposition of $\mathbf{S}_k^{-1}$. In case of unsupervised learning, the same strategy may apply but $\mathcal{L}_2$ will be the likelihood of the observations given the estimated SPDE parameters since in this case, the true states would not be available during the training process.

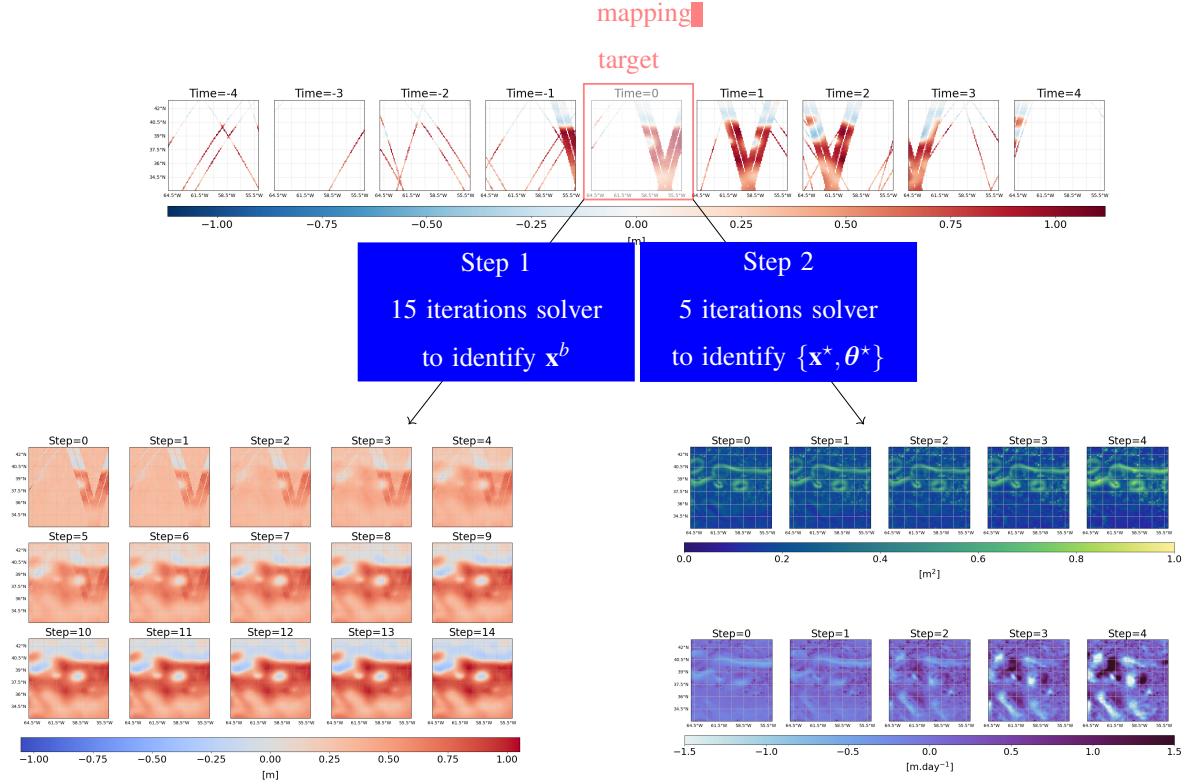
\begin{figure}[H]
\centering
\scalebox{.85}{
\tikz{
    \node(obs){
       \includegraphics[width=15cm]{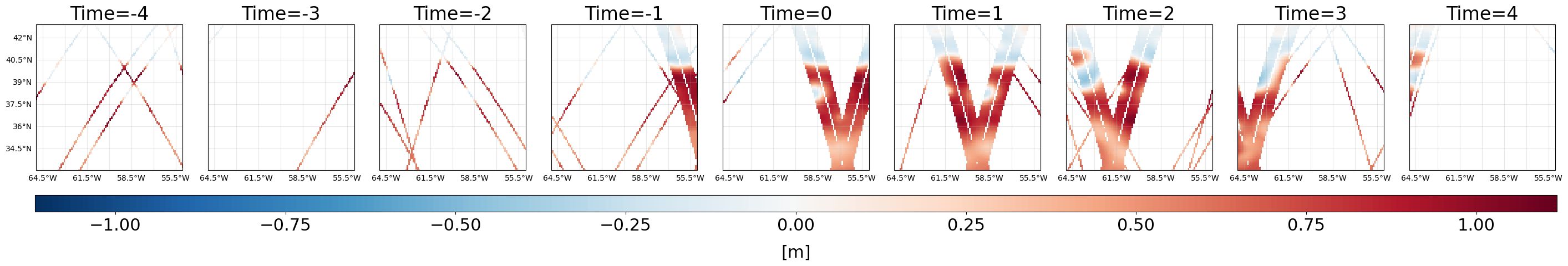}
    };
    \node[below = 4cm of obs, xshift=2cm] (fg) at (obs.west)   {%
        \includegraphics[width=8cm]{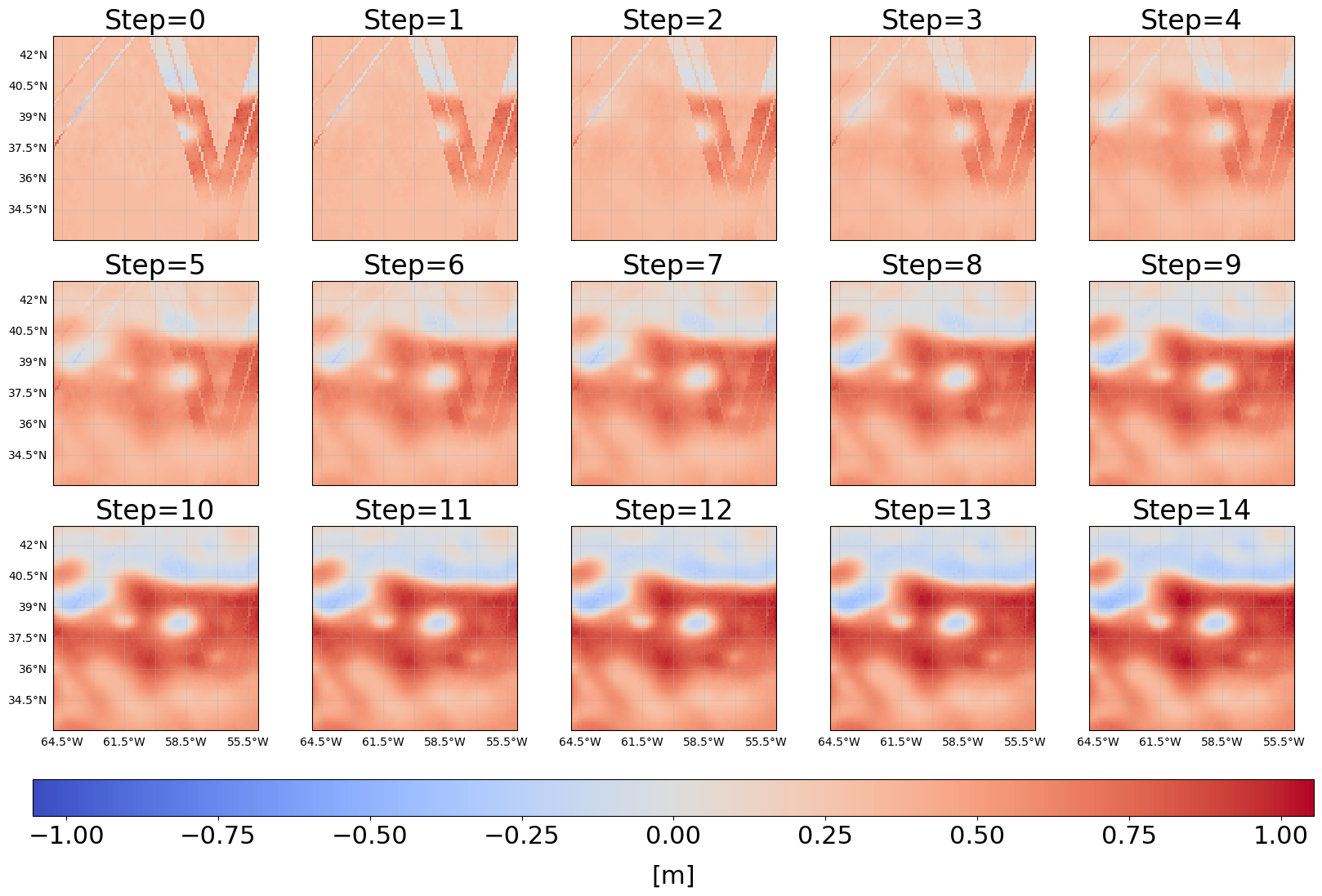}%
    };
    \node[below = 4cm of obs, xshift=-3cm] (tau) at (obs.east) {%
        \includegraphics[width=8cm]{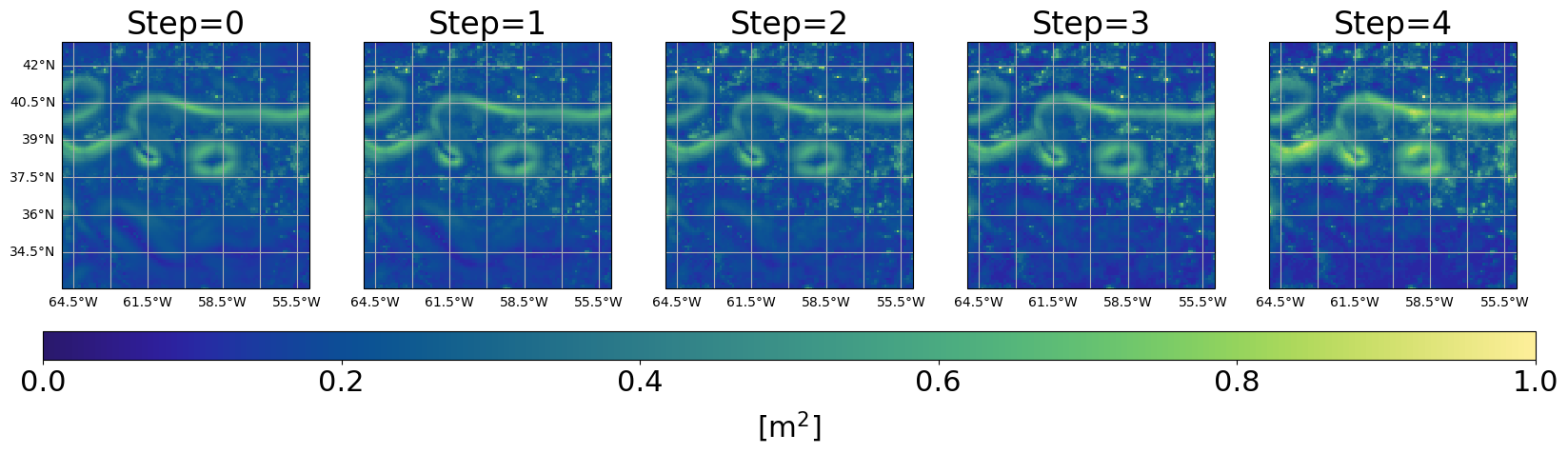}%
    };
    \node[below = 7cm of obs, xshift=-3cm] (m) at (obs.east) {%
        \includegraphics[width=8cm]{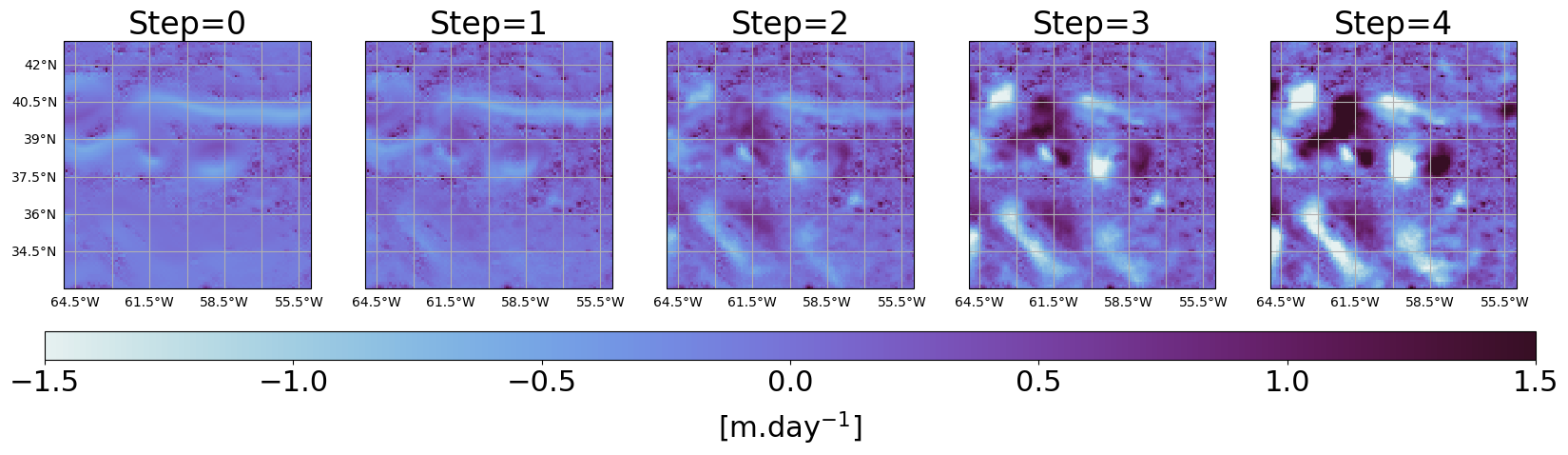}%
    };
    \begin{scope}[transparency group, opacity=.5]
        \draw[color=red, fill=white, thick] (obs.center) ++ (-.7,-.5) rectangle (1,1.3) node[above, text width=1cm, xshift=-1cm, align=center] (target) {mapping target};
    \end{scope}
    \draw (obs.center) ++ (-.7,-.5)  edge[->] node[draw, fill=blue!50, blue, thick, text width=4cm, align=center] {\textcolor{white}{Step 1\\ 15 iterations solver to identify $\x^b$}} (fg);
    \draw (obs.center) ++ (1,-.5)  edge[->] node[draw, fill=blue!50, blue, thick, text width=4cm, align=center] {\textcolor{white}{Step 2\\  5 iterations solver to identify $\lbrace{ \x^\star, \boldsymbol{\theta}^\star \rbrace}$}} (tau);
}
}%
\caption{Two-step adaptation of the 4DVarNet scheme: on the top panel, an example of 4 nadirs+SWOT observations along the 9-day data assimilation window, see Application 2 in Section \ref{results}. The target reconstruction day is at the center of the DAW. On the bottom-left panel is displayed the 15 iterations of the first 4DVarNet solver to retrieve the prior mean $\x^b$: as initial condition $\x^{(0)}$, the non-observed parts of the domain are filled with the global mean of the training dataset (0 when normalized). On the bottom-right panel is displayed the 5 iterations of the modified 4DVarNet solver with augmented state formalism to retrieve both $\x^\star$ and $\boldsymbol{\theta}^\star$: at the beginning, the parameters in $\boldsymbol{\theta}$ are initialized with partial  derivatives of $\x^b$, see Section 3.\ref{neuralsolver}} 
\label{2step_ls}
\end{figure}

\paragraph{Two-step learning schemes.} In the initial version of 4DVarNet schemes, the output $\x^\star$ provided by the neural formulation is deterministic and can be seen as the posterior mean of the state given the observations. Within this SPDE-based parametrization of the prior $\x^b$, we also aim at providing the distribution of the prior as a GP process. Though, for realistic geophysical fields, even the prior cannot be considered as a zero-mean Gaussian field and some non-linearities have to be accounted for in its deterministic mean $\x^b$. To solve for this specific case, we involve a 2-step learning process in which the deterministic mean $\x^b$ is first estimated by a 4DVarNet scheme applied on coarser resolution than the actual observations. Second, the modified 4DVarNet scheme is involved to estimate jointly the posterior mean $\x^\star$ together with the SPDE parametrization $\boldsymbol{\theta}^\star$ of the prior $\x \sim \mathcal{N}\left(\x^b,\mathbf{Q}^b_{\boldsymbol{\theta}^\star}\right)$. This two-steps procedure also enables to simplify the SPDE training scheme: the first guess is estimated based on the entire set of observations along the state sequence, while the SPDE parametrization is estimated only on a reduced window of length $5$, centered on the targeted time of interest. This considerably reduces the size of precision matrix $\mathbf{Q}^b_{\boldsymbol{\theta}}$, making the algorithm scalable for any application. Fig. \ref{2step_ls} shows a schematic overview of this two-steps learning scheme with illustrations coming from realistic Sea Surface Heigh datasets provided in Application 2, Section \ref{results}. In the end, this will lead to a stochastic version of the neural variational scheme based on the prior GP distribution in which we can sample ensemble members. 

\paragraph{Complementary PyTorch developments.} Regarding the implementation of our model, we use Pytorch \citep{pytorch} whose sparse linear algebra is not providing yet a sparse Cholesky algorithm and a sparse solver of linear systems, which is critical, especially when computing the likelihood $\mathcal{L}_2$ in the training loss function. As a consequence, despite the theoretical tools have been fully detailed in the previous sections, we had to implement new functionalities based on scipy sparse linear algebra \citep{scipy} to store the precision matrices in an efficient way and compute the inner variational cost, see again Eq.\ref{var_cost}. In particular, we provide a PyTorch extension of sparse cholesky matrices, see Appendix \ref{App:AppendixB}, based on the CHOLMOD supernodal Cholesky factorization \citep{cholmod} and draw from \citet{Seeger_2019} to provide the backward pass of the sparse Cholesky decomposition.\\

\section{Results}
\label{results}
In this Section, we provide two applications of this work:
\begin{itemize}
\item The first example relies on a spatio-temporal GP simulation driven by a non-stationary spatial diffusion tensor. Because the dynamical process is linear, the best reconstruction is provided by the optimal interpolation using the SPDE parameters used in the simulation to fill in the precision matrix.
\item The second example uses an Observation System Simulation Experiment (OSSE) of the Sea Surface Height (SSH) along the Gulf Stream. We will use the SPDE-based prior as a surrogate model along the data assimilation window to provide ensemble members of the posterior distribution.
\end{itemize}
Each experiment comes from three detailed subsections, respectively presenting the dataset, the training setting (how training, validation and test data are separately generated) and finally the results.

\subsection{Diffusion-based non stationary GP}

\textbf{Dataset}. In this first application, we simulate 500 states of a GP driven by the following diffusion SPDE:
\begin{align} 
\label{spde_diff}
\left\lbrace \kappa^2 - \nabla \cdot\mathbf{H}(\mathbf{s})\nabla \right \rbrace^{\alpha/2} \mathbf{x}(\mathbf{s},t)=\tau \mathbf{z}(\mathbf{s},t)
\end{align}
The regularity parameter $\kappa=0.33$ is fixed over space and time. To ensure the GP to be smooth enough, we use a value of $\alpha=4$. Such a formulation enables to generate GPs driven by local anisotropies in space leading to non stationary spatio-temporal fields with eddy patterns. The diffusion tensor $\mathbf{H}$ is a 2-dimensional diffusion tensor generated by drawing from the spatial statistics literature, see e.g. \citep{fuglstad_2015a}. We introduce a generic decomposition of $\mathbf{H}(\mathbf{s},t)$ through the equation:
\begin{align*} 
\mathbf{H}=\gamma \mathbf{I}_2 + \beta \mathbf{v}(\mathbf{s})^{\mathrm{T}}\mathbf{v}(\mathbf{s})
\end{align*}
with $\gamma=1$, $\beta=25$ and $\mathbf{v}(\mathbf{s})=(v_1(\mathbf{s}),v_2(\mathbf{s}))^{\mathrm{T}}$ using a periodic formulation of its two vector fields components, see Section \ref{spde_param}. We use the Finite Difference Method in space coupled with an implicit Euler scheme in time to solve the equation. Let $\mathcal{D}=[0, 100] \times [0, 100]$ be the square spatial domain of simulation and $\mathcal{T}=[0, 500]$ the temporal domain. Both spatial and temporal domains are discretized so that the simulation is made on a uniform Cartesian grid consisting of points ($x_i$, $y_j$, $t_k$) where $x_i$=$i \Delta x$, $y_j$=$j \Delta j$, $t_k$=$k \Delta t$ with $\Delta x$, $\Delta y$ and $\Delta t$ all set to 1.

\begin{figure}[H]
\centering
\includegraphics[width=16cm]{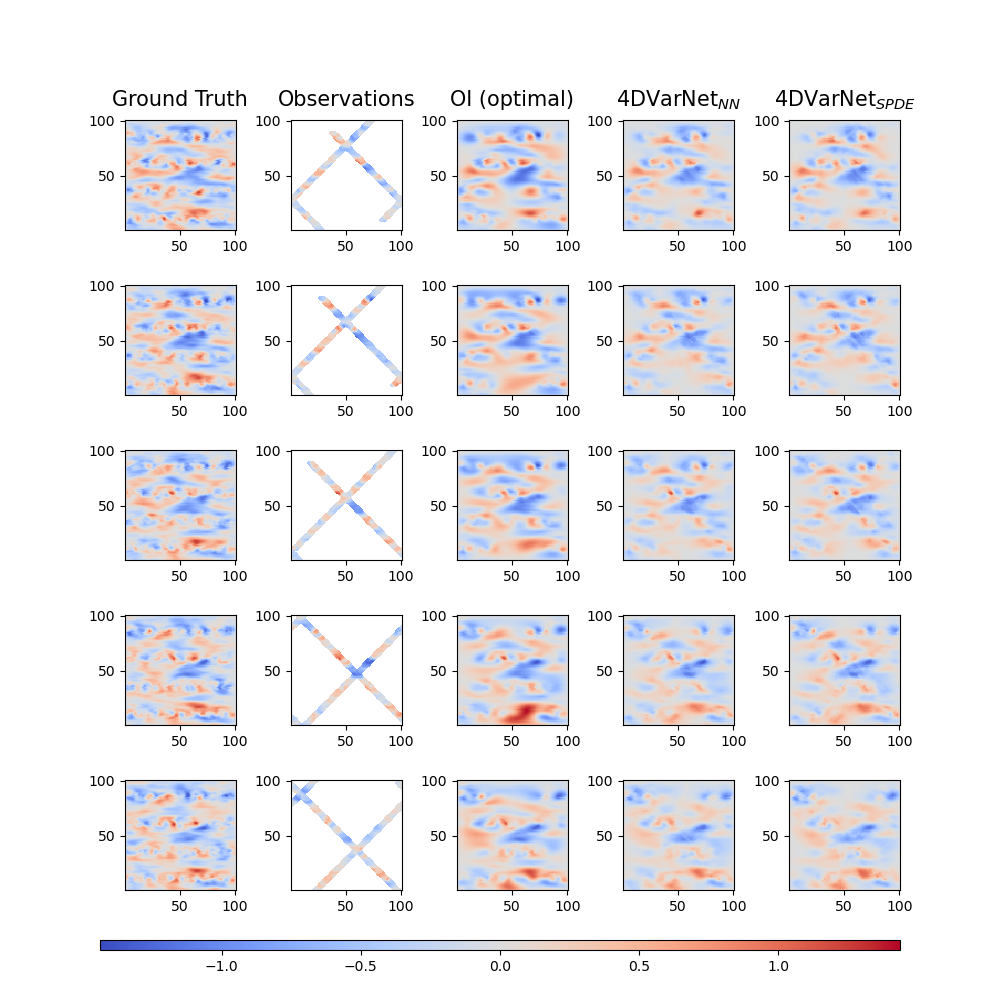}
\caption{From left to right: Gaussian process Ground Truth simulation for 5 days in the test period (top to bottom), pseudo-observations, Optimal Interpolation (with know covariance matrix, hence the best solution possible), neural variational scheme with UNet-based prior $\Phi$ and SPDE-based prior operator $\mathcal{N} \sim (\mathbf{0},\mathbf{Q}^b_\theta)$. A data assimilation window of length 5 is used}
\label{rec_GP}
\end{figure}

\textbf{Training setting}. We train the neural architectures for both UNet-based and SPDE-based priors with Adam optimizer on 50 epochs. The training period goes from timestep 100 to 400. During the training procedure, we select the best model according to metrics computed over the validation period from timestep 30 to 80. Overall, the set of metrics is computed on a test period going from timestep 400 to 500. No further improvements in the training losses are seen when training the model longer. We use a data assimilation window of length 5 and generate pseudo-observations from the ground truth, inspired by orbiting satellites tracks around the earth, see Application 2 in Section \ref{realistic_SSH}.\\

\textbf{Results}. Fig. \ref{rec_GP} displays the results obtained by Optimal Interpolation and the neural implicit solver with both UNet and SPDE-based parametrization of the prior. Neural-based reconstructions are optimized at the center of the assimilation window ($T=2$), which is why the performance may be affected for other leadtimes. There is no significant differences between the two prior formulations used in the neural scheme, which overall retrieve the main patterns of the OI. Some artefacts may appear due to the observation term in the inner variational cost that tends to lead the solution towards the observation in its close neighbourhood. Improvements may be expected when adding regularization terms in the training loss to counteract such effects, see e.g. \citet{beauchamp_2023b}. Regarding the derived framework proposed in this work, one of the question was: is it possible to retrieve interpretable SPDE parametrizations from the joint learning setting? Two configurations were considered: when using as initial condition for the parametrization a gradient-based information of the accumulated alongtrack observations, i.e. $\mathbf{H}_{11}=\nabla_{\overrightarrow{x}}{\mathbf{y}}$ and $\mathbf{H}_{22}=\nabla_{\overrightarrow{y}}{\mathbf{y}}$; or an isotropic initialization, i.e.  $\mathbf{H}=\mathbf{I}$. The first configuration enables to identify patterns in zonal and meridional components of the true diffusion tensor and leads to an optimal parametrization $\boldsymbol{\theta}^\star$ very close to the true diffusion tensor. While leading to different SPDE parametrizations $\boldsymbol{\theta}^\star$, the interpolation metrics are similar in the end for the two initializations. In addition, using an isotropic initial condition is more general (see the next realistic SSH application) and also retrieves in the end the main zonal flow directions encoded by $\mathbf{H}_{22}$, while the meriodional and periodic structures of $\mathbf{H}_{11}$ and $\mathbf{H}_{12}$ are partly seen as well. 

\begin{minipage}{.95\textwidth}
\centering
\begin{figure}[H]
\centering
\includegraphics[width=8cm]{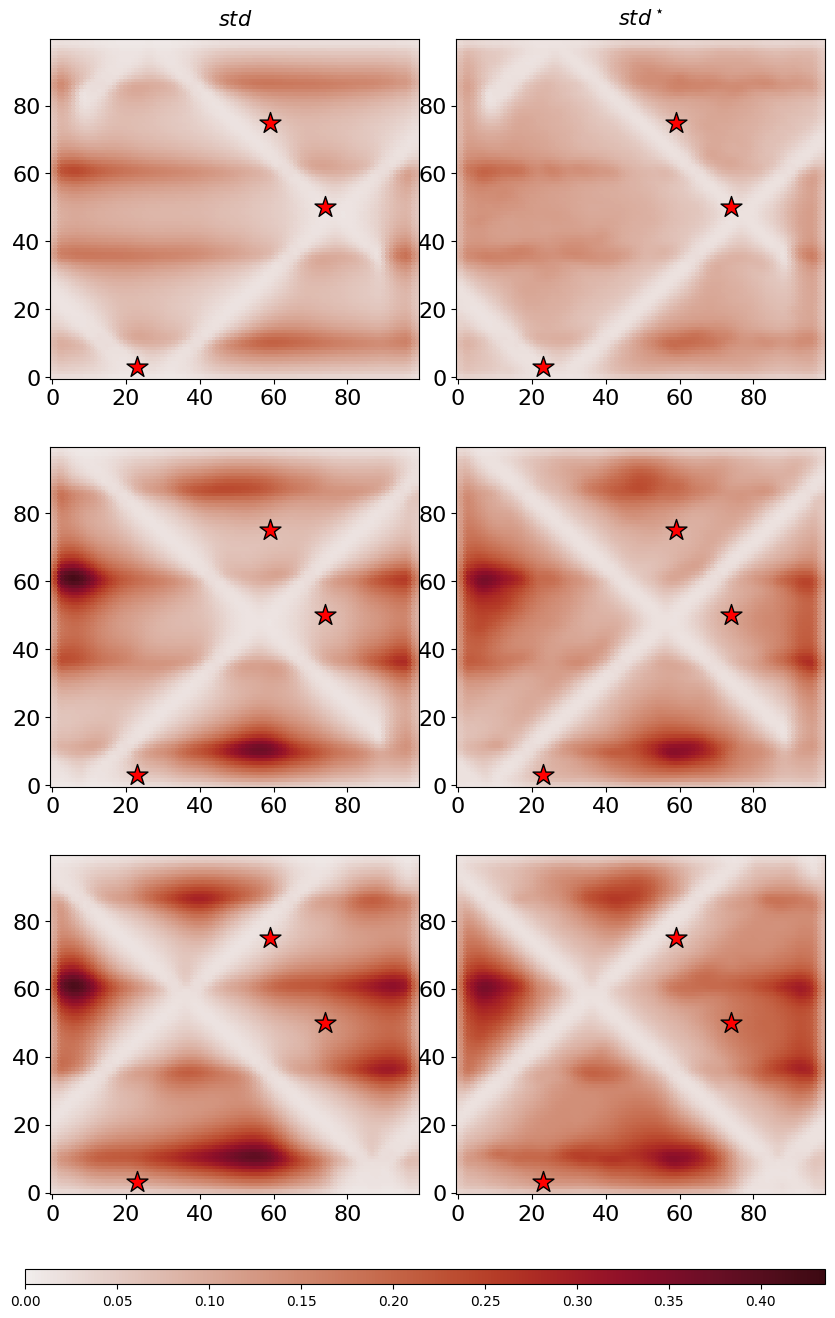}
\caption{True (left) and estimated (right) posterior standard deviations at the beginning, center and end of the assimilation window $time={0,2,4}$ for a given time in the test period}
\label{posterior_cov_GP}
\end{figure}
\end{minipage}
\begin{minipage}{.95\textwidth}
\centering
\begin{figure}[H]
\centering
\includegraphics[width=7.5cm]{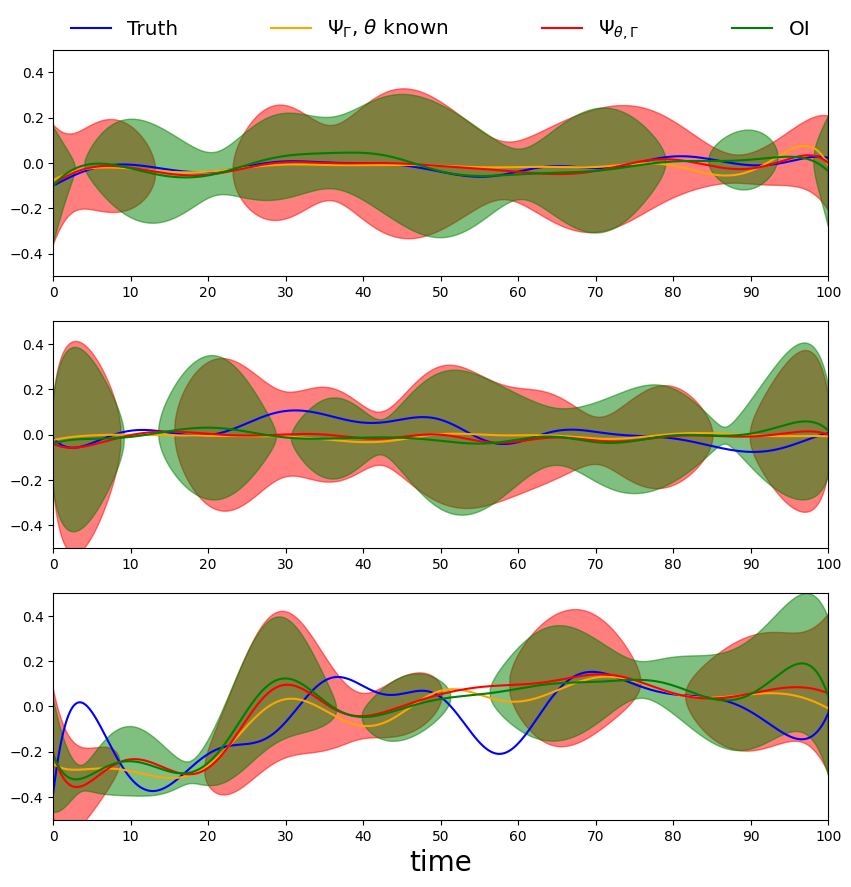}
\caption{Ground truth, OI and its posterior variance (blue), neural scheme with SPDE parametrization $\boldsymbol{\theta}$ known (orange) and with inference of $\boldsymbol{\theta}$ (red) for the same three points identified in Fig. \ref{posterior_cov_GP} along the 100 time step test period}
\label{prior_cov_GP_ts}
\end{figure}
\end{minipage}

Because we know the process is Gaussian, we know that $\mathbf{Q}(\x|\y)=\mathbf{Q}^b +  \mathbf{H}^{\mathrm{T}} \mathbf{R}^{-1} \mathbf{H}$, see Section \ref{back}, to compute the closed form of the posterior pdf. Looking at the estimated posterior variance obtained starting from isotropic initial condition of the parameters, see Fig. \ref{posterior_cov_GP}, we understand that even if the initial set of parameters is not retrieved, they still remain interpretable. Visually, a simulation produced by this set of SPDE parameters is also consistent and propose a spatio-temporal diffusion process close to the original one. It also highlights the dependency of the OI uncertainty quantification to the sampling scheme, especially here where no noise is added to the partial set of observations. On Fig. \ref{prior_cov_GP_ts} is also shown the time series on three different locations (red stars) $\in \mathcal{D}$, see again Fig. \ref{posterior_cov_GP}, of the GT (blue line) along the test period (100 time steps), the OI and corresponding standard deviation (green line), and the neural scheme estimations when the SPDE parametrization is known (orange line) or estimated (red line with associated uncertainty). Both neural scheme configurations are generally close which validates the capability of this framework to estimate jointly both state and prior parametrization. Because the neural scheme is optimized on the global MSE, its solution may deviate more or less significantly from the OI depending of the position we are looking at in domain $\mathcal{D}$.\\

\begin{figure}[H]
\centering
\includegraphics[width=8cm]{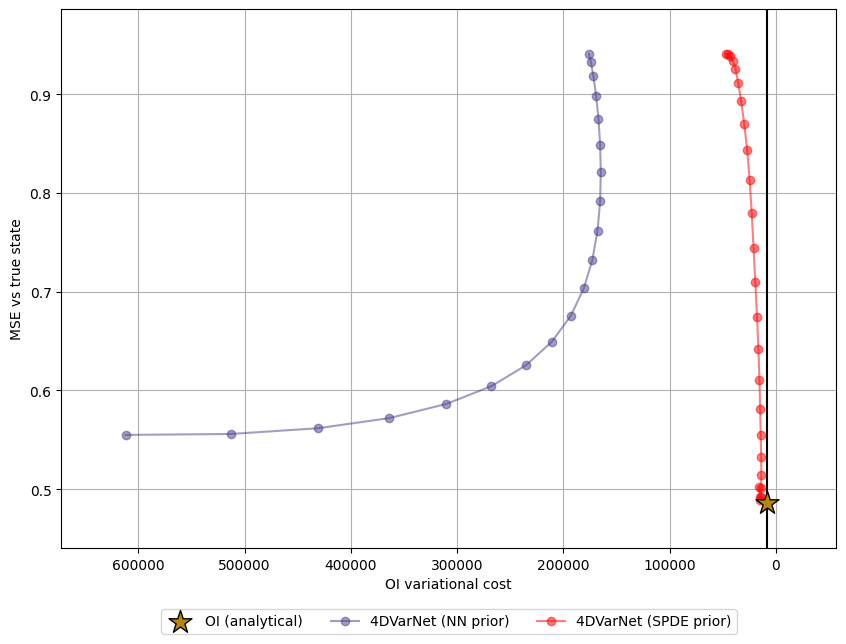}
\caption{Optimal Interpolation derived variational cost vs Mean Squared Error (MSE) loss (a) for the gradient-based descent of the variational cost, the classical implementation of the neural scheme and its SPDE-based prior formulation. For the analytical Optimal Interpolation solution (the yellow star), there is no iterations, then a single point is displayed.}
\label{cmp_loss_GP}
\end{figure}

Last, Fig. \ref{cmp_loss_GP} provides the scatterplot of the global MSE w.r.t the OI variational cost:
\begin{align*} 
  \mathcal{J}_{OI}(\y, \mathbf{x},\boldsymbol{\theta}^\star) =  ||\y-\mathbf{H}\mathbf{x}||^2 + \mathbf{x}^\mathrm{T} \mathbf{Q}^b_{\boldsymbol{\theta}^\star}\mathbf{x}
\end{align*}
throughout the iteration process after training of the neural schemes. For SPDE-based prior, the initial parametrization $\boldsymbol{\theta}^{(0)}$ relates to an isotropic GP process. LSTM-based iterative solvers are all consistent with the optimal solution, i.e. the best linear unbiased estimator with minimal error variance, in terms of MSE. When the latter is used as training loss, 20 iterations is enough to reach satisfactory performance. Using the same loss for inner  variational cost and outer training loss (not shown here), see \citet{beauchamp_2022b}, would require more iterations to converge. Also, constraining the prior to follow the same SPDE simulation ensures to also jointly minimize the OI variational cost (asymptotic convergence of the red line to the yellow star on Fig. \ref{cmp_loss_GP}) which is not the case when looking for an optimal solution within the bi-level neural optimization of prior and solver (blue line) that may lead to deviate from the original variational cost to minimize. Let note that by construction, the analytical OI solution is optimal regarding the OI variational cost: it is  unbiased with minimal variance. In other words, at a given spatio-temporal location $(\mathbf{s},t)$, its variance (which is the local MSE) is minimal. In our case, we compute the global MSE over the entire domain $\mathcal{D} \times [t_k-2,t_k+2]$ w.r.t the true state because we have only one single realization to compute this metrics. This is why the global MSE (see $\mathcal{L}_1$) of the OI may be outperformed by learning-based methods.

\subsection{Realistic SSH datasets}
\label{realistic_SSH}

\begin{figure}[H]
\centering
  \includegraphics[width=12cm]{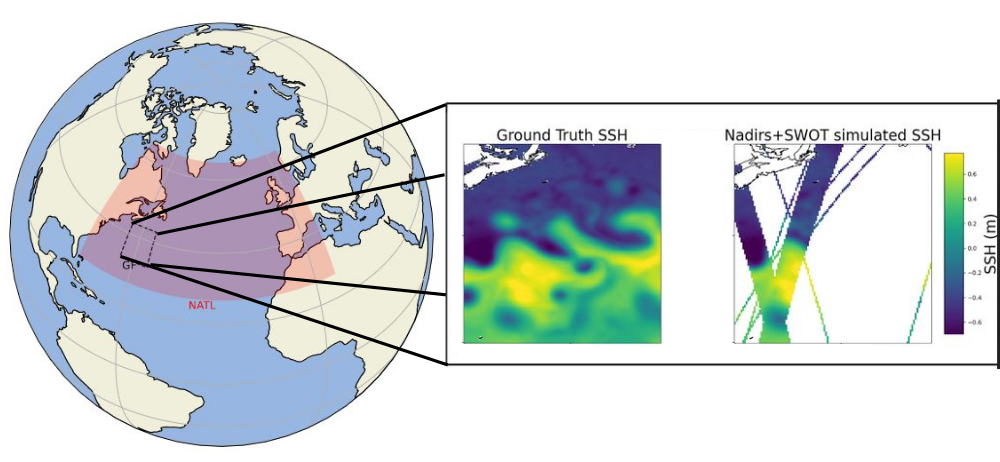}
  \caption{NATL60 and GULFSTREAM domain and zoom-in picture for one day Ground Truth and accumulated along-track nadir + wide-swath SWOT SSH pseudo-observations}
  \label{fig_ssh_data_GULFSTREAM}
\end{figure}

\textbf{Dataset}. In this application on Sea Surface Height (SSH) spatio-temporal fields, we focus on a small part of the GULFSTREAM, see Fig. \ref{fig_ssh_data_GULFSTREAM}, mainly driven by energetic mesoscale dynamics, to illustrate how our framework may help to solve for the oversmoothing of the state-of-the-art Optimal interpolation (OI) and how the SPDE formulation of the prior is a consistent linearization of the dynamics in the data assimilation window that helps to generate ensemble members in the posterior distribution. We use an Observation System Simulation Experiment (OSSE) with the NEMO (Nucleus for European Modeling of the Ocean) model NATL60 high resolution basin-scale configuration \cite{jean-marc_molines_meom-configurations}. Based on this one-year long simulation, we generate pseudo along-track nadir data for the current capabilities of the observation system \citep{Ballarotta_2019} and pseudo wide-swath SWOT data in the context of the upcoming SWOT mission \citep{metref_2020}, with additional observation errors \citep{dufau_mesoscale_2016, esteban_2014,gaultier_2010}. The two types of observations may be merged, see Fig. \ref{fig_ssh_data_GULFSTREAM}, to produce a one-year long daily datasets of partial and noisy observations of the idealized Ground Truth (GT). Last the DUACS operational system (CMEMS/C3S Copernicus program) provides the Optimal Interpolation baseline \citep{taburet_duacs_2019} as daily gridded (0.25$^\circ$x0.25$^\circ$) products

\vspace{-.2cm}
\vspace{-.5cm}
\begin{figure}[H]
\centering
\includegraphics[width=15cm]{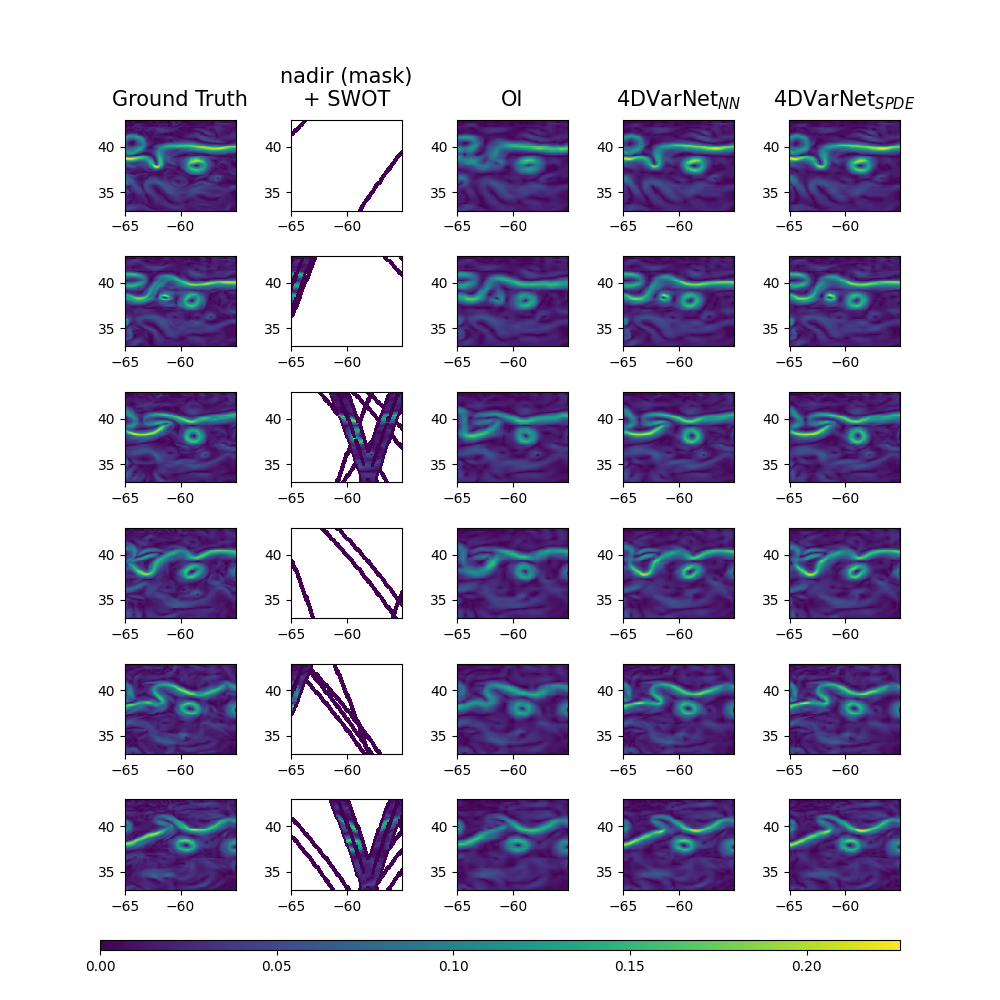}
\caption{From left to right: SSH Gradient Ground Truth, Pseudo-observations (nadir mask and SWOT Gradient field), DUACS Optimal Interpolation, and neural variational scheme results obtained with UNet-based $\Phi$ and SPDE-based prior operator $\mathcal{N} \sim (\mathbf{x}^b,\mathbf{Q}^b_\theta)$. A data assimilation window of length 5 is used}
\label{rec_osse}
\end{figure}

\textbf{Training setting}. All the datasets are downscaled from the original resolution of 1/60$^\circ$ to 1/10$^\circ$. For the training, the dataset spans from mid-February 2013 to October 2013, while the validation period refers to January 2013. All methods are tested on the test period from  October 22, 2012 to December 2, 2012. We still use Adam optimizer with 1100 epochs. Regarding the metrics, we use the ocean data challenge \footnote{https://github.com/ocean-data-challenges/2020a\_SSH\_mapping\_NATL60} strategy looking at RMSE-score, and both spatial and temporel minimal scales resolved. The reader may refer to \citep{beauchamp_2023b} for more details.\\

\textbf{Results}. Because we aim at assessing how the use of SPDE priors in the neural architecture is relevant, we used a Gaussian prior formulation $\mathbf{x} \sim \mathcal{N}\left( \x^b, \mathbf{Q}^b_{\boldsymbol{\theta}} \right)$ where the mean $\x^b$ is provided as a first guess by a preliminary 4DVarNet run. Doing so, the SPDE parametrization $\boldsymbol{\theta}$ is focused on the reconstruction and surrogate parametrization of the small scales not catched by this deterministic mean. Then, we can reduce the data assimilation window used to estimate the prior mean $\x^b$, to a reasonable length $L=5$ here, so that the storage of multiple (sparse) precision matrices throughout the computational graph remains possible, see Section 3.\ref{learning_scheme}.  Moving to longer time windows including mesoscale-related autocorrelations (more than 10 days) in this SPDE framework would lead to similar results, see e.g. \citet{febvre_2022} but would require moving to matrix-free formulations, with potential existing solutions, see e.g. \citet{pereira_2022}.

Last point on this experimental configuration: because the pseudo-observations are subsampled from hourly simulations but we target daily reconstructions, they are noisy due to representativity errors between the two temporal resolutions. This is not currently adressed by our framework where the observation term in the minimization cost, see Eq. \ref{var_cost}, is only the L2-norm of the innovations. But it might be easily considered, either by using a known observation error covariance matrix $\mathbf{R}$ or by learning one of its possible parametrization as an additional feature of the neural scheme.

\begin{figure}[H]
\centering
\subfloat[nRMSE]{
\includegraphics[width=6cm]{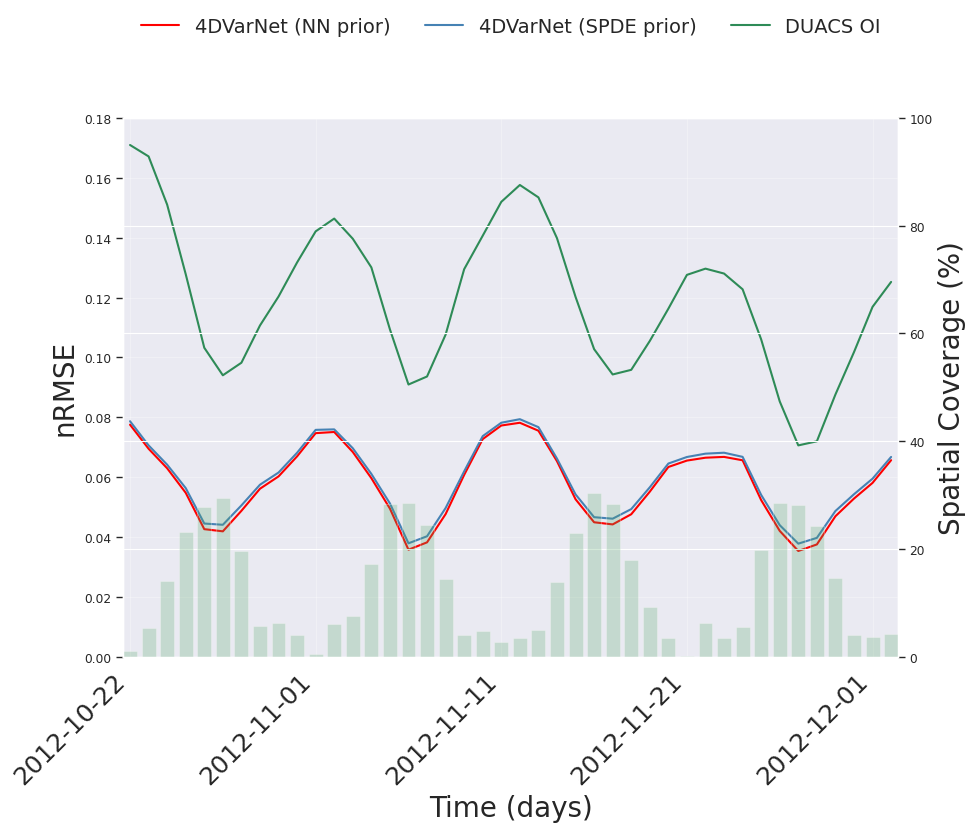}
\label{ts_osse}
}
\subfloat[Spectrum]{
\includegraphics[width=8cm]{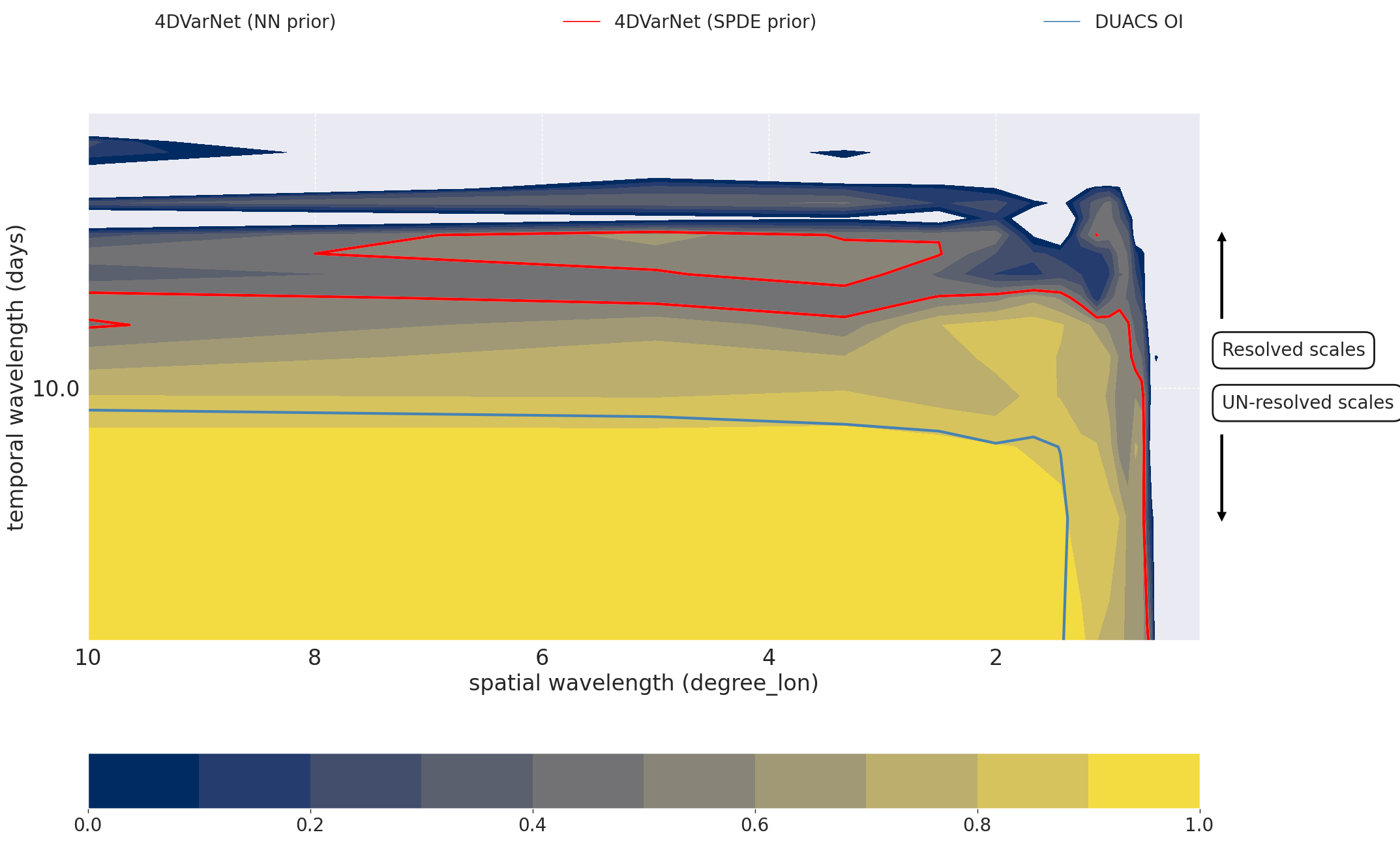}
\label{spectrum_osse}
}
\caption{For DUACS Optimal Interpolation, neural solvers with UNet-based and SPDE-based prior, (a) provides their temporal performance, i.e. nRMSE time series along the BOOST-SWOT DC evaluation period  ; and (b) displays their spectral performance, i.e. the PSD-based score is used to evaluate the spatio-temporal scales resolved in the GULFSTREAM domain (yellow area)}
\end{figure}

As in Example 7.1, we provide in Fig. \ref{rec_osse} the reconstructions, as the SSH gradients, obtained from DUACS OI baseline, and both neural solver formulations, the one using as prior a UNet-based parametrization, see e.g. \citep{beauchamp_2023a} and our SPDE-based formulation. As expected and already seen in previous related studies \citep{beauchamp_2023b, beauchamp_2023c, Fablet_2021}, the neural schemes improve the baseline by retrieving the dynamics along the main meander of the Gulf Stream and additional small energetic eddies. Again, there is no significant differences between the two neural formulations, which was again expected because the aim of the SPDE formulation is not to improve the mean state estimation obtained when using a neural prior operator, which is even more general, but to provide a a stochastic framework for interpretability and uncertainty quantification. This is supported by Figs. \ref{ts_osse} and \ref{spectrum_osse} respectively showing the normalized RMSE and the space-time spectrum along the test period. The periodic improvements of the score are due to the SWOT sampling that does not provide informations every day on this Gulf Stream domain. Overall, the nRMSE is in average improved by 60\% when using the neural architecture. For the spectrum, minimal spatio-temporal scales $\lambda_x$ and $\lambda_t$ also improve resp. by 30\% and 60\%. From both figures and scores provided in Table \ref{table_scores}, we can see that the UNet formulation of prior $\Phi$ leads to a small improvement in the reconstruction, which was expected because this is the only task optimized by the pure neural formulation. Introducing the SPDE formulation leads to optimize both reconstruction and likelihood of the parameters, which is more difficult. Though, the reconstruction performed by the latter is satisfactory enough and very close to the original solution proposed in \citet{Fablet_2021}. It also competes with  other state-of-the-art method available in the ocean data challenges 2020a, among which DUACS OI \citep{taburet_duacs_2019}, MIOST (Multi-scale OI) \citet{Ardhuin_2020} or a 4DVar scheme based on a QG dynamical model \citet{LEGUILLOU_2022}. While DUACS OI has minimal spatial and temporal resolution of 1.22° and 11.15 days, 4DVarNet reaches 0.62° and 4.35 days, which reaches similar order of performance that combining a 4DVar with a QG model.

\begin{figure}[H]
\centering
\includegraphics[width=16cm]{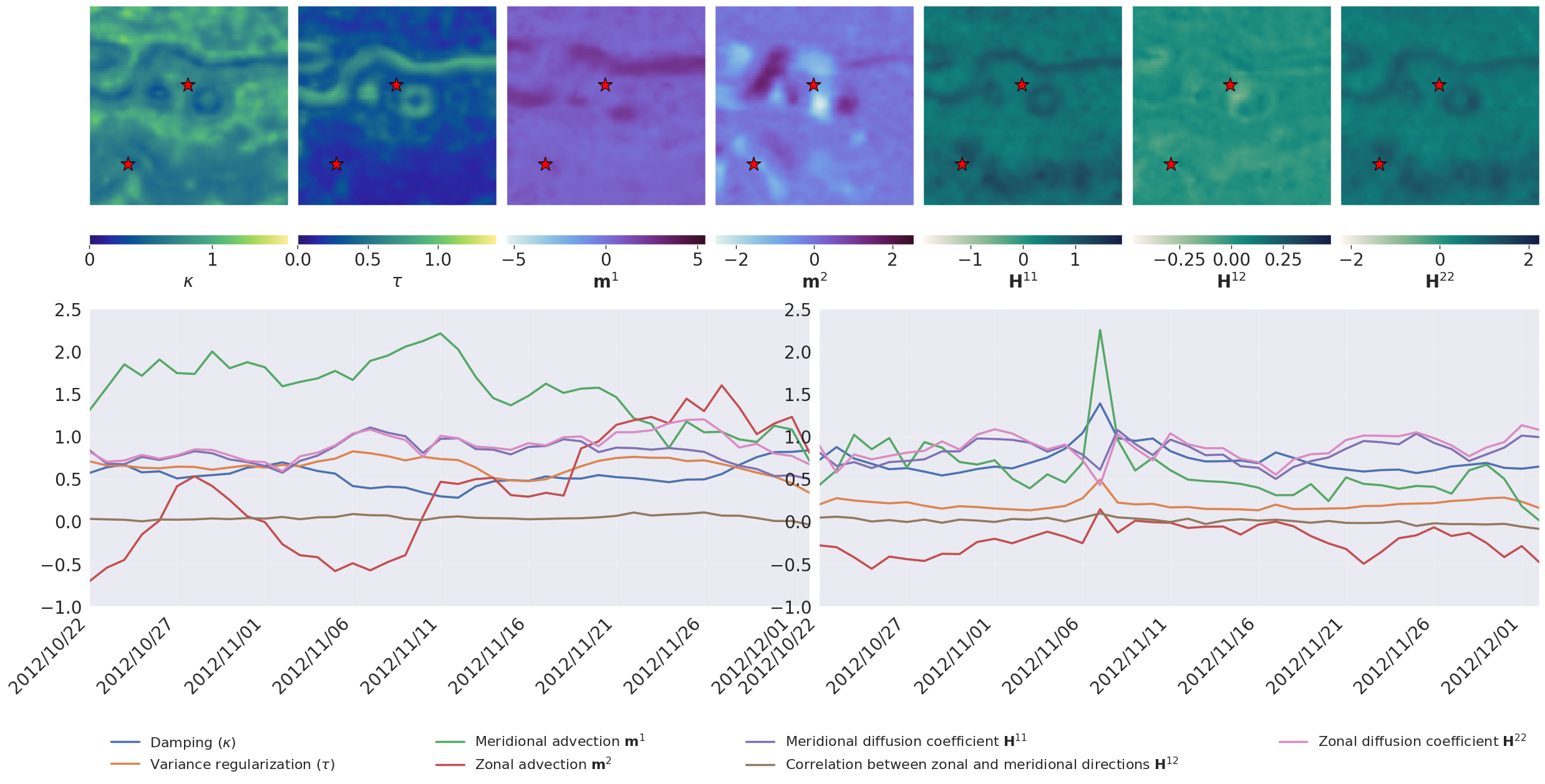}
\caption{Parameter estimation of the SPDE prior:  in the neural scheme along the 42 days test period and every 6 days. Top, from left to right: $\boldsymbol{\tau}$, $\mathbf{m}^1$, $\mathbf{m}^2$ (advection fields), $\mathbf{H}^{1,1}$, $\mathbf{H}^{1,2}$ and $\mathbf{H}^{2,2}$ (diffusion tensor) estimated on the first day of the test period. Bottom: time series of these parameters along the test period on the two locations identified by the red stars. The first one (left) is located along the Gulf Stream meander and the second one (right) is in the less energetic left-lower part of the domain.}
\label{param_osse}
\end{figure}

In this very general setup, where the equation-based dataset provide a supervised learning setting on state $\mathbf{x}$, but not on the SPDE parameters $\boldsymbol{\theta}$, two main questions are raised. On the first one: does the parameters retrieved by the iterative solver are interpretable? If considering Fig. \ref{param_osse} that shows parameters  $\boldsymbol{\tau}$, $\mathbf{m}_1$, $\mathbf{m}_2$ (advection fields), $\mathbf{H}_{11}$, $\mathbf{H}_{12}$ and $\mathbf{H}_{22}$ (diffusion tensor) on the first day of the test period, they seem consistent with the SSH field $\mathbf{x}$ that partially encodes the SPDE parametrization, which also opens avenue for state-dependent parameters. We also show the time series along the 42 days of the test period for these parameters on two locations of the Gulf Stream domain (red stars): a first one right in the Gulf Stream meander (left time series) and a second one in the left-lower part of the domain, with less variability (right time series). Interestingly, in less energetic areas the parameters are almost perfectly correlated while in the Gulf Stream, they might behave differently. Playing with both damping and variance regularization parameters provide a flexible way to handle complex GP priors with both low and high marginal variances for a given time. This is a key aspect here because the range of possible values attributed to the anomaly generally differs according to the spatio-temporal dynamics of the SSH: it is high along the main meander of the Gulf Stream and eddies not catched up by the OI, and lower elsewhere. This is also supported by generating non-conditional simulations based on these parameters and comparing it to the ground truth anomaly field, see again Fig. \ref{ex_simu}. The spatio-temporal fields are clearly consistent with the original simulation which also makes the link between our approach and generative modeling. The average of a large number of simulations would be a zero state, while its covariance relates to the model error matrix used in equation-based DA to perturb the members in ensemble methods. 

\begin{table}[H]
\centering
\begin{tabular}{l r r r r}
\hline
 & $\mu$(RMSE) & $\sigma$(RMSE) & $\lambda_x$ (degree) & $\lambda_t$ (days) \\
\hline
  OI & 0.92 & 0.02 & 1.22 & 11.06 \\
  MIOST & 0.94 & 0.01 & 1.18 & 10.33 \\ 
  4DVar-QG & 0.96 & 0.01 & 0.66 & 4.65 \\ 
  4DVarNet (NN prior) & 0.96 & 0.01 & 0.62 & 4.35 \\
  4DVarNet (SPDE prior) & 0.96 & 0.01 & 0.64 & 5.03 \\
\hline
\end{tabular}
\caption{Evaluation of the performance between OI, MIOST, 4DVar-QG, 4DVarNet schemes with UNet and SPDE-based prior parametrization. The OSSE involves 1 SWOT + 4 nadirs as pseudo-observations.}
\label{table_scores}
\end{table}

\begin{figure}[H]
\centering
\subfloat[$\x-\x^\star$]{\includegraphics[width=15cm]{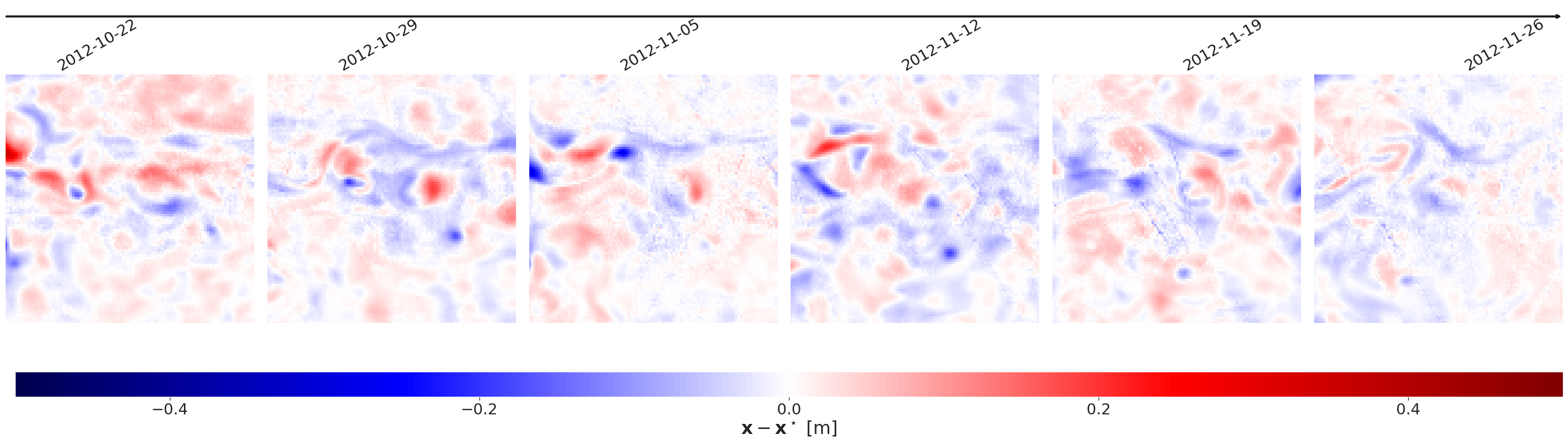}}\\
\subfloat[posterior standard deviation]{\includegraphics[width=15cm]{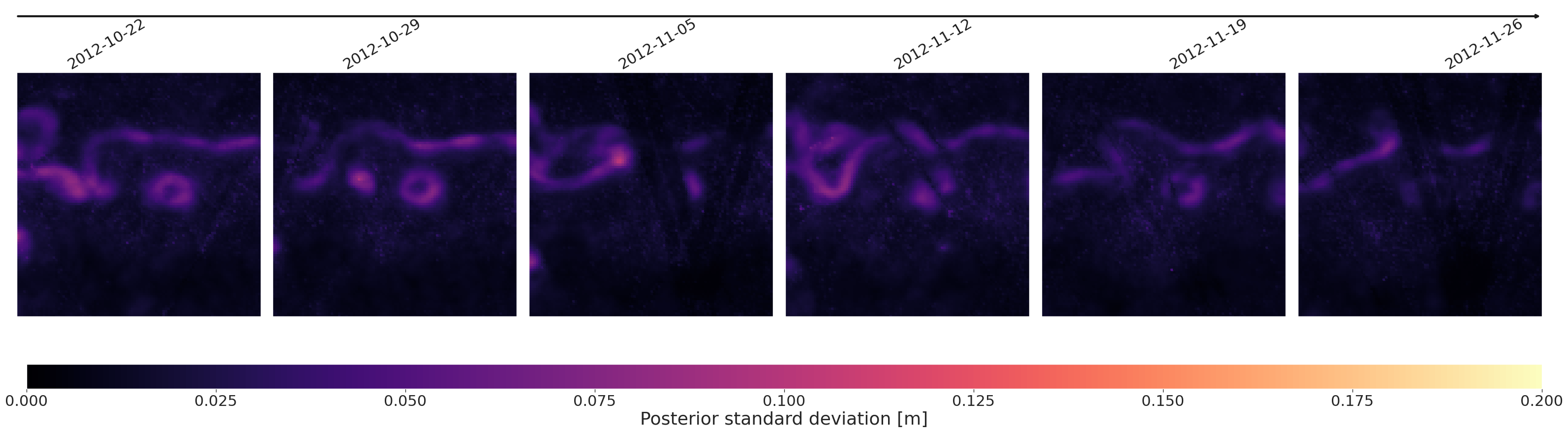}}\\
\subfloat[continuous ranked probability score (CRPS) maps]{\includegraphics[width=15cm]{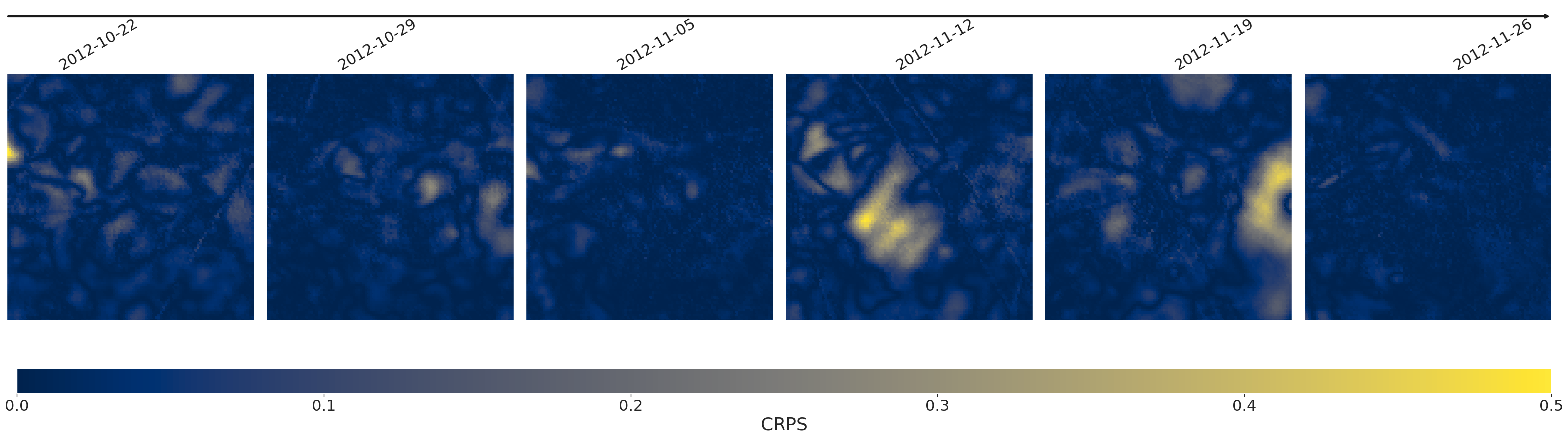}}
\caption{For six days along the OSSE test period: a) corresponding reconstruction error at the center of the assimilation window, b) ensemble-based posterior standard deviations and c) Continuous Ranked Probability Score}
\label{posterior_var_osse}
\end{figure}

The second question is: does this approach, SPDE-based generation of members, followed by their conditioning with observations, is efficient to estimate the posterior pdf $p_{\mathbf{x}|\mathbf{y}}$? Fig. \ref{posterior_var_osse} shows in a) the reconstruction error $\mathbf{x}-\mathbf{x}^\star$ for six days along the test period and in b) the empirical posterior standard deviations computed from 200 members. In c), we also provide pointwise Continuous Ranked Probability Score (CRPS) maps to assess the accurary of the ensemble-based predictions. Given the observations $\y_{ijk}$, we compute the empirical CDF of the stochastic process $X$ at a given spatio-temporal location $\lbrace{i,j,k\rbrace}$ as $F_{ijk}(z)=\mathbf{P}\left[ X_{ijk} \le z \right]$:
\begin{align*}
CRPS(F_{ijk},\y_{ijk}) = \int_{-\infty}^{+\infty} \left( F(z) -\mathbf{1}_{z-\y_{ijk}} \right)^2 dz
\end{align*}
Looking at the estimated standard deviation produced by the ensemble of neural variational reconstructions, they are not as dependent of the observations as an OI scheme would be, which validates the flow-dependency discussed in Section \ref{uq_scheme}. We can still see the observation mask as blurry areas but the standard deviations are rather increasing and continuous along the main meander of the Gulf Stream. The CRPS is often close to zero, which indicates the ensemble of recontruction is wholly accurate with a realistic posterior standard deviation. The highest CRPS values observed are about 0.4 which remains reasonable, and can be explained by high reconstruction errors outside the main Gulf Stream meander that are not correctly handled by the ensemble.

\section{Conclusion}

We explore a new neural architecture to tackle the reconstruction inverse problem of a dynamical process from partial and potentially noisy observations. We provide a joint end-to-end learning scheme of both stochastic prior models and solvers. The idea is to optimize in the same time the state and the stochastic parametrization of the prior so that we minimize the mean squared error between the reconstruction and the true states, while we are also able to provide uncertainty quantification of the mean state reconstruction, either analytically in the gaussian case, or ensemble-derived in the more general configuration.\\

In our work, we draw from recent advances in geostatistical simulations of SPDE-based Gaussian Processes to provide a flexible trainable prior embedded in a neural architecture backboned on variational data assimilation. The SPDE parameters are added as latent variables in an augmented state the trainable solver has to reconstruct. A bi-level optimization scheme is used to optimize in the same time:
\begin{itemize}
\item the inner variational cost derived from OI-based formulations, which depends on state $\mathbf{x}$, observations $\mathbf{y}$ and SPDE parameters $\boldsymbol{\theta}$ and,
\item the outer training loss function of the neural architecture, which drives the optimization of the LSTM-based residual solver parameters $\boldsymbol{\omega}$ leading to the reconstruction of the augmented state. 
\end{itemize}
The first application of the framework on a diffusion-based GP showed that it reaches the same performance, in terms of MSE w.r.t the ground truth, observed when using a neural-based prior in the classic implementation fo the neural scheme \citep{fablet2020joint}, which asymptotically converges towards the optimal solution. In addition, the posterior variance ot the mean state derived from the SPDE parameters is close to the true variance of the Optimal Interpolation, which demonstrates the potentiality of the proposed scheme to handle uncertainty quantifications. Indeed, if we showed that retrieving the original set of SPDE parameters might be difficult, the local minimum found after the training process leads to a parametrization with high likelihood, similar diffusion-based spatial patterns and spatio-temporal covariances structures. Though not always physically explainable, the SPDE prior formulation still helps to interprete the dynamical process in terms of statistical properties. \\
In a more general setup, we also present an application on Sea Surface Height dynamics based on Observation System Simulation Experiment (OSSE), for which the ground truth is given by a state-of-the-art ocean model and pseudo-observations are generated by a realistic satellite subsampling of the ground truth. In this case, the process is not linear and the gaussian framework does not apply. Then, the idea is to use the GP linear SPDE formulation as a surrogate model to linearize the prior dynamics along the data assimilation window. It provides an efficient way for fast sampling of a huge set of members in the prior distribution within a few minutes. Based on the neural solver, able to handle non-linear and non-gaussian dynamics based on its supervised training, the conditioning of these simulations leads to the estimation of the posterior distribution. The preliminary conclusions made on the GP experiment holds for the reconstruction of the mean state, i.e. no significant differences observed when using a trainable neural prior. The key aspect of the framework is here revealed by the SPDE parametrization, which is fully non-stationary in space and time, and allows for online estimation after training for any new set of input observations, in contrast with other approaches in most of the spatial statistics literature which requires offline new parameter inference. We showed that the prior parametrization is both statistically consistent with the original ground truth used in the training and physically sounded with similar patterns observed in the simulations. Comparing the posterior pdf retrieved from the ensemble members, higher variances are observed in areas of high reconstruction errors which is a good indicator of the framework capability to quantify realistic uncertainties. Additional applications would be necessary on real datasets to complement these preliminary conclusions. \\

Regarding the potential extensions of this methodology, it is important to understand that such an SPDE parametrization of the prior term is a way of going to generative modelling, as it is called in the machine learning community. In our case, the SPDE is linear and already provide an efficient way to produce fast, large and realistic ensembles, both in terms of spatial rendering and likelihood. Promising avenues would be to draw from the existing link between diffusion models and SDE to enrich our framework, provide stochastic non-linear priors and see if and how it helps to improve the results obtained in this work. Also, we made the choice to model the prior with a linear SPDE before conditioning it with our neural solver. A direct use of the SPDE formulation for the posterior would have been possible, but restrictive to GP reconstructions which is in most cases rather limited. Back to the non-linear neural diffusion operators, an other way of addressing the problem would be to directly optimize the sampling in the posterior pdf. \\

Regarding how this work is purely idealized or not, several options could be considered to apply it in real-world: make the loss function compliant with observation-only datasets, which implied to change the likelihood-based regularization term, see e.g. \citet{clarotto_2022}, or use transfer learning technique to apply what is learnt with an idealized Ground Truth on real observations, which has already been successfully tested in the original approach, see e.g. \citet{febvre_2022}. Such strategies will be considered in the framework of this original 4DVarNet scheme extension in future works.

\acknowledgments

This study was supported by Public Funds through Ministère de l’Education Nationale, Ministère de l’Enseignement Supérieur et de la Recherche, Fonds Européen de DEveloppement Régional (FEDER), Région Bretagne, Conseil Général du Finistère, Brest Métropole. This work has also been supported by the ANR Projects Melody and OceaniX and GENCI-IDRIS (grant no. 2020-101030).
%
%
\datastatement
The open-source 4DVarNet with SPDE priors version of the code is available at https://github.com/CIA-Oceanix/4dvarnet-starter/tree/maximebeauchamp. The datasets are shared through the ocean data challenge 2020a also available on GitHub \url{https://github.com/ocean-data-challenges/2020a\_SSH\_mapping\_
NATL60, last access: 2022}

\clearpage
\appendix[A] 

\appendixtitle{Finite difference schemes for SPDE}

\label{App:AppendixA}

In this work, we propose to use the Euler implicit scheme as a discretization method for the stochastic PDEs. We need to define the discretized version of several differential operators on the 2D regular grid, namely:

\subsection{Discretization of the spatial differential operators}

\begin{subequations}
\begin{align}
\Delta \mathbf{x}_{i,j} &= \frac{\partial^2{\mathbf{x}}}{\partial{x}^2}_{i,j} +  \frac{\partial^2{\mathbf{x}}}{\partial{y}^2}_{i,j} \nonumber \\
& = \frac{\partial}{\partial{x}} \left(\frac{\partial{\mathbf{x}}}{\partial{x}}\right)_{i,j} 
+ \frac{\partial}{\partial{y}} \left(\frac{\partial{\mathbf{x}}}{\partial{y}}\right)_{i,j} \nonumber \\
& = \frac{\partial}{\partial{x}} \left(\frac{\mathbf{x}_{i+1,j}-\mathbf{x}_{i-1,j}}{2d{x}} \right) + \frac{\partial}{\partial{y}} \left(\frac{\mathbf{x}_{i,j+1}-\mathbf{x}_{i,j-1}}{2d{y}} \right) \nonumber \\
& =  \left( \frac{\mathbf{x}_{i+1,j}}{dx^2} - \frac{\mathbf{x}_{i,j}}{dx^2} \right) -
     \left( \frac{\mathbf{x}_{i,j}}{dx^2} - \frac{\mathbf{x}_{i-1,j}}{dx^2} \right) \nonumber \\         
&+   \left( \frac{\mathbf{x}_{i,j+1}}{dy^2} - \frac{\mathbf{x}_{i,j}}{dy^2} \right) -
     \left( \frac{\mathbf{x}_{i,j}}{dy^2} - \frac{\mathbf{x}_{i,j-1}}{dy^2} \right) \nonumber \\
& = \frac{\mathbf{x}_{i+1,j} -2\mathbf{x}_{i,j} + \mathbf{x}_{i-1,j}}{dx^2} + \frac{\mathbf{x}_{i,j+1} -2\mathbf{x}_{i,j} + \mathbf{x}_{i,j-1}}{dy^2} 
\end{align}
\begin{align}
\nabla \mathbf{H} \cdot \nabla \mathbf{x}_{i,j} &= \left( \mathbf{H}^{1,1}\frac{\partial}{\partial{x}^2} + \mathbf{H}^{1,2}\frac{\partial}{\partial{x}\partial{y}}+ \mathbf{H}^{2,1}\frac{\partial}{\partial{x}\partial{y}} + \mathbf{H}^{2,2}\frac{\partial}{\partial{y}^2} \right) \mathbf{x}_{i,j} \nonumber \\
& = \mathbf{H}^{1,1} \frac{\mathbf{x}_{i+1,j} -2\mathbf{x}_{i,j} + \mathbf{x}_{i-1,j}}{dx^2} \nonumber \\
& + \mathbf{H}^{2,2} \frac{\mathbf{x}_{i,j+1} -2\mathbf{x}_{i,j} + \mathbf{x}_{i,j-1}}{dy^2} \nonumber \\
& + \mathbf{H}^{1,2} \frac{\mathbf{x}_{i+1,j+1} -\mathbf{x}_{i+1,j-1} - \mathbf{x}_{i-1,j+1} + \mathbf{x}_{i-1,j-1}}{4dxdy} \nonumber \\
& + \mathbf{H}^{2,1} \frac{\mathbf{x}_{i+1,j+1} -\mathbf{x}_{i-1,j+1} - \mathbf{x}_{i+1,j-1} + \mathbf{x}_{i-1,j-1}}{4dxdy} \nonumber \\
& = \mathbf{H}^{1,1} \frac{\mathbf{x}_{i+1,j} -2\mathbf{x}_{i,j} + \mathbf{x}_{i-1,j}}{dx^2} \nonumber \\
& + \mathbf{H}^{2,2} \frac{\mathbf{x}_{i,j+1} -2\mathbf{x}_{i,j} + \mathbf{x}_{i,j-1}}{dy^2} \nonumber \\
& + \mathbf{H}^{1,2} \frac{\mathbf{x}_{i+1,j+1} -\mathbf{x}_{i+1,j-1} - \mathbf{x}_{i-1,j+1} + \mathbf{x}_{i-1,j-1}}{2dxdy} \nonumber \\
\end{align}
\begin{align}
\mathbf{m} \cdot  \nabla \mathbf{x}_{i,j} & = \mathbf{m^1} \frac{\partial}{\partial{x}}\mathbf{x}_{i,j} +  \mathbf{m^2} \frac{\partial}{\partial{y}}\mathbf{x}_{i,j} \nonumber \\
& =  \mathbf{m^1} \frac{\mathbf{x}_{i+1,j}-\mathbf{x}_{i-1,j}}{2dx} +  \mathbf{m^2} \frac{\mathbf{x}_{i,j+1}-\mathbf{x}_{i,j-1}}{2dy}
\end{align}
\end{subequations}

\subsection{Upwind schemes for advection-dominated SPDE}

With such an advection-diffusion framework detailed in Section \ref{spde_param}, it is known that the solution to the space-centered scheme does not oscillate only when the Peclet number is lower than 2 and the Courant–Friedrichs–Lewy condition (CFL) condition $Cr=dt\left( \mathbf{m}^1/dx + \mathbf{m}^2/dy \right) \le 1$ is satisfied \citep{Lewy_1928, Price_1966}. For  unsatisfied Peclet conditions, damped oscillations occur with nonreal eigenvalues \citep{Finlayson_1992, Price_1966}, while in the limiting case of pure advection $\mathbf{H} \rightarrow \mathbf{0}$, such a scheme would be unconditionally unstable \citep{Finlayson_1992, Strikwerda_1989}. Because the velocity fiels is allowed to vary in space and time, the CFL number is different at each discrete space-time location $(i,j,t)$. A necessary condition for convergence is that the CFL condition be satisfied at each point location, the velocity and diffusion parameters being unknown, we have to choose the timestep $dt$ small enough so that the maximum CFL number (observed in space at each time step) satisfies the CFL condition. Again, because the velocity field is trained, it may happen that the CFL condition is not satisfied if the latter is more and more dominant during the training. A simple way of counteracting this problem is to use an activation function on the two velocity components by clipping their maximum value.\\

One way of addressing this problem of stabilities in FDM when the advection term is predominant over the diffusion relates to the class of upwind schemes (UFDM). It is used to numerically simulate more properly the direction of propagation of the state in a flow field. The first order upwind FDM uses a one-sided finite difference in the upstream direction to approximate the advection term in the transport SPDE. The spatial accuracy of the first-order upwind scheme can be improved by choosing a more accurate finite difference stencil for the approximation of spatial derivative. Let note that UFDM scheme for SPDE eliminates the nonphysical oscillations in the space-centered scheme and generate stable solutions even for very complicated flows.

\subsubsection{First-order upwind scheme (UFDM1)}

Instead of using centered differences:
\begin{subequations}
\begin{align}
\left(\frac{\partial{\mathbf{x}}}{\partial{x}}\right)_{i,j} = \frac{\mathbf{x}_{i+1,j}-\mathbf{x}_{i-1,j}}{2dx}
\end{align}
\begin{align}
\left(\frac{\partial{\mathbf{x}}}{\partial{y}}\right)_{i,j} = \frac{\mathbf{x}_{i,j+1}-\mathbf{x}_{i,j-1}}{2dy}
\end{align}
\end{subequations}

We use the one-sided upwind differences : 
\begin{subequations}
\begin{align}
\begin{cases}
\displaystyle \left(\frac{\partial{\mathbf{x}}}{\partial{x}}\right)_{i,j} = \frac{\mathbf{x}_{i,j}-\mathbf{x}_{i-1,j}}{dx}
\ \mathrm{if} \  \mathbf{m}^1_{i,j}>0 \\
\displaystyle \left(\frac{\partial{\mathbf{x}}}{\partial{x}}\right)_{i,j} = \frac{\mathbf{x}_{i+1,j}-\mathbf{x}_{i,j}}{dx}
\ \mathrm{if} \  \mathbf{m}^1_{i,j}<0
\end{cases}
\end{align}
and 
\begin{align}
\begin{cases}
\displaystyle \left(\frac{\partial{\mathbf{x}}}{\partial{y}}\right)_{i,j} = \frac{\mathbf{x}_{i,j}-\mathbf{x}_{i,j-1}}{dy}
\ \mathrm{if} \  \mathbf{m}^2_{i,j}<0 \\
\displaystyle \left(\frac{\partial{\mathbf{x}}}{\partial{y}}\right)_{i,j} = \frac{\mathbf{x}_{i,j+1}-\mathbf{x}_{i,j}}{dy}
\ \mathrm{if} \  \mathbf{m}^2_{i,j}<0
\end{cases}
\end{align}
\end{subequations}

\subsubsection{Third-order upwind scheme (UFDM3)}

It can be shown, see e.g. \citep{Eewing_2001} that the UFDM scheme is actually a second-order approximation of the SPDE with a modified diffusion term. Along this line, it comes with the family of methods that may introduce excessive numerical diffusion in the solution with large gradients. Thus, we use a third order upwind scheme for the approximation of spatial derivatives with four points instead of two, with only a reduced increase in the degree of sparsity of the precision matrix. This scheme is less diffusive compared to the second-order accurate scheme. It comes with four points instead of two for the approximation, with only a reduced increase in the degree of sparsity of the discretized differential operator.

It can be expressed as follows:
\begin{subequations}
\begin{align}
\begin{cases}
 \displaystyle \left(\frac{\partial{\mathbf{x}}}{\partial{x}}\right)_{i,j} = \frac{2\mathbf{x}_{i+1,j} + 3\mathbf{x}_{i,j} -6\mathbf{x}_{i-1,j} + \mathbf{x}_{i-2,j}}{6dx}\ \mathrm{if}\ \mathbf{m}^1_{i,j} >0 \\
 \displaystyle \left(\frac{\partial{\mathbf{x}}}{\partial{x}}\right)_{i,j} = \frac{-\mathbf{x}_{i+2,j} +6 \mathbf{x}_{i+1,j}-3\mathbf{x}_{i,j}-2\mathbf{x}_{i-1,j}}{6dx}\ \mathrm{if}\ \mathbf{m}^1_{i,j} <0 
\end{cases}
\end{align}
and 
\begin{align}
\begin{cases}
 \displaystyle \left(\frac{\partial{\mathbf{x}}}{\partial{y}}\right)_{i,j} = \frac{2\mathbf{x}_{i,j+1} + 3\mathbf{x}_{i,j} -6\mathbf{x}_{i,j-1} + \mathbf{x}_{i,j-2}}{6dy}\ \mathrm{if}\ \mathbf{m}^2_{i,j} >0 \\
 \displaystyle \left(\frac{\partial{\mathbf{x}}}{\partial{y}}\right)_{i,j} = \frac{-\mathbf{x}_{i,j+2} +6 \mathbf{x}_{i,j+1}-3\mathbf{x}_{i,j}-2\mathbf{x}_{i,j-1}}{6dy}\ \mathrm{if}\ \mathbf{m}^2_{i,j} <0 
\end{cases} 
\end{align}
\end{subequations}

\subsection{Discretization of the spatio-temporal SPDE}

Based on the discretization of the spatial operators using upwind finite difference schemes for the advection term and centered difference schemes for the diffusion term, we involve an implicit Euler scheme to solve the advection-diffusion SPDE:
\begin{align*} 
\frac{\partial{\mathbf{x}}}{\partial{t}}+\left\lbrace \boldsymbol{\kappa}^2(\mathbf{s},t) + \mathbf{m}(\mathbf{s},t) \cdot \nabla - \nabla \cdot\mathbf{H}(\mathbf{s},t)\nabla \right \rbrace^{\alpha/2} \mathbf{x}(\mathbf{s},t)=\boldsymbol{\tau}(\mathbf{s},t) \mathbf{z}(\mathbf{s},t)
\end{align*}

Because the upwind advection schemes is not symmetric and does not involve the same neighbours in the difference scheme approximation according to the predominant flow direction, we introduce the following generic notations. 
\begin{align*}
\begin{cases}
 \displaystyle \mathbf{m}^{1,t,-}_{i,j} =\left(\frac{\partial{\mathbf{x}}}{\partial{x}}\right)^t_{i,j} \ \mathrm{if}\ \mathbf{m}^{1,t}_{i,j} >0 \\
 \displaystyle \mathbf{m}^{1,t,+}_{i,j} = \left(\frac{\partial{\mathbf{x}}}{\partial{x}}\right)^t_{i,j} \ \mathrm{if}\ \mathbf{m}^{1,t}_{i,j} <0 
\end{cases} \ \mathrm{and} \ 
\begin{cases}
 \displaystyle \mathbf{m}^{2,t,-}_{i,j} = \left(\frac{\partial{\mathbf{x}}}{\partial{y}}\right)^t_{i,j} \ \mathrm{if}\ \mathbf{m}^{2,t}_{i,j} >0 \\
 \displaystyle \mathbf{m}^{2,t,+}_{i,j} = \left(\frac{\partial{\mathbf{x}}}{\partial{y}}\right)^t_{i,j} \ \mathrm{if}\ \mathbf{m}^{2,t}_{i,j} <0 
\end{cases} 
\end{align*}

and by denoting $\mathbf{a}^{1,t,+}_{i,j} = \max(\mathbf{m}^{1,t}_{i,j},0)$, $\mathbf{a}^{1,t,-}_{i,j} = \min(\mathbf{m}^{1,t}_{i,j},0)$, $\mathbf{a}^{2,t,+}_{i,j} = \max(\mathbf{m}^{2,t}_{i,j},0)$, $\mathbf{a}^{2,t,-}_{i,j} = \min(\mathbf{m}^{2,t}_{i,j},0)$, the resulting UFDM, whatever the order of the scheme, can be written in its compact form as:
\begin{align*}
\mathbf{x}^{t+1}_{i,j} = \mathbf{x}^{t}_{i,j} + dt &\Big[\kappa^t_{i,j}\mathbf{x}^t_{i,j} + 
\left( \mathbf{a}^{1,t,+}_{i,j}\mathbf{m}^{1,t,-}_{i,j} + \mathbf{a}^{1,t,-}_{i,j}\mathbf{m}^{1,t,+}_{i,j}\right) + 
\left( \mathbf{a}^{2,t,+}_{i,j}\mathbf{m}^{2,t,-}_{i,j} + \mathbf{a}^{2,t,-}_{i,j}\mathbf{m}^{2,t,+}_{i,j}\right) \\
&+ \mathbf{H}^{1,1,t}_{i,j} \frac{\mathbf{x}^{t}_{i+1,j} -2\mathbf{x}^{t}_{i,j} + \mathbf{x}^{t}_{i-1,j}}{dx^2} + \mathbf{H}^{2,2,t}_{i,j} \frac{\mathbf{x}^{t}_{i,j+1} -2\mathbf{x}^{t}_{i,j} + \mathbf{x}^{t}_{i,j-1}}{dy^2} \\
& + \mathbf{H}^{1,2,t}_{i,j} \frac{\mathbf{x}^{t}_{i+1,j+1} -\mathbf{x}^{t}_{i+1,j-1} - \mathbf{x}^{t}_{i-1,j+1} + \mathbf{x}^{t}_{i-1,j-1}}{2dxdy} + \tau^t_{i,j} \mathbf{z}_{i,j} \Big]
\end{align*}

Using again notation $i=\lfloor {k/N_x} \rfloor$ and $j=k \ \mathrm{mod}\ {N_x}$, operator $\mathbf{A}_t$ associated to UFDM1 finally writes:
\begin{equation}
\label{DF}
\mathbf{A}_{k,l}(t) =
\left.
\begin{cases}
&\mathbf{H}^{1,2,t}_{i,j}/2dxdy \text{ if } l=k\pm (N_x+1) \\
&\mathbf{H}^{1,2,t}_{i,j}/2dxdy \text{ if } l=k\pm (N_x-1) \\
&-\mathbf{H}^{1,1,t}_{i,j}/dx^2 + \left( \mathbf{a}^{1,t,+}_{i,j} + \mathbf{a}^{1,t,-}_{i,j} \right)/dx \text{ if } l=k-1\\
&-\mathbf{H}^{1,1,t}_{i,j}/dx^2 + \left( \mathbf{a}^{1,t,+}_{i,j} + \mathbf{a}^{1,t,-}_{i,j} \right)/dx \text{ if } l=k+1\\
&-\mathbf{H}^{2,2,t}_{i,j}/dy^2 + \left( \mathbf{a}^{2,t,+}_{i,j} + \mathbf{a}^{1,t,-}_{i,j} \right)/dy \text{ if } l=k-N_x\\
&-\mathbf{H}^{2,2,t}_{i,j}/dy^2 + \left( \mathbf{a}^{2,t,+}_{i,j} + \mathbf{a}^{1,t,-}_{i,j} \right)/dy \text{ if } l=k+N_x\\
&\left(\kappa^t_{i,j}\right)^2+2(\mathbf{H}^{1,1,t}_{i,j}/dx^2+\mathbf{H}^{2,2,t}_{i,j}/dy^2)\\
& + \left( \mathbf{a}^{1,t,+}_{i,j} + \mathbf{a}^{1,t,-}_{i,j} \right)/dx + \left( \mathbf{a}^{2,t,+}_{i,j} + \mathbf{a}^{2,t,-}_{i,j} \right)/dy \text{ if } k=l \\
&0 \mathrm{\ otherwise}
\end{cases}
\right.
\end{equation}

The same operator associated to UFDM3 writes:
\footnotesize
\begin{align}
\label{DF}
&\mathbf{A}_{k,l}(t) = \nonumber \\
&\left.
\begin{cases}
\mathbf{H}^{1,2,t}_{i,j}/2dxdy &\text{ if } l=k\pm (N_x+1) \\
\mathbf{H}^{1,2,t}_{i,j}/2dxdy &\text{ if } l=k\pm (N_x-1) \\
 \mathbf{a}^{1,t,+}_{i,j}/6dx  &\text{ if } l=k-2\\
-\mathbf{H}^{1,1,t}_{i,j}/dx^2 - \left( 6\mathbf{a}^{1,t,+}_{i,j} + 2\mathbf{a}^{1,t,-}_{i,j} \right)/6dx &\text{ if } l=k-1\\
-\mathbf{H}^{1,1,t}_{i,j}/dx^2 + \left( 2\mathbf{a}^{1,t,+}_{i,j} + 6\mathbf{a}^{1,t,-}_{i,j} \right)/6dx &\text{ if } l=k+1\\
 -\mathbf{a}^{1,t,-}_{i,j}/6dx &\text{ if } l=k+2 \\
 \mathbf{a}^{2,t,+}_{i,j}/6dy &\text{ if } l=k-2Nx\\
-\mathbf{H}^{2,2,t}_{i,j}/dy^2 - \left( 6\mathbf{a}^{2,t,+}_{i,j} + 2\mathbf{a}^{2,t,-}_{i,j} \right)/6dy &\text{ if } l=k-N_x\\
-\mathbf{H}^{2,2,t}_{i,j}/dy^2 + \left( 2\mathbf{a}^{2,t,+}_{i,j} + 6\mathbf{a}^{2,t,-}_{i,j} \right)/6dy &\text{ if } l=k+N_x\\
 - \mathbf{a}^{2,t,-}_{i,j}/6dy &\text{ if } l=k+2Nx\\
\left(\kappa^t_{i,j}\right)^2+2(\mathbf{H}^{1,1,t}_{i,j}/dx^2+\mathbf{H}^{2,2,t}_{i,j}/dy^2) &\\
 + 3\left( \mathbf{a}^{1,t,+}_{i,j} - \mathbf{a}^{1,t,-}_{i,j} \right)/6dx & \\ + 3\left( \mathbf{a}^{2,t,+}_{i,j} - \mathbf{a}^{2,t,-}_{i,j} \right)/6dy &\text{ if } k=l 
\end{cases}
\right.
\end{align}
\normalsize
and 0 otherwise.

\newpage

\appendix[B] 

\appendixtitle{SPDE-based precision matrix}
\label{App:AppendixB}

Let define the spatio-temporal SPDE prior $\mathbf{x}=\lbrace \x_0,\cdots,\x_{Ldt} \rbrace$. From now on, $\x_0 \sim \mathcal{N}(\mathbf{0},\mathbf{P_0})$ denotes the initial state and $\mathbf{Q}_0=\mathbf{P_0}^{-1}$ is always taken as the precision matrix obtained after a given stabilization run, i.e. the evolution of the dynamical system over $N$ timesteps using as stationary parameters the initial parametrization $\boldsymbol{\theta}_0$ of the SPDE at time $t=0$. 
We can rewrite :
\[ \lbrace \x_0,\cdots,\x_{Ldt} \rbrace = \mathbf{M}_G \left[\begin{array}{c}\mathbf{x}_0\\ \mathbf{z}\end{array}\right]\]

with $\mathbf{z}=[\mathbf{z}_1,\dots,\mathbf{z}_t]^{\mathrm{T}}$ and 

\[\mathbf{M}_G = \left[\begin{array}{ccccccc} 
\mathbf{I} & 0  & 0 & 0 & 0 &\dots& 0 \\
\mathbf{M}_1 & \mathbf{T}_1 & 0 &0  & 0 & \dots &0 \\
\mathbf{M}_2\mathbf{M}_1 & \mathbf{M}_2\mathbf{T}_1 & \mathbf{T}_2 & 0& 0 &\dots& 0\\
\mathbf{M}_3\mathbf{M}_2\mathbf{M}_1 & \mathbf{M}_3\mathbf{M}_2\mathbf{T}_1 & \mathbf{M}_3\mathbf{T}_2 & \mathbf{T}_3 & 0 &\dots & 0\\
\vdots &\ddots &\ddots &\ddots & \ddots & \ddots & 0 \\
\vdots &\ddots &\ddots &\ddots & \ddots & \ddots & 0 \\
\vdots &\ddots &\ddots &\ddots & \ddots & \ddots & \mathbf{T}_L \\
\end{array}\right]\]

Despite its apparent complexity, $\mathbf{M}_G$ has a particular structure which allows to easily compute its inverse:
\[\mathbf{M}_G^{-1} = \left[\begin{array}{ccccccc} 
\mathbf{I}   & 0   & 0 & 0 & 0 &\dots& 0 \\
-\mathbf{T}_1^{-1}\mathbf{M}_1 & \mathbf{T}_1^{-1} & 0 &0  & 0 & \dots &0 \\
0 & -\mathbf{T}_2^{-1}\mathbf{M}_2 & \mathbf{T}_2^{-1} & 0& 0 &\dots& 0\\
0 & 0 & -\mathbf{T}_3^{-1}\mathbf{M}_3 & \mathbf{T}_3^{-1} & 0 &\dots & 0\\
0 &\ddots &\ddots &\ddots & \ddots & \ddots & 0 \\
\vdots &\ddots &\ddots &\ddots & \ddots & \ddots & 0 \\
0 &\ddots &\ddots &\ddots & \ddots & -\mathbf{T}_L^{-1}\mathbf{M}_L & \mathbf{T}_L^{-1} \\
\end{array}\right]\]

All this information is embedded in $\mathbf{Q}^b$ (Eq. \ref{global_prec_mat}), which is the inverse of the prior covariance matrix $\mathbf{B}$:

\begin{align}
\label{evol_cov}
\mathbf{Q}^b  & = \mathbf{B}^{-1} = 
\begin{bsmallmatrix} 
\mathbf{P}_0 & \mathbf{P}_{0,1} & \dots & \dots & \mathbf{P}_{0,L} \\
\mathbf{P}_{1,0} & \mathbf{P}_1 & \dots & \dots & \mathbf{P}_{1,L} \\
\vdots & \mathbf{P}_{2,1} & \mathbf{P}_2 & \dots & \mathbf{P}_{2,L} \\
\vdots &\ddots &\ddots &\ddots & \vdots \\
\mathbf{P}_{L-1,0} & \dots & \dots & \mathbf{P}_{L-1}  & \mathbf{P}_{L-1,L} \\
\mathbf{P}_{L,0} & \mathbf{P}_{L,1} & \dots & \mathbf{P}_{L,L-1}  & \mathbf{P}_L \\
\end{bsmallmatrix}^{-1} = {\mathbf{M}_G^{-1}}^{\mathrm{T}}\left[\begin{array}{llll}
\mathbf{P}_0^{-1} & 0 & \dots & 0\\
0  & \mathbf{I} &\dots & 0\\
\vdots & \ddots & \ddots & \vdots\\
0 & 0 & \dots & \mathbf{I}
\end{array}\right] \mathbf{M}_G^{-1}
\end{align}

By denoting $\mathbf{S}_k=\mathbf{T}_k\mathbf{T}_k^{\mathrm{T}}$, we have

\begin{align}
\label{global_prec_mat}
\mathbf{Q}^b  = \begin{bsmallmatrix} 
\mathbf{P}_0^{-1}+\mathbf{M}_1^{\mathrm{T}}\mathbf{S}_1^{-1}\mathbf{M}_1   & -\mathbf{M}_1^{\mathrm{T}}\mathbf{S}_1^{-1}   & 0 & 0 & 0 &\dots& 0 \\
-\mathbf{S}_1^{-1}\mathbf{M}_1 & \mathbf{S}_1^{-1}+\mathbf{M}_2^{\mathrm{T}}\mathbf{S}_2^{-1}\mathbf{M}_2 & -\mathbf{M}_2^{\mathrm{T}}\mathbf{S}_2^{-1}  &0  & 0 & \dots &0 \\
0 & -\mathbf{S}_2^{-1}\mathbf{M}_2 & \mathbf{S}_2^{-1} +\mathbf{M}_3^{\mathrm{T}}\mathbf{S}_3^{-1}\mathbf{M}_3& -\mathbf{M}_3^{\mathrm{T}}\mathbf{S}_3^{-1}& 0 &\dots& 0\\
0 &\ddots &\ddots &\ddots & \ddots & \ddots & 0 \\
0 &\ddots &\ddots &\ddots & \ddots & \ddots & 0 \\
\vdots &\ddots &\ddots &\ddots & -\mathbf{S}_{L-1}^{-1}\mathbf{M}_{L-1} & \mathbf{S}_{L-1}^{-1}+\mathbf{M}_L^{\mathrm{T}}\mathbf{S}_L^{-1}\mathbf{M}_L & -\mathbf{M}_L^{\mathrm{T}}\mathbf{S}_L^{-1} \\
0 &\ddots &\ddots &\ddots & 0 & -\mathbf{S}_{L}^{-1}\mathbf{M}_{L} & \mathbf{S}_L^{-1} \\
\end{bsmallmatrix}
\end{align}

Because of the formulation of $\mathbf{M}_t$ and $\mathbf{T}_t$, the precision matrix $\mathbf{Q}^b $ with the FDM scheme also writes:

\begin{align}
\hspace{-1.5cm}
\label{global_prec_mat_2_appendix}
\mathbf{Q}^b  &= \frac{1}{dt} \begin{bsmallmatrix} 
\mathbf{P}_0^{-1}+\widetilde{\mathbf{Q}}_{s,1}  & -\widetilde{\mathbf{Q}}_{s,1}\mathbf{M}_1^{-1} & 0 & 0 & 0 &\dots& 0 \\
-\left(\mathbf{M}_1^{\mathrm{T}}\right)^{-1}\widetilde{\mathbf{Q}}_{s,1} & \mathbf{M}^{\mathrm{T}}_1\widetilde{\mathbf{Q}}_{s,1}\mathbf{M}_1 +\widetilde{\mathbf{Q}}_{s,2} & -\widetilde{\mathbf{Q}}_{s,2}\mathbf{M}_2^{-1} &0  & 0 & \dots &0 \\
0 & -\left(\mathbf{M}_2^{\mathrm{T}}\right)^{-1}\widetilde{\mathbf{Q}}_{s,2} & \mathbf{M}^{\mathrm{T}}_2\widetilde{\mathbf{Q}}_{s,2}\mathbf{M}_2 + \widetilde{\mathbf{Q}}_{s,3} & -\widetilde{\mathbf{Q}}_{s,3}\mathbf{M}_3^{-1} & 0 & \dots & 0\\
0 &\ddots &\ddots &\ddots & \ddots & \ddots & 0 \\
0 &\ddots &\ddots &\ddots & \ddots & \ddots & 0 \\
\vdots &\ddots &\ddots &\ddots & -\left(\mathbf{M}_{L-1}^{\mathrm{T}}\right)^{-1}\widetilde{\mathbf{Q}}_{s,L-1} & \mathbf{M}^{\mathrm{T}}_{L}\widetilde{\mathbf{Q}}_{s,L-1} \mathbf{M}_{L-1} + \widetilde{\mathbf{Q}}_{s,L} & -\widetilde{\mathbf{Q}}_{s,L}\mathbf{M}_{L}^{-1} \\
0 &\ddots &\ddots &\ddots & 0 & -\left(\mathbf{M}_{L}^{\mathrm{T}}\right)^{-1}\widetilde{\mathbf{Q}}_{s,L} & \mathbf{M}^{\mathrm{T}}_{L}\widetilde{\mathbf{Q}}_{s,L}\mathbf{M}_{L} \\
\end{bsmallmatrix}
\end{align}
where $\widetilde{\mathbf{Q}}_{s,t}$ is the precision matrix of the colored noise weighted by the non-uniform regularization variance $\boldsymbol{\tau}_t$.\\

\appendix[C] 

\appendixtitle{PyTorch implementation of sparse linear algebra}
\label{App:AppendixC}

Currently, two pieces of codes are missing in the PyTorch sparse linear algebra to achieve a fully sparse implementation of our algorithm: the automatic differentiation tools for
\begin{itemize}
\item solving sparse linear systems 
\item running sparse Cholesky decomposition
\end{itemize}

First, regarding the implementation of the backward pass for solving linear systems, we start by writing the forward pass of this system of equation:
\begin{align*}
    x = \mathbf{A}^{-1}\mathbf{b}
\end{align*}
where $\mathbf{A}$ denotes a $2D$ square matrix and $\mathbf{b}$ a one-dimensional vector.\\
We need to provide the gradients wrt both $\mathbf{\mathbf{A}}$ and $\mathbf{\mathbf{b}}$:
\begin{align*}
    \frac{\partial L}{\partial \mathbf{b}} &= \frac{\partial L}{\partial x_i} \frac{\partial x_i}{\partial \mathbf{b}_j} 
        = \frac{\partial L}{\partial x_i} \frac{\partial}{\partial \mathbf{b}_k} ( \mathbf{A}^{-1}_{ij} \mathbf{b}_j )  
        = \frac{\partial L}{\partial x_i} \mathbf{A}^{-1}_{ij} \frac{\partial \mathbf{b}_j}{\partial \mathbf{b}_k}\\
        &= \frac{\partial L}{\partial x_i} \mathbf{A}^{-1}_{ij} \delta_{jk}
        = \frac{\partial L}{\partial x_i} \mathbf{A}^{-1}_{ik} 
        = \big(\mathbf{A}^{-1}\big)^{T} \frac{\partial L}{\partial x} \\
        &= \mathrm{solve}\big( \mathbf{A}^{\mathrm{T}}\,,\,\,  \frac{\partial L}{\partial x} \big) 
\end{align*}

and

\begin{align*}
\frac{\partial L}{\partial \mathbf{A}} 
        &= \frac{\partial L}{\partial x_i} \frac{\partial x_i}{\partial \mathbf{A}_{mn}}  
        = \frac{\partial L}{\partial x_i} \frac{\partial}{\partial \mathbf{A}_{mn}} ( \mathbf{A}^{-1}_{ij} \mathbf{b}_j ) \\
        &= -\frac{\partial L}{\partial x_i} \mathbf{A}^{-1}_{ij} \frac{\partial \mathbf{A}_{jk}}{\partial \mathbf{A}_{mn}} \mathbf{A}^{-1}_{kl} \mathbf{b}_l \\
        &= -\frac{\partial L}{\partial x_i} \mathbf{A}^{-1}_{ij} \delta_{jm} \delta_{kn} \mathbf{A}^{-1}_{kl} \mathbf{b}_l \\
        &= -\frac{\partial L}{\partial x_i} \mathbf{A}^{-1}_{im} \mathbf{A}^{-1}_{nl} \mathbf{b}_l \\
        &= -\left(\big(\mathbf{A}^{-1}\big)^{T} \frac{\partial L}{\partial x}\right)\otimes\left( \mathbf{A}^{-1} \mathbf{b} \right) \\
        &= -\frac{\partial L}{\partial \mathbf{b}} \otimes x
\end{align*}

where we used the einstein summation convention during the derivations, as well as the following identity:
\begin{align*}
\frac{\partial (\mathbf{A}^{-1})}{\partial p} = - \mathbf{A}^{-1} \frac{\partial \mathbf{A}}{\partial p} \mathbf{A}^{-1}.
\end{align*}
These two expressions $\frac{\partial L}{\partial \mathbf{b}}$ and $\frac{\partial L}{\partial \mathbf{A}}$ are easily implemented in PyTorch based on sparse representations.\\

Second, the backward pass for a sparse Cholesky decomposition may be found for instance in \citet{Seeger_2019}: given a symmetric, positive deﬁnite matrix $\mathbf{A}$, its
Cholesky factor $\mathbf{L}$ is lower triangular with positive diagonal, such that the forward pass of the Cholesky decomposition is defined as $\mathbf{A}=\mathbf{L}\mathbf{L}^{\mathrm{T}}$. Given the output gradient $\overline{\mathbf{L}}$ and the Cholesky factor $\mathbf{L}$, the backward pass compute the input gradient $\overline{\mathbf{A}}$ defined as :
\begin{align}
\overline{\mathbf{A}} = \frac{1}{2} \mathbf{L}^{-\mathrm{T}} ltu(\mathbf{L}^{\mathrm{T}}\overline{\mathbf{L}})\mathbf{L}^{-1}
\end{align}
where $ltu(\mathbf{X})$ generates a symmetric matrix by copying the lower triangle to the upper
triangle. In our work, such an operation is useful when optimizing the likelihood in the outer training cost function of the neural scheme, see Eq. \ref{log_det_Q}, which involves the determinant of sparse matrices Cholesky decomposition. 


\bibliographystyle{abbrvnat}
\bibliography{refimta}

\end{document}